\documentclass{article}

\PassOptionsToPackage{numbers,sort&compress}{natbib}
\usepackage[main,preprint]{neurips_2026}
\makeatletter
\renewcommand{\@noticestring}{Accepted for presentation at the 40th Annual Conference on Neural Information Processing Systems (NeurIPS~2026).}
\makeatother

\usepackage[utf8]{inputenc}
\usepackage[T1]{fontenc}

\usepackage[hypertexnames=false]{hyperref}
\usepackage{url}
\usepackage{booktabs}
\usepackage{amsmath,amssymb,amsfonts,amsthm,mathtools}
\usepackage{bm}
\usepackage{graphicx}
\usepackage{microtype}
\usepackage{xcolor}
\usepackage{enumitem}
\usepackage{algorithm}
\usepackage{algpseudocode}
\usepackage{float}
\usepackage{placeins}

\graphicspath{{./}{./manuscript_plots/}{./sim_appendix_plots/}}

\theoremstyle{definition}
\newtheorem{definition}{Definition}
\newtheorem{assumption}{Assumption}
\newtheorem{condition}{Condition}

\theoremstyle{plain}
\newtheorem{theorem}{Theorem}
\newtheorem{proposition}{Proposition}
\newtheorem{corollary}{Corollary}

\theoremstyle{remark}
\newtheorem{remark}{Remark}

\newcommand{\R}{\mathbb{R}}
\newcommand{\E}{\mathbb{E}}
\newcommand{\Pbb}{\mathbb{P}}
\newcommand{\1}{\bm{1}}
\newcommand{\diag}{\operatorname{diag}}
\newcommand{\col}{\operatorname{col}}

\newcommand{\norm}[1]{\left\lVert #1 \right\rVert}

\newcommand{\sinTheta}{\operatorname{sin}\Theta}
\newcommand{\Err}{\operatorname{Err}}

\title{Communication-Efficient Federated Learning of Latent Patient Representations from Multi-Institutional EHRs}

\author{%
  Zhiyu Yan\thanks{Equal contribution.\quad \textsuperscript{\(\dagger\)}Equal contribution.}\\
  Department of Biostatistics\\
  Harvard University\\
  \texttt{romanyan@g.harvard.edu}
  \And
  Shiao Liu\footnotemark[1]\\
  Department of Biostatistics\\
  Harvard University\\
  \texttt{liushiao1@gmail.com}
  \And
  Rahul Mahajan\\
  Department of Neurology\\
  Mass General Brigham\\
  \texttt{rmahajan@mgb.org}
  \AND
  Anand Viswanathan\\
  Department of Neurology\\
  Mass General Brigham\\
  \texttt{aviswanathan1@mgb.org}
  \And
  Christopher D. Anderson\textsuperscript{\(\dagger\)}\\
  Department of Neurology\\
  Mass General Brigham\\
  \texttt{cdanderson@mgb.org}
  \And
  Rui Duan\textsuperscript{\(\dagger\)}\\
  Department of Biostatistics\\
  Harvard University\\
  \texttt{rduan@hsph.harvard.edu}
}

\begin{document}

\maketitle

\begin{abstract}
Electronic health records (EHRs) contain rich health information for uncovering latent patient representations and clinically meaningful subgroups. Learning from EHR data across institutions is valuable because representations learned at a single site may reflect local data patterns and lack generalizability, yet pooling patient-level records across institutions is often impractical because of privacy constraints and communication costs. We propose a communication-efficient, non-iterative federated spectral method for learning shared latent representation components from multi-institutional EHRs, in which each institution shares aggregate marginal frequencies and a low-rank spectral summary while keeping patient-level records local. A central server aggregates these summaries with the pooled marginal frequencies to recover both the shared low-rank structure and the canonical row-normalization direction required for interpretable representation recovery. We prove row-wise recovery guarantees and error bounds, with a leading federated variance term matching a corresponding lower bound under explicit regularity conditions. Simulations show that our method closely matches pooled analysis in estimating latent representation components, improves as more sites contribute summaries, and outperforms single-site and existing federated benchmarks across a range of data settings. Applied to real-world multi-facility EHR data for stroke patients, the method learns interpretable patient representations that highlight clinically distinct subgroups and could help inform subgroup-specific care.
\end{abstract}

\section{Introduction}\label{sec:intro}

Electronic health records (EHRs) provide a broad view of patient care and are increasingly
used to generate real-world evidence, characterize disease phenotypes, support risk
stratification, and inform clinical decisions. Shaped by complex care
processes, EHR data are inherently high-dimensional, heterogeneous, and sparse. Patient representation learning offers a way to summarize these complex records into lower-dimensional latent variables or distributions that can support downstream subgroup discovery, outcome prediction, and generative modeling, which help guide more personalized care strategies~\citep{zhang2022clinical,khurshid2022cohort,simmons2016text, ding2022data,
miceli2023ai_stroke,ryu2024deep}.

Arranging EHRs into concept-by-patient counts provides a clean input format for patient
representation learning. Here \emph{concepts} are derived clinical
data fields that can be counted within a patient record. Structured data may contribute
diagnosis, medication, procedure, or laboratory concepts, while unstructured notes may
contribute concepts after tokenization or other natural-language preprocessing. This count-based format naturally fits topic modeling, in which each
topic is a shared probability distribution over concepts and each patient
is represented by nonnegative topic-membership weights that sum to
one~\citep{hofmann1999probabilistic,blei2003latent,zhang2023postacute,ye2024dynamic}.
We refer to the topics as shared representation components and to a
patient's membership-weight vector as the patient representation.
This parameterization gives a direct interpretation to both the
concept-level loadings and the patient-level coordinates.

Learning patient representations from EHR data collected at multiple institutions is
important for improving generalizability, increasing sample size, and studying rare clinical events. Representation components learned at a
single hospital may instead reflect local patient case mix or care processes rather than more general disease patterns. However,
directly pooling patient-level EHR records across institutions is often infeasible due to data-governance restrictions, privacy concerns,
infrastructure barriers, and the computational and communication costs of centralizing large clinical
datasets.

We propose Federated Topic-SCORE, a communication-efficient non-iterative spectral method for learning shared
latent patient representation components from multi-institutional EHRs. The method
builds on the singular value decomposition (SVD)-based Topic-SCORE framework~\citep{Ke02012024}, which uses low-rank
spectral coordinates, row normalization, and simplex geometry to estimate topic
loadings. Standard Topic-SCORE, however, assumes access to the
pooled concept-by-patient matrix, which is precisely what multi-institutional EHR
settings often cannot provide. In Federated Topic-SCORE, each site keeps patient-level records local and sends only a
marginal concept-frequency vector and a low-rank SVD summary. The server combines
these summaries to recover the shared low-rank structure and the normalization
needed to estimate the global representation component matrix.

Federated Topic-SCORE is distinguished from existing related work in two respects. First,
existing federated representation learning and topic modeling methods, such as federated latent Dirichlet allocation (LDA) and federated neural topic modeling, typically rely on iterative 
updates within collaborating sites, iterative communication across sites, or both~\citep{jiang2019,Ma_2023,
calvo2023federatedneuraltopicmodels, wang2020fedlda, kushwaha2025federated}. In iterative federated learning settings, each site repeats local computation, summary preparation and transfer, personnel coordination, and waiting for other sites in each iteration, which can significantly impede the practical deployment of the federated algorithm~\citep{tong2025unlocking,jiang2023agedependent}. Federated Topic-SCORE instead estimates the shared topic-loading matrix through a non-iterative spectral scheme, reducing both computation and communication costs. Second, a related line of work on distributed principal component analysis and shared singular-subspace estimation also
aggregates local spectral information across sites~\citep{fan2019distributed,
ma2024optimalestimationsharedsingular}. However, these methods only focus on subspace
estimation, whereas Federated Topic-SCORE further recovers the row-normalized spectral
coordinates and simplex vertices needed for interpretable nonnegative topic loadings.

The resulting federated estimator of topic loadings comes with row-wise recovery guarantees for the normalized
spectral coordinates used in the simplex step, as well as error bounds
whose leading federated variance term matches a corresponding lower bound under
explicit regularity conditions. Simulations show that our method closely matches the
pooled Topic-SCORE without sharing patient-level records, improves as more sites
participate, and outperforms single-site and existing federated topic modeling benchmarks under varying parameter and data
settings. In a multi-facility EHR application, Federated Topic-SCORE
recovers representation components closer to the results obtained from the pooled data than any single-site analysis
and produces interpretable representations whose dominant memberships highlight patient subgroups
with distinct survival patterns.

Our contributions are:
\begin{itemize}[leftmargin=*,itemsep=1pt,topsep=2pt]
    \item We develop a communication-efficient federated
    spectral algorithm for learning shared latent patient representation components from
    multi-institutional EHR data, without sharing patient-level records or requiring
    iterative cross-site communication.
    \item We establish finite-sample guarantees for our method, including
    row-wise control of the normalized spectral coordinates used for latent component recovery and error bounds under heterogeneous site-level membership mixtures.
    \item We demonstrate the improved performance of our method over comparable benchmarks through simulations, and illustrate the utility of our method in a real-world multi-facility EHR application.
\end{itemize}

\section{Problem setup and method}\label{sec:method}

\subsection{Federated latent representation model}\label{subsec:model}

Suppose there are \(M\) institutions
sharing a common, consistently indexed vocabulary of \(p\) EHR concepts and a central
server that aggregates summaries from all sites. Site \(m\) has a sample size of \(n_m\) and forms the concept-by-patient frequency matrix
\(D^{(m)}=[d_1^{(m)},\ldots,d_{n_m}^{(m)}]\in\mathbb R^{p\times n_m}\), where each
column is the patient's raw concept-count vector normalized by its total count. Write
\(D=[D^{(1)},\ldots,D^{(M)}]\) and \(n_{\rm tot}=\sum_m n_m\).

Let \(K\) denote the prescribed number of latent representation components.
The shared components are parameterized by the topic-loading matrix
\(A=[A_{\cdot1},\ldots,A_{\cdot K}]\in\mathbb R^{p\times K}\), whose columns are
probability distributions over EHR concepts. Patient \(i\) at site \(m\) has
membership weights \(w_i^{(m)}\). Specifically, \(A_{\cdot k}\in\Delta^{p-1}\) and
\(w_i^{(m)}\in\Delta^{K-1}\), where
\(\Delta^{q-1}=\{x\in\mathbb R^q:x\ge0,\mathbf 1^\top x=1\}\).
For simplicity, in the analysis below, we assume a common total concept count \(N\) per patient, so that \(N d_i^{(m)}\) is the raw concept-count vector. Conditional on these weights,
\begin{equation}\label{eq:plsi-main}
    N d_i^{(m)}\sim {\rm Multinomial}\{N,Aw_i^{(m)}\},
\end{equation}
independently across all patients and sites. The topic-loading matrix \(A\) is shared
across sites, while the membership vectors \(w_i^{(m)}\) may differ across sites.
Thus, institutions may have different patient mixtures while sharing the same latent
representation components. Remark~\ref{rem:unequal-length-proof-note} discusses the
modifications needed for unequal count lengths.

Let \(W^{(m)}=[w_1^{(m)},\ldots,w_{n_m}^{(m)}]\) and
\(W=[W^{(1)},\ldots,W^{(M)}]\). Throughout the theoretical analysis, \(A\) and the
membership matrices are fixed unknown parameters, and all probabilities and
expectations are conditional on \(W\). Define
\(D_0^{(m)}=\mathbb E[D^{(m)}\mid W]=AW^{(m)}\) and
\(D_0=[D_0^{(1)},\ldots,D_0^{(M)}]=AW\). Define the pooled and local membership
second moments as \(\Sigma_W=n_{\rm tot}^{-1}WW^\top\) and
\(\Sigma_{W,m}=n_m^{-1}W^{(m)}W^{(m)\top}\). Each site first sends only its marginal
concept-frequency vector \(\bar d^{(m)}=n_m^{-1}D^{(m)}\1\) to the server. The server aggregates
\(\bar d=n_{\rm tot}^{-1}\sum_{m=1}^M n_m\bar d^{(m)}\) and
\(M_d=\diag(\bar d)\). At the population level, with \(w=n_{\rm tot}^{-1}W\1\),
define \(m_0=Aw\) and \(M_0=\diag(m_0)\).

The spectral step targets the marginal-frequency normalized loading space
\begin{equation}\label{eq:common-space-main}
    \mathcal S=\col(M_0^{-1/2}A).
\end{equation}
The pooled marginal normalization is what places all sites in this common spectral
coordinate system.
If every \(\Sigma_{W,m}\) is nonsingular, then the columns of
\(M_0^{-1/2}D_0^{(m)}\) span \(\mathcal S\). Its left singular vectors may
nevertheless differ across sites because \(\Sigma_{W,m}\) varies with \(m\). Thus, local patient mixtures may differ,
but the population representation component subspace is shared. 

We next specify the regularity conditions used in our analysis.
\begin{assumption}[Regularity: anchor concepts and locally rich mixtures]
\label{ass:regular-plsi}
For \(a_j^\top\), the \(j\)-th row of \(A\), let \(h_j=\|a_j\|_1\),
\(h_{\min}=\min_j h_j\), \(H=\diag(h_1,\ldots,h_p)\), and
\(\Sigma_A=A^\top H^{-1}A\). There exists a constant \(c>0\), independent of
\(M,N,p\) and the site sample sizes \(n_1,\ldots,n_M\), such that: (i) each latent representation component \(k\) has at least one anchor concept,
defined as a row \(j\) with \(A(j,k)>0\) and \(A(j,\ell)=0\) for all \(\ell\ne k\); 
(ii) \(h_{\min}\ge cK/p\); and (iii)
\(\lambda_{\min}(\Sigma_A)\ge c\), \(\lambda_{\min}(\Sigma_W)\ge c\), and
\(\min_{m\le M}\lambda_{\min}(\Sigma_{W,m})\ge c\).
\end{assumption}

The anchor-concept condition enables simplex-based topic recovery.
The lower bound on \(\lambda_{\min}(\Sigma_A)\) prevents degeneracy of the
topic-loading geometry, while the bounds on \(\Sigma_W\) and \(\Sigma_{W,m}\)
ensure sufficiently rich pooled and site-specific membership mixtures.
The local condition requires enough variation within each site to identify
the common signal space. The setting with near-violation of this condition is examined in Appendices~\ref{subsec:upturn} and~\ref{app:sim-heterogeneous}.

\subsection{Canonical coordinate geometry and vertex hunting}
\label{subsec:canonical-main}

Under Assumption~\ref{ass:regular-plsi}, we fix a strictly positive first
basis vector of \(\mathcal S\) using the pooled marginal.

\begin{proposition}[Pooled-marginal canonical direction]
\label{prop:canonical-direction}
Under Assumption~\ref{ass:regular-plsi},
\[
    \xi_1=M_0^{1/2}\1\in\mathcal S,\qquad \|\xi_1\|_2=1,
\]
and \(\xi_1\) is strictly positive. Let \(\Xi_\perp\) be any orthonormal basis of
\(\mathcal S\cap \xi_1^\perp\) and set \(\Xi=[\xi_1,\Xi_\perp]\). Then there is a
unique nonsingular \(V\in\mathbb R^{K\times K}\) such that
\[
    \Xi=M_0^{-1/2}AV,\qquad V_{\cdot,1}=w.
\]
For each concept \(j\), define the oracle ratio point
\(r_j\in\mathbb R^{K-1}\) by \(r_j(\ell)=\Xi_\perp(j,\ell)/\xi_1(j)\),
\(1\le\ell\le K-1\). For each component \(k\), define
\(v_k^\star\in\mathbb R^{K-1}\) by
\(v_k^\star(\ell)=V_{k,\ell+1}/V_{k,1}\).
The concept-level barycentric coordinate vector \(\pi_j\in\Delta^{K-1}\)
has entries
\(\pi_j(k)=A(j,k)V_{k,1}/\sum_{s=1}^K A(j,s)V_{s,1}
=A(j,k)w(k)/m_0(j)\). These quantities satisfy
\begin{equation}\label{eq:oracle-simplex-main}
    r_j=\sum_{k=1}^K\pi_j(k)v_k^\star,
    \qquad
    A\diag(w)=M_0^{1/2}\diag(\xi_1)\Pi,
    \qquad \Pi=[\pi_1,\ldots,\pi_p]^\top .
\end{equation}
An anchor concept for component \(k\) has \(r_j=v_k^\star\).
The \(K\) vertices are affinely independent, so their convex hull is a
\((K-1)\)-dimensional simplex containing all ratio points.
Recovering the barycentric coordinates gives \(A\diag(w)\)
through~\eqref{eq:oracle-simplex-main}. Column normalization removes
the positive factors \(w(k)\), recovering \(A\) up to a column permutation.
\end{proposition}

The server therefore estimates the pooled-marginal canonical direction
and chooses an orthonormal basis for its complement in the estimated signal
space. Different complementary bases rotate the ratio points and their
vertices together, leaving the barycentric coordinates unchanged.

After forming empirical ratio points \(\widehat r_j\), vertex hunting estimates the
simplex vertices \(v_1^\star,\ldots,v_K^\star\). We use augmented successive
projection by default, while any routine satisfying the stability condition stated in
Section~\ref{sec:theory} can also be used. Given the vertices, we then use the standard
Topic-SCORE barycentric-coordinate recovery described in
Appendix~\ref{subsec:vh-barycentric}, solving the  system in
\eqref{eq:barycentric-appendix} and then normalizing the result to obtain the topic loadings.

\subsection{Federated Topic-SCORE}
\label{subsec:algorithm-main}

We now turn the preceding geometric framework into the federated protocol in
Algorithm~\ref{alg:fed-topic-score-main}. Patient-level records remain local,
while sites share aggregate summaries.
Each site first uploads \(\bar d^{(m)}\in\mathbb R^p\). The server then uses these uploads to computes \(M_d\) and broadcasts it back to the sites.
Each site next uploads \(\widehat\Xi^{(m)}\in\mathbb R^{p\times K}\),
whose columns are the leading \(K\) left singular vectors of
\(M_d^{-1/2}D^{(m)}\). Finally, the server broadcasts
\(\widehat A_{\rm fed}\) for local membership weight estimation.
The entire protocol requires no repeated cross-site model-update rounds.

\begin{algorithm}[!htbp]
\caption{Federated Topic-SCORE to estimate latent representation components and memberships}
\label{alg:fed-topic-score-main}
\begin{algorithmic}[1]
\Require Local concept-by-patient matrices \(\{D^{(m)}\}_{m=1}^M\) (separately supplied by each site and staying local), number of latent representation components \(K\),
weights \(\{\alpha_m\}_{m=1}^M\), vertex-hunting routine
\Ensure Topic loading estimate \(\widehat A_{\rm fed}\) and local memberships
\(\{\widehat w_i^{(m)}\}\)
\Statex \textit{Marginal pre-normalization.}
\State Each site sends \(\bar d^{(m)}=n_m^{-1}D^{(m)}\1\) to the central server.
\State Server forms \(\bar d=n_{\rm tot}^{-1}\sum_m n_m\bar d^{(m)}\) and sets
\(M_d=\diag(\bar d)\).
\State Server broadcasts \(M_d\) to all sites.
\Statex \textit{Local spectral upload.}
\State Each site computes \(D_f^{(m)}=M_d^{-1/2}D^{(m)}\) and uploads
\(\widehat\Xi^{(m)}\in\mathbb R^{p\times K}\), whose columns are its leading
\(K\) left singular vectors.
\Statex \textit{Server-side simplex recovery.}
\State Form \(\bar P=\sum_m\alpha_m\widehat\Xi^{(m)}
\widehat\Xi^{(m)\top}\) and extract its top-\(K\) projector \(\widehat P\).
\State Set
\(\widehat\xi_1=\widehat P M_d^{1/2}\1/\|\widehat P M_d^{1/2}\1\|_2\).
\State Choose an orthonormal \(\widehat\Xi_\perp\) spanning
\(\operatorname{range}(\widehat P-\widehat\xi_1\widehat\xi_1^\top)\), and set
\(\widehat r_j(\ell)=\widehat\Xi_\perp(j,\ell)/\widehat\xi_1(j)\).
\State Set \(\widehat y_j=(1,\widehat r_j^\top)^\top\in\mathbb R^K\) and
use vertex hunting to obtain
\(\widehat v_1,\ldots,\widehat v_K\in\mathbb R^{K-1}\).
\State Form
\(\widehat Q=\begin{pmatrix}1&\cdots&1\\
\widehat v_1&\cdots&\widehat v_K\end{pmatrix}\).
For each concept \(j\), solve \(\widehat Q\widehat\pi_j^\star=\widehat y_j\)
and set \(\widehat\pi_j=\mathcal T(\widehat\pi_j^\star)\), with
\(\mathcal T\) defined after~\eqref{eq:wls-step-main}.
\State Set \(\widehat\Pi=[\widehat\pi_1,\ldots,\widehat\pi_p]^\top\) and
\(\widetilde A_{\rm fed}=M_d^{1/2}\diag(\widehat\xi_1)\widehat\Pi\).
\State Compute
\(\widehat A_{{\rm fed},\cdot k}
=\widetilde A_{{\rm fed},\cdot k}/
\|\widetilde A_{{\rm fed},\cdot k}\|_1\) for \(k=1,\ldots,K\).
\Statex \textit{Local membership estimation.}
\State Server broadcasts \(\widehat A_{\rm fed}\) to all site, and each site estimates
\(\widehat w_i^{(m)}\) for its own patients using~\eqref{eq:wls-step-main}.
\end{algorithmic}
\end{algorithm}

% CAMERA M-04: rebuttal 430; later normalized-subspace clarification.
As shown in Algorithm~\ref{alg:fed-topic-score-main}, sites use truncated SVD to form their local estimates of \(\mathcal S\). The server perform a weighted aggregation of the projection matrices based on the local estimates, denoted as \(\bar P\), using nonnegative weights \(\alpha_m\ge0\), where \(\sum_m\alpha_m=1\). A natural default choice for the weights is \(\alpha_m=n_m/n_{\rm tot}\), which assigns greater weight to larger sites. The projector \(\widehat P\) represents the orthogonal projector onto the span of eigenvectors associated
with the \(K\) largest eigenvalues of \(\bar P\).
Aggregating projectors removes dependence on rotations of the local orthonormal bases, since such rotations leave the corresponding projection matrices unchanged.
After topic recovery, each site estimates patient memberships by
unconstrained weighted least squares followed by truncation and
\(\ell_1\)-normalization:
\begin{equation}\label{eq:wls-step-main}
    \widehat w_i^{(m),\star}
    =\underset{b\in\mathbb R^K}{\arg\min}\,
    \left\|M_d^{-1/2}\{d_i^{(m)}-\widehat A_{\rm fed}b\}\right\|_2^2,
    \qquad
    \widehat w_i^{(m)}=\mathcal T(\widehat w_i^{(m),\star}),
\end{equation}
where \(\mathcal T(b)=[b]_+/\|[b]_+\|_1\), with
\(\mathcal T(b)=K^{-1}\1\) if \([b]_+=0\).

\begin{remark}[Working local rank]
\label{rem:working-rank}
Although our theory is developed in the setting with local upload of rank \(K\), sites may upload a modestly larger
rank \(L\ge K\) before the server extracts a rank-\(K\) aggregate. This can improve
finite-sample coverage under noisy or heterogeneous local spectra, but overly large
\(L\) can add noisy directions. We discuss more details of the local SVD rank in Appendix~\ref{app:practical-svd-rank}.
\end{remark}

\section{Theoretical guarantees}\label{sec:theory}

We study the theoretical guarantees for the estimation accuracy of Federated Topic-SCORE. Specifically, we derive finite-sample error bounds for estimation of the ratio coordinates and the shared topic loading matrix \(A\). These bounds quantify how estimation error depends on the vocabulary size, local sample size, record length, and number of participating sites.

We establish the topic-loading error bounds in three stages. First, projector averaging recovers the shared spectral space up to a federated variance term, a local second-order spectral bias, and a row-wise range term. Second, the pooled marginal identifies the canonical positive direction inside this recovered space. Third, row-wise error bounds for the resulting canonical basis yield bounds for the ratio coordinates and, through vertex hunting, for the topic loading matrix.

Note that the main upper bounds below assume equal site sizes \(n_m=n\), equal aggregation weights \(\alpha_m=1/M\), fixed \(K\), and \(N\ge2\). The weighted projector-averaging argument extends to unequal site sizes and general deterministic weights \(\{\alpha_m\}_{m=1}^M\). Constants \(C\) may vary between displays and depend only on \(K\), the regularity constants in Assumption~\ref{ass:regular-plsi}, and, where applicable, the stability constants in Assumption~\ref{ass:vh} below. Appendix~\ref{app:proofs} provides the full finite-sample statements, constant dependencies, normalization-bridge decomposition, and proofs, including the extension to settings with unequal site sizes.

For column-normalized topic loading matrices, we measure estimation error using the permutation-aligned \(\ell_1\) distance
\[
    L(\widehat A,A)
    =\min_{\kappa\in S_K}\sum_{k=1}^K
    \|\widehat A_{\cdot,\kappa(k)}-A_{\cdot,k}\|_1,
\]
where \(S_K\) denotes the set of permutations of \(\{1,\ldots,K\}\). To state the finite-sample bounds, fix a failure probability \(\delta\in(0,1)\) and let
\(\ell=\ell_{n,M}(\delta)=\log(2pM/\delta)\).
We distinguish three contributions to the error bounds,
\[
    T_{\rm var}
    =\beta_n\sqrt{\frac{p\ell}{NnM}},
    \qquad
    T_{\rm loc}
    =\beta_n^2\frac{p\ell}{Nn},
    \qquad
    T_{\rm row}
    =\frac{\sqrt p\,\ell}{M}.
\]
Here, \(\beta_n\ge1\) is a deterministic factor arising in the concentration bounds for normalized multinomial counts, with its definition given in Appendix~\ref{subsec:rates}. The term \(T_{\rm var}\) is the federated variance contribution, \(T_{\rm loc}\) is the local second-order spectral bias that is not averaged away across sites, and \(T_{\rm row}\) arises from bounding the averaged projector error uniformly across rows.

We combine these contributions with the remaining error terms by defining
\begin{equation}\label{eq:main-radius}
    \mathcal R_{n,M}(\delta)
    =T_{\rm var}+T_{\rm loc}+T_{\rm row}
    +\mathrm{Rem}_{n,M,N,p}(\delta).
\end{equation}
 
 The remainder accounts for empirical marginal normalization, the diagonal bias of ordinary SVD, and higher-order terms in the projector perturbation expansion. The complete expression for \(\mathcal R_{n,M}(\delta)\) is given in Appendix~\ref{subsec:rates}.

The theorem below uses the deterministic smallness conditions stated in Theorem~\ref{thm:fed-rowwise} in Appendix~\ref{subsec:rates}. These conditions keep the perturbations sufficiently small for the subspace and ratio constructions to remain stable. Let \(e_j\in\mathbb R^p\) denote the \(j\)th standard basis vector. The following theorem bounds the row-wise errors in the canonical basis and the associated ratio coordinates.

\begin{theorem}[Row-normalized spectral coordinate recovery]\label{thm:row-main}
Suppose Assumption~\ref{ass:regular-plsi} holds, \(n_m=n\), \(\alpha_m=1/M\), and the
deterministic smallness conditions in Theorem~\ref{thm:fed-rowwise} are satisfied.
Then, with probability at least \(1-\delta\),
there exists an orthogonal matrix \(O_\perp\in\mathbb R^{(K-1)\times(K-1)}\) such that  
\[
\max_j h_j^{-1/2}
\left\|e_j^\top\{[\widehat\xi_1,\widehat\Xi_\perp O_\perp]-[\xi_1,\Xi_\perp]\}\right\|_2
+
\max_j\|O_\perp^\top\widehat r_j-r_j\|_2
\le C\mathcal R_{n,M}(\delta).
\]
\end{theorem}

 While Theorem~\ref{thm:row-main} above controls the estimation error in the ratio coordinates, recovering the topic-loading matrix also requires the vertex-hunting step to be stable under small uniform perturbations of these coordinates. The following assumption formalizes this stability requirement.

\begin{assumption}[Stable vertex hunting]\label{ass:vh}
The vertex-hunting routine is stable under uniform ratio perturbations: if
\(\max_j\|O^\top\widehat r_j-r_j\|_2\le\varepsilon\) for some orthogonal
\(O\in\mathbb R^{(K-1)\times(K-1)}\) and sufficiently small \(\varepsilon\), then,
up to a permutation, \(\max_k\|O^\top\widehat v_k-v_k^\star\|_2\le C\varepsilon\).
\end{assumption}

Note that Proposition~\ref{prop:sp-vh} in Appendix~\ref{app:proofs}
shows that the augmented successive-projection routine
satisfies Assumption~\ref{ass:vh} under the anchor-concept
and nondegenerate-simplex conditions.

Combining Theorem~\ref{thm:row-main} with Assumption~\ref{ass:vh} yields the following topic-loading recovery guarantee.

\begin{theorem}[Topic loading recovery]\label{thm:A-main}
Suppose Assumptions~\ref{ass:regular-plsi} and~\ref{ass:vh} hold, \(n_m=n\),
\(\alpha_m=1/M\), and the deterministic smallness conditions in
Theorem~\ref{thm:fed-rowwise} are satisfied, with the additional
vertex-hunting restriction specified in Theorem~\ref{thm:fed-A}.
Then, with probability at least \(1-\delta\), up to a common permutation
of latent representation components,
\[
    \max_j \frac{\|\widehat a_j-a_j\|_1}{h_j}
    +L(\widehat A_{\rm fed},A)
    \le C\mathcal R_{n,M}(\delta).
\]
Here \(\widehat a_j^\top\) is the \(j\)th row of
\(\widehat A_{\rm fed}\) after the common column permutation.
\end{theorem}

The following theorem establishes a minimax lower bound
of order \(\sqrt{p/(NnM)}\) for estimating the topic-loading
matrix over the stated parameter class.

\begin{theorem}[Minimax lower bound for estimating the topic-loading matrix]
\label{thm:lower-main}
Fix \(K\ge2\). There exist constants
\(c_\star(K)\in(0,1/(2K)]\) and \(p_0(K)\) such that the
following holds. For any fixed \(c_{\mathrm{reg}}\in(0,c_\star(K)]\),
let the supremum below range over all pLSI parameters
\((A,W^{(1:M)})\) satisfying
Assumption~\ref{ass:regular-plsi} with \(c=c_{\mathrm{reg}}\).
Suppose \(n_m=n\ge K\), \(p\ge p_0(K)\), and
$\log(nM)\le\min\{p,N\}$, $p\log(nM)=o(NnM)$. Then there exist constants \(c,c'>0\), depending only on
\(K\) and \(c_{\mathrm{reg}}\), such that, along any sequence
satisfying these conditions and for all sufficiently large
problem sizes,
\[
\inf_{\widehat A}
\sup_{(A,W^{(1:M)})}
\mathbb P_{A,W^{(1:M)}}\!\left\{
L(\widehat A,A)\ge c\sqrt{\frac{p}{NnM}}
\right\}\ge c'.
\]
\end{theorem}

The lower bound has the same dependence on \(p,N,n,M\) as the
federated variance component \(T_{\rm var}\), up to logarithmic
and \(\beta_n\) factors. The complete finite-sample upper
guarantee is given by~\eqref{eq:main-radius}.

The error bound in~\eqref{eq:main-radius} also clarifies
how the local sample size \(n\), record length \(N\),
and number of participating sites \(M\) affect estimation
accuracy. Ignoring logarithmic and \(\beta_n\) factors,
increasing \(n\) or \(N\) reduces both \(T_{\rm var}\)
and \(T_{\rm loc}\), while increasing \(M\) reduces
\(T_{\rm var}\) and \(T_{\rm row}\). The local spectral
bias \(T_{\rm loc}\) arises from the site-level SVDs
computed before aggregation. Averaging more site summaries
does not remove this bias. Consequently, at fixed \(n\)
and \(N\), local bias can eventually dominate the error
bound as \(M\) increases. In our analysis, \(T_{\rm row}\) arises when bounding
the errors in the estimated ratio coordinates for
all concepts simultaneously.

\section{Numerical experiments and real-world EHR application}
\label{sec:experiments}

\subsection{Simulations}\label{sec:simulations}

We conduct simulation studies to compare Federated Topic-SCORE with pooled Topic-SCORE, single-site Topic-SCORE, and a
federated latent Dirichlet allocation (LDA) benchmark, using the permutation-aligned $\ell_1$ distance from the simulated ground-truth topic matrix $A$ as the evaluation metric. The Federated LDA benchmark is computed based on a recent federated
LDA topic modeling method~\citep{kushwaha2025federated}. This method avoids repeated cross-site synchronization, but its site-level LDA fits are
iterative and computationally heavier than Federated Topic-SCORE. We begin with a base configuration that serves as the base setting for all subsequent simulation variations and sensitivity analyses, where $M = 5, n_m = 200$ for all sites, $p = 1000, N = 2000$, and $K = 5$. In our simulations, we consider both homogeneous and heterogeneous topic membership weight distributions across sites. Appendix~\ref{app:sim-methods}
provides details of the full data generating processes, and Appendix~\ref{app:code-data} provides a link to the code repository. 

\begin{figure}[!htbp]
\centering
\includegraphics[width=\linewidth]{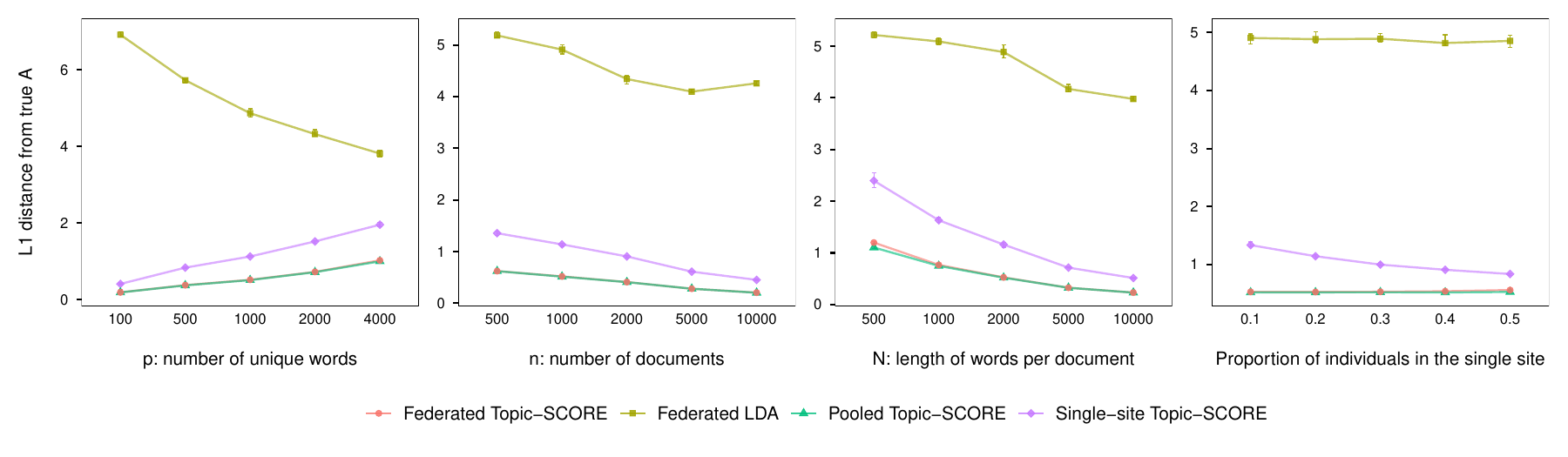}
\caption{Estimation accuracy of the topic loading matrix \(A\) under homogeneous weight distributions across sites. From left to right, panels vary vocabulary size \(p\), total sample size \(n\), count length \(N\), and the fraction of patients at a single site (with the remaining patients distributed equally among the other sites). Points show the median of the $\ell_1$ distance across 30 simulation replicates, and error bars show the interquartile ranges. Distinct point shapes and colors identify the methods. The Federated Topic-SCORE and pooled Topic-SCORE curves nearly overlap across the displayed settings.}
\label{fig:sim-homo}
\end{figure}

\paragraph{Federated and pooled Topic-SCORE achieve similar recovery performance when membership weight distributions are homogeneous across sites.} We performed a series of sensitivity analyses by varying one data-generating parameter at a time while holding all others fixed at their base-case values. These results are consistent with the leading terms in $R_{n,M}$ in that increasing $p$ raises the error, whereas increasing $n$ or $N$ reduces the variance and local bias contributions. Across these settings, the federated estimator
nearly overlaps with pooled Topic-SCORE and remains more accurate than single-site
estimation, including under site-size imbalance (Figure~\ref{fig:sim-homo}).
Federated LDA has substantially larger error throughout the homogeneous settings,
consistent with the original Topic-SCORE evidence that LDA-type estimators can be
less accurate for topic-loading recovery~\citep{Ke02012024}.
Additional sensitivity analyses for settings where membership weight distributions are homogeneous across sites are provided in Appendix~\ref{app:sim-extra}.

\FloatBarrier

\begin{figure}[!htbp]
\centering
\includegraphics[width=0.8\linewidth]{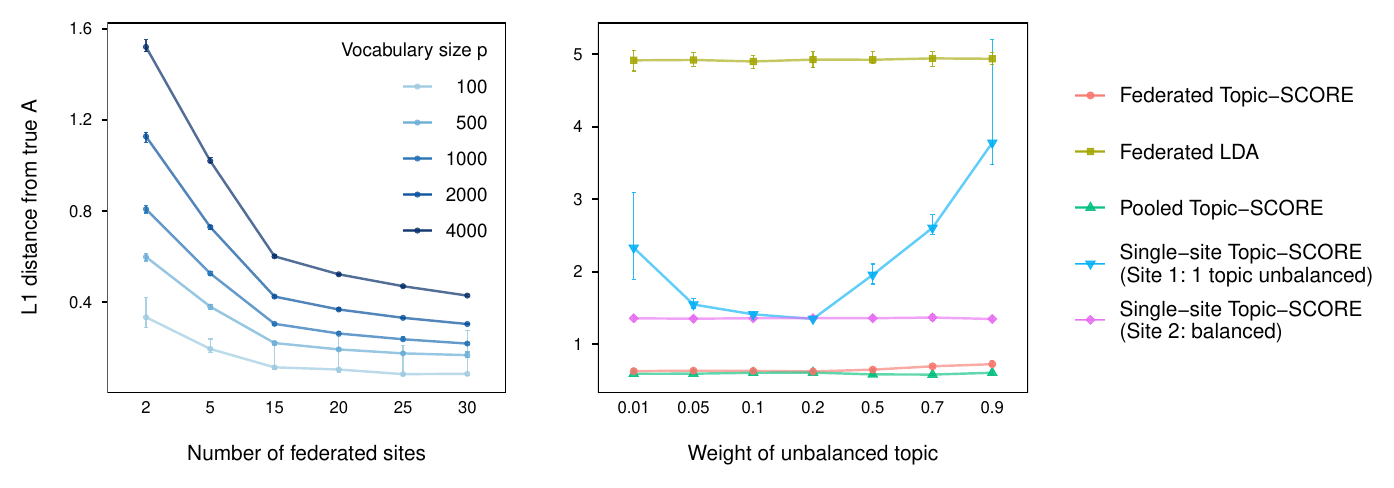}
\caption{Estimation accuracy of \(A\) under varying numbers of participating sites and heterogeneous membership weight distributions across sites. Left: Federated Topic-SCORE error versus the number of participating sites, with per-site sample size fixed as in the base setting. Colors denote vocabulary size \(p\). Right: comparison of methods as the membership weight of one topic for all patients at Site 1 varies, while all other sites remain balanced. Distinct point shapes and colors identify the methods and different single sites. Points show median $\ell_1$ distance across 30 simulation replicates, and error bars show interquartile ranges.}
\label{fig:sim-sites-het}
\end{figure}

\paragraph{Federated Topic-SCORE achieves better recovery as more sites participate.} As more sites contribute summary statistics, the estimation error of Federated Topic-SCORE decreases across vocabulary sizes (Figure~\ref{fig:sim-sites-het}, left). This is consistent with the error bound in Theorem~\ref{thm:A-main}, whose federated variance and row-wise terms decrease with \(M\) when the per-site sample size is fixed.

\paragraph{Federated Topic-SCORE performs well under heterogeneous membership weight distributions across sites.}
When membership weights at one site become increasingly concentrated on a single topic while the other sites remain balanced, the single-site estimator fitted to that site becomes less accurate. Such concentration can make \(\Sigma_{W,m}\) nearly singular, undermining the local richness condition in Assumption~\ref{ass:regular-plsi}. Despite the undermined local topic richness, Federated Topic-SCORE remains close to pooled Topic-SCORE in our experiment (Figure~\ref{fig:sim-sites-het}, right), suggesting that incorporating summary statistics from the other balanced sites can mitigate the effect of topic imbalance at one site. Federated LDA is relatively stable as the site becomes more imbalanced, but its estimation error remains substantially higher than that of Federated Topic-SCORE. Additional simulations with more severe topic imbalance at one or multiple sites are described in Appendix~\ref{app:sim-heterogeneous}.

\FloatBarrier

\subsection{Real-world EHR application: learning stroke patient representations}
\label{sec:realdata}

We applied Federated Topic-SCORE to real-world EHR data of stroke patients from four hospitals in the Mass General Brigham healthcare system, including Massachusetts General Hospital (MGH), Newton--Wellesley Hospital (NWH), Brigham and Women's Hospital (BWH), and Spaulding Rehabilitation Hospital (SRH), with the goal of learning latent representation components of stroke subtypes and characterizing patient subgroups using the representations. The final cohort included 7,057 patients. For each patient, we set their first qualifying acute ischemic stroke admission date from January 1, 2005 through January 31, 2025 as the index date. At each hospital, we constructed a concept-by-patient count matrix using diagnosis codes recorded during the first 30 days after the index date. Patients were followed for 1-year survival, with death as the event and censoring at the last available encounter or the end of follow-up. We used \(K=5\) for interpretability and local working rank \(L=10\), as discussed in Appendix~\ref{app:practical-svd-rank}. The study was approved by the Mass General Brigham Institutional Review Board. Appendix~\ref{app:real-details} provides details about the cohort description and concept construction.

\begin{figure}[!htbp]
\centering
\includegraphics[width=0.8\linewidth]{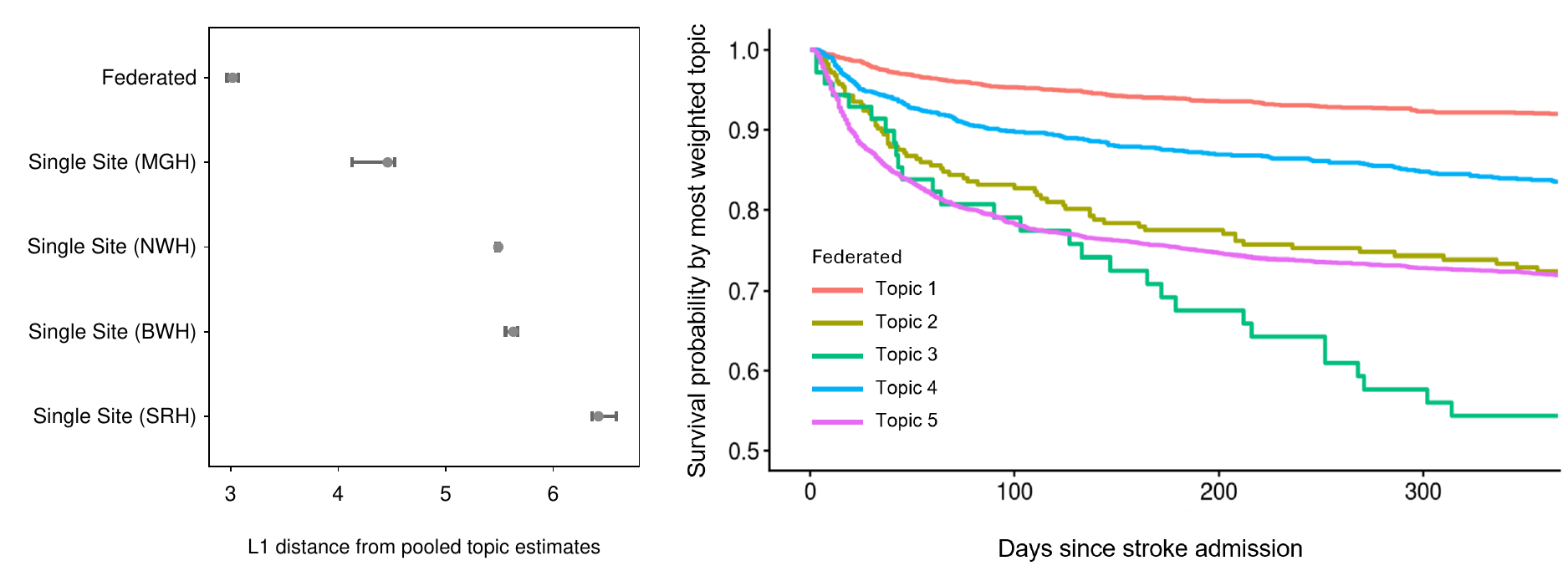}
\caption{Federated Topic-SCORE estimation and subgroup survival patterns in the stroke EHR cohort. Left: \(\ell_1\) distance from pooled Topic-SCORE for federated estimation and single-site estimation at Massachusetts General Hospital (MGH), Newton--Wellesley Hospital (NWH), Brigham and Women's Hospital (BWH), and Spaulding Rehabilitation Hospital (SRH). Points and error bars show medians and interquartile ranges across 20 replicates for the vertex hunting procedure. Right: 1-year survival curves for patient subgroups defined by their largest estimated topic membership weights.}
\label{fig:real-benchmark-survival}
\end{figure}

\paragraph{Federated Topic-SCORE estimates are closer to the pooled benchmark than single-site estimates.}
Because the ground-truth topic loading matrix is unknown for real EHR data, we use pooled Topic-SCORE as a centralized reference to understand the performance of our method. We compare federated and single-site estimates with the pooled estimate using the permulation-aligned \(\ell_1\) distance. Federated Topic-SCORE has the smallest distance to the pooled estimate, whereas single-site estimates differ across hospitals and are farther from the reference (Figure~\ref{fig:real-benchmark-survival}, left). These differences may reflect variation in local case mix and coding practices.

\begin{figure}[!htbp]
\centering
\includegraphics[width=\linewidth]{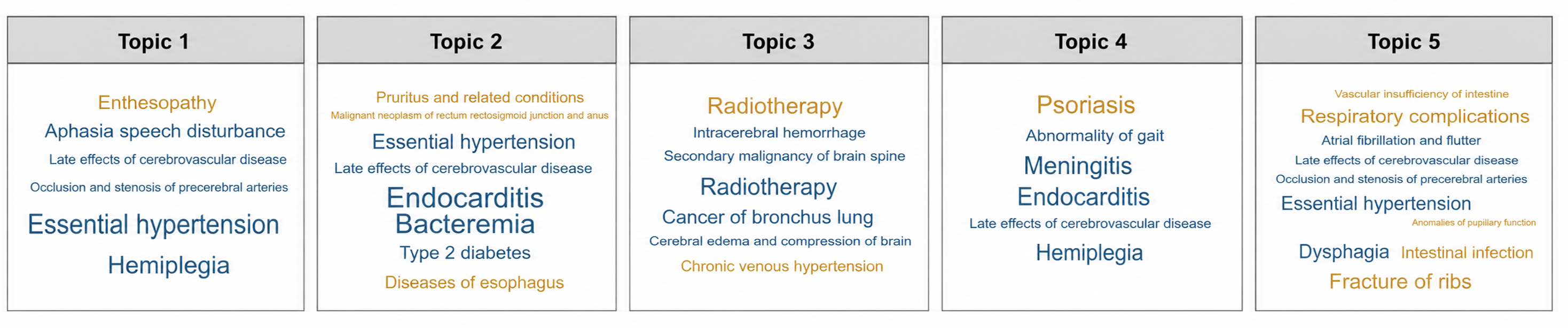}
\caption{Topics by Federated Topic-SCORE learned from EHR concepts for stroke patients. Each panel summarizes one latent representation component (topic) using diagnosis-code-based concepts. Blue (darker) terms denote the highest-loading concepts, whereas gold (lighter) terms denote anchor concepts.}
\label{fig:real-representations}
\end{figure}

\paragraph{Learned representation components highlight clinically distinct patient subgroups.} The latent components learned by Federated Topic-SCORE yield clinically interpretable post-stroke EHR concept profiles, as shown by the highest-loading and anchor concepts in Figure~\ref{fig:real-representations}. We then assigned each patient to the subgroup corresponding to their largest membership weight. Each EHR concept profile is clinically consistent with the corresponding patient subgroup's 1-year survival pattern (Figure~\ref{fig:real-benchmark-survival}, right). The subgroup characterized by cancer-care and hemorrhagic-event concepts shows the sharpest early survival decline, while the remaining profiles form a graded survival structure over follow-up. The highest-survival subgroup is dominated by vascular-sequelae and neurological-symptom concepts such as hypertension, aphasia, and hemiplegia, whereas lower-survival groups involve more medically complex multi-system profiles, including infections, cardiac and respiratory complications, vascular insufficiency, and fracture. These profiles should be read as EHR-derived phenotype summaries and require downstream clinical validation, but they may provide hypotheses for more personalized stroke and post-stroke management.

\FloatBarrier

\section{Discussion, limitations, and broader impact}
\label{sec:discussion}

Federated Topic-SCORE provides a non-iterative approach to learn shared latent patient
representation components from multi-institutional EHR data. Each site communicates a
marginal concept-frequency vector and one low-rank spectral summary, allowing the
server to recover latent component structure close to pooled Topic-SCORE without
patient-level data sharing. Our simulations show that Federated Topic-SCORE has lower estimation error with larger local sample sizes, longer records, and more participating sites, but higher error with larger vocabularies. These trends are consistent with the bounds that we derived on topic loading estimation error. The method performs well when membership weight distributions differ across sites, although severe topic imbalance at some of the sites can reduce estimation accuracy. The method also substantially outperforms the existing federated LDA benchmark, which is a useful non-iterative cross-site comparator but relies on computationally intensive local LDA fits and has weaker topic-loading recovery,
consistent with prior Topic-SCORE comparisons to LDA-type estimators \citep{Ke02012024}.  

In our application to real-world EHR data from stroke patients, the Federated Topic-SCORE estimate was closer to the pooled reference than any single-site estimate. The model then represented each patient by membership weights across topics learned without using any specific outcome labels. These weights form representations for discovery of patient subgroups potentially informative for multiple outcomes. In this study, we assigned each patient to the topic with the largest weight as a simple way to define subgroups for interpretation and comparison and obtained subgroups with distinct survival patterns. In general, patient subgroups can also be defined using the full membership weight vectors, corresponding to each patient's full representation.

Specifying the number of representation components rank \(K\) controls representation granularity. Smaller \(K\) yields broader and
often more stable profiles, whereas larger \(K\) may reveal finer subgroups at the cost
of interpretability and finite-sample stability. In the EHR application, we use
\(K=5\) for interpretability and local working rank \(L=10\) to improve finite-sample
subspace coverage before extracting the global rank-\(K\) component. Technical discussion of a pilot-stage rule for selecting \(K\) and \(L\) is deferred to Appendix~\ref{app:proof-WK} and
Appendix~\ref{app:practical-svd-rank}, respectively. In practice, representation
interpretability is also an important consideration when choosing both ranks.

We emphasize that Federated Topic-SCORE can be applied to context-aware EHR concepts based on large language models (LLMs). Recent work on patient-centered topic modeling with LLMs suggests that context-aware models can help construct clinically meaningful concepts or topic-like summaries from unstructured EHR data such as clinical notes. Such LLM-derived embeddings could be created locally and mapped to a shared vocabulary or count data format as described in the existing study~\citep{austern2025pptm}, after which our method can be applied directly. This application in a multi-institutional setting for unstructured data is an important future step for federated representation learning.

The shared topic loading assumption has the advantage of directly modeling a general population in which shared disease- or biology-related profiles can be studied and expressed, and site-specific $W^{(m)}$ can capture differences in patient case mix and heterogeneity across sites. While the varying data collection approaches across sites could imply different concept meanings, this issue can be mitigated by data harmonization preprocessing or a common data model that aligns concept semantics before modeling ~\citep{hripcsak2019facilitating}.

Several limitations should be considered when interpreting the findings of our study. First, the theory assumes a
shared-topic pLSI model based on a harmonized EHR concept vocabulary, anchor concept structure, rare-concept preprocessing, and sufficient local
representation richness. These conditions are important for identifiability and local
spectral recovery. They can be partly assessed through vocabulary checks and anchor
diagnostics, and the robustness of our method under near-violations of some assumptions is discussed in the appendix. Second, the federated LDA is a practical topic modeling benchmark for the same non-iterative cross-site communication regime, but we did not provide an exhaustive comparison with iterative federated topic
models as they fall into a different communication scheme. Third, the real-world analysis uses data from one multi-facility healthcare network, so its clinical findings may lack generalizability
to other health systems and patient
populations. We therefore view the application as a feasibility and clinical interpretability demonstration rather than external validation. Fourth, the learned
representations and associated survival differences are hypothesis-generating. The
survival analysis is associative rather than causal or predictive, and the
representations require clinical review and external validation before being used to
guide care. Finally, summary sharing reduces patient-level raw-record exposure but
does not provide formal differential privacy. Marginal and spectral summaries may
still reveal aggregate site-level information, especially for rare concepts or
specialized facilities. The current setting is designed for addressing privacy concerns with respect to institutional data-governance restrictions and can be combined with differential privacy approaches as a future direction.

By enabling non-iterative multi-site representation learning without patient-level data
pooling, this work may make collaborative patient profiling across broader and more
generalizable patient populations practically feasible. The non-iterative
summary-sharing design reduces repeated coordination burden, which is often a major
barrier to sustained clinical data collaboration. The resulting interpretable EHR concept
representations can support hypothesis generation for risk stratification,
care-pathway evaluation, and more personalized care.

\pagebreak

\begin{ack}
We thank the reviewers and the area chair for their constructive feedback, which helped improve this work. Rui Duan acknowledges support for this work from the National Institutes of Health (R01 GM148494) and the Patient-Centered Outcomes Research Institute (PCORI) Project Program Award ME-2024C1-37351. Zhiyu Yan is supported by the American Heart Association (26GWTGDRASp1611353 and 26PRE1563536). Christopher D. Anderson is supported by the National Institutes of Health (U01NS069673 and RF1NS139183) and the Global Brain Care Coalition. Christopher D. Anderson reports sponsored research support from Bayer AG and consulting for Apopharma and MPM BioImpact, unrelated to the present work.
\end{ack}

\bibliographystyle{plainnat}
\bibliography{ref_camera_v2}

\newpage
\appendix

\section{Proofs and technical details}\label{app:proofs}

% Local proof-appendix helpers.  These are harmless if the macros/environments
% are already supplied by the main preamble.
\providecommand{\Var}{\operatorname{Var}}
\providecommand{\Cov}{\operatorname{Cov}}
\makeatletter
\@ifundefined{lemma}{\newtheorem{lemma}{Lemma}}{}
\makeatother

% Self-contained proof appendix adapted for the EHR paper.
% word/document/topic corresponds to code/patient/profile.
\subsection{Notation map and proof-appendix scope}\label{app:proof-scope}
This appendix gives a self-contained proof of the theoretical results for the
federated Topic-SCORE estimator used in the paper.  We use the standard
pLSI/topic-model terminology in the proof: words are EHR concepts or codes,
documents are patient records, and topics are shared representation components.
The topic matrix \(A\) is the shared topic-loading matrix in the main text:
its columns describe the shared components. Each column \(w_i\) of the
topic-weight matrix \(W\) is a patient representation.
Topic-SCORE refers to the SCORE-style spectral recovery chain: estimate a low-rank
spectral space, choose a positive first direction, divide the remaining coordinates
rowwise by that direction to form ratio coordinates, and use simplex geometry to
recover topic loadings.  Whitening in this appendix means marginal-frequency
pre-normalization by \(M_0^{-1/2}\) or \(M_d^{-1/2}\), rather than covariance
whitening.  All probabilities and expectations are conditional on the fixed
membership matrices \(W^{(1:M)}\), exactly as in the fixed-design Topic-SCORE
analysis.  The appendix restates the model, regularity
assumptions, oracle simplex geometry, projector-averaging bounds,
empirical-pre-normalization bridge, row-wise canonical-basis recovery,
deterministic simplex propagation, the minimax lower bound, the held-out
membership-weight guarantee, and the optional overspecified-rank calculation.
The proof-level radius \(\varepsilon_{n,M}(\delta)\) defined in
\eqref{eq:epsilon-nM} is the same quantity denoted \(\mathcal R_{n,M}(\delta)\)
in the main text; the notation \(\varepsilon\) is used in the appendix only to keep
formulas inherited from the spectral perturbation proof compact.

\subsection{Model and the federated setting}\label{sec:model}

\subsubsection{pLSI model}

The vocabulary size is $p$ and the number of topics is $K\ll \min\{p,n_{\mathrm{tot}}\}$. All documents have the common length $N$. Throughout the upper-bound theory we assume $N\ge2$, so the signal factor $1-1/N$ is bounded away from zero. The empirical word-frequency vector of document $i$ is
\[
d_i(j)=\frac{\#\{\text{occurrences of word }j\text{ in document }i\}}{N},
\qquad j=1,\dots,p.
\]
Let $D=[d_1,\dots,d_{n_{\mathrm{tot}}}]\in\R^{p\times n_{\mathrm{tot}}}$ be the corpus matrix.

\begin{definition}[Topic matrix and topic weights]
The topic matrix is $A=[A_1,\dots,A_K]\in\R^{p\times K}$, where each column is a probability mass function on the vocabulary: $A_k\ge 0$ and $\1^\top A_k=1$. Document $i$ has topic weights $w_i\in\R^K$ with $w_i\ge 0$ and $\1^\top w_i=1$. Write
\[
W=[w_1,\dots,w_{n_{\mathrm{tot}}}]\in\R^{K\times n_{\mathrm{tot}}}.
\]
\end{definition}

\begin{assumption}[pLSI generation mechanism]
For each document $i$,
\[
N d_i\sim \mathrm{Multinomial}(N,Aw_i),
\]
and documents are independent conditional on $W$. Hence the mean matrix is $D_0:=\E[D\mid W]=AW$.
\end{assumption}

\subsubsection{Federated partition and pooled notation}

There are $M$ clients. Client $m$ owns $n_m$ documents and forms
\[
D^{(m)}=[d_1^{(m)},\dots,d_{n_m}^{(m)}]\in\R^{p\times n_m},
\qquad
W^{(m)}=[w_1^{(m)},\dots,w_{n_m}^{(m)}]\in\R^{K\times n_m}.
\]
We work in the shared-topic regime: the topic matrix $A$ is common across clients, while the client-level topic-weight matrices $W^{(m)}$ may differ across sites. Define
\[
n_{\mathrm{tot}}:=\sum_{m=1}^M n_m,
\qquad
D=[D^{(1)},\dots,D^{(M)}],
\qquad
W=[W^{(1)},\dots,W^{(M)}].
\]
For any corpus matrix $X\in\R^{p\times n}$, let
\[
\bar x:=\frac1nX\1,
\qquad M_X:=\diag(\bar x).
\]
The pooled population and empirical marginals are
\[
m_0:=\frac{1}{n_{\mathrm{tot}}}D_0\1,
\qquad M_0:=\diag(m_0),
\qquad
\bar d:=\frac{1}{n_{\mathrm{tot}}}D\1,
\qquad M_d:=\diag(\bar d).
\]
For each client,
\[
\bar d^{(m)}:=\frac1{n_m}D^{(m)}\1,
\qquad
M_d^{(m)}:=\diag(\bar d^{(m)}),
\]
and let the corresponding local population marginal be
\[
m_{0,m}:=\frac1{n_m}D_0^{(m)}\1,
\qquad
M_{0,m}:=\diag(m_{0,m}).
\]
Throughout, $A$ and $W^{(1:M)}$ are treated as fixed unknown matrices satisfying deterministic regularity conditions. All probabilities and expectations are conditional on $W^{(1:M)}$. This is the same fixed-design convention as in the single-site Topic-SCORE analysis of Ke and Wang's SVD Topic-SCORE paper.

For the topic weights, define the pooled second moment and the client-level second moments
\[
\Sigma_W:=\frac1{n_{\mathrm{tot}}}WW^\top,
\qquad
\Sigma_{W,m}:=\frac1{n_m}W^{(m)}W^{(m)\top},
\qquad 1\le m\le M.
\]
When $\lambda_{\min}(\Sigma_{W,m})>0$, the columns of
\[
M_0^{-1/2}D_0^{(m)}=M_0^{-1/2}AW^{(m)}
\]
span the common space
\[
\mathcal S:=\col(M_0^{-1/2}A).
\]
This $K$-dimensional space, rather than any client-specific spectral basis, is the deterministic target of the federated spectral procedure.

% CAMERA_V1_BEGIN P-A03
\begin{remark}[Shared loading matrices and site heterogeneity]\label{rem:shared-A-heterogeneity}
The shared-$A$ assumption does not require the participating sites to have identical patient populations. The site-specific membership matrices $W^{(m)}$ may vary arbitrarily and therefore capture differences in case mix, topic prevalence, and patient-level mixtures.

More generally, if a site-specific loading matrix satisfies $A_m = A B_m$ for some column-stochastic matrix $B_m$, then
\[
A_m W^{(m)} = A\bigl(B_m W^{(m)}\bigr).
\]
Thus, heterogeneity lying within the convex hull of the shared topic dictionary can be absorbed exactly into the membership weights.

What the current model does not capture is topic-conditional coding variation outside this shared span, such as when the same clinical phenotype induces systematically different code distributions across sites. Allowing site-specific loading matrices $A_m$ would address this issue, but would also require additional topic-alignment constraints to preserve cross-site comparability. We will clarify this distinction and identify a shared-plus-local model,
\[
A_m = A + \Delta_m,
\]
as an important extension for applications involving more heterogeneous healthcare systems.
\end{remark}
% CAMERA_V1_END P-A03

% =========================================================
\subsection{Topic-SCORE preliminaries}

\subsubsection{Anchor words and regularity assumptions}

Let $a_j^\top$ denote the $j$th row of $A$ and define
\[
h_j:=\|a_j\|_1=\sum_{k=1}^K A(j,k),
\qquad
h_{\min}:=\min_{1\le j\le p}h_j,
\qquad
\bar h:=\frac1p\sum_{j=1}^p h_j=\frac Kp.
\]
Also define $H:=\diag(h_1,\dots,h_p)$ and $\Sigma_A:=A^\top H^{-1}A$.

\begin{definition}[Anchor word]
Word $j$ is an anchor word if row $j$ of $A$ has exactly one nonzero entry. If that nonzero entry is in column $k$, then word $j$ is an anchor word for topic $k$.
\end{definition}

\begin{assumption}[At least one anchor word per topic]\label{assump:anchor}
For each $k\in[K]$, there exists at least one anchor word for topic $k$.
\end{assumption}

\begin{assumption}[No extremely rare words]\label{assump:hmin}
There exists $c_1\in(0,1)$ such that
\[
h_{\min}\ge c_1\bar h=c_1\frac Kp.
\]
\end{assumption}

\begin{assumption}[Pooled weight regularity and topic overlap]\label{assump:Sigma}
There exists $c_2\in(0,1)$ such that
\[
\lambda_{\min}(\Sigma_W)\ge c_2,
\qquad
\lambda_{\min}(\Sigma_A)\ge c_2.
\]
\end{assumption}

\begin{remark}[Why the federated upper bounds use a weaker topic-overlap condition]\label{rem:SigmaA-weaker-than-single-site}
Ke and Wang's single-site SVD Topic-SCORE analysis assumes the stronger condition
$\min_{k,\ell}\Sigma_A(k,\ell)\ge c_2$ together with $\lambda_{\min}(\Sigma_W)\ge c_2$ and
$\lambda_{\min}(\Sigma_A)\ge c_2$; compare their condition~(10). In that paper the extra entrywise-overlap lower bound is tied to the Perron-based choice of the first singular direction. In the present federated analysis, the first SCORE direction is recovered canonically from the projected pooled marginal,
\[
\xi_1=M_0^{1/2}\1,
\]
so the oracle/simplex part uses only the positivity of $w=n_{\mathrm{tot}}^{-1}W\1$ and the invertibility of
$A^\top M_0^{-1}A$. Accordingly, the federated upper-bound chain requires only $\lambda_{\min}(\Sigma_W)\ge c_2$ and $\lambda_{\min}(\Sigma_A)\ge c_2$. The lower-bound construction below is carried out on a convenient subclass that in fact also has strictly positive off-diagonal overlap, but that stronger property is no longer part of the theorem interface.
\end{remark}

\begin{assumption}[Client-level weight richness]\label{assump:local-richness}
There exists $c_{\mathrm{loc}}\in(0,1)$ such that
\[
\lambda_{\min}(\Sigma_{W,m})\ge c_{\mathrm{loc}},
\qquad 1\le m\le M.
\]
\end{assumption}

\begin{table}[t]
\centering
\caption{How the assumptions are used.}
\small
\begin{tabular}{p{0.22\linewidth}p{0.31\linewidth}p{0.36\linewidth}}
\toprule
Assumption & Single-site role & Federated role \\
\midrule
Anchor words & Guarantees the ideal simplex and justifies vertex hunting. & Same role; used only after the ratio cloud has been constructed. \\
$h_{\min}\gtrsim p^{-1}$ & Gives $M_0(j,j)\asymp h_j$ and the row-wise scale. & Same role; also turns pooled marginal concentration into the $h_j^{-1/2}$ row scale. \\
Pooled $\Sigma_W$ and $\lambda_{\min}(\Sigma_A)$ & Enter oracle geometry, row scales, and the recovery identity. & Enter the canonical positive direction $\xi_1=M_0^{1/2}\1$, bounded simplex geometry, and the lower bound. \\
Client-level $\Sigma_{W,m}$ & Not present. & Controls local signal gaps and local projector perturbation. \\
\bottomrule
\end{tabular}
\end{table}

For two topic matrices with columns normalized in $\ell_1$, we measure error by the permutation-invariant loss
\begin{equation}\label{eq:L-loss-def}
L(\hat A,A):=\min_{\kappa\in S_K}\sum_{k=1}^K\|\hat A_{\kappa(k)}-A_k\|_1.
\end{equation}
All row-wise statements below are to be read after the same topic-label permutation.

\begin{remark}[Scope and relation to existing theory]\label{rem:scope-related}
The theoretical contribution is the federated interface needed for a Topic-SCORE guarantee. The single-site ratio, simplex, and recovery algebra follows the SVD Topic-SCORE chain of Ke and Wang's SVD Topic-SCORE paper; the new ingredients are the federated projector-averaging step, the empirical-whitening bridge, and recovery of the canonical pooled-marginal direction. The mathematical statements below are written so that this proof chain can be checked independently of implementation details and of the empirical study.
\end{remark}

\subsubsection{The pooled marginal induces the canonical positive direction}\label{subsec:canonical-direction}

The common signal space is
\[
\mathcal S:=\col(M_0^{-1/2}A).
\]
Let
\[
w:=\frac1{n_{\mathrm{tot}}}W\1\in\R^K.
\]
Since $D_0\1=n_{\mathrm{tot}}m_0$, we have
\[
M_0^{-1/2}Aw=M_0^{-1/2}m_0=M_0^{1/2}\1.
\]
Because $\1^\top m_0=1$, the vector
\[
\xi_1:=M_0^{1/2}\1
\]
already has Euclidean norm one. It is strictly positive, belongs to $\mathcal S$, and is therefore the natural pooled analogue of the positive SCORE denominator used in single-site Topic-SCORE.

Choose any orthonormal basis $\Xi_\perp\in\R^{p\times(K-1)}$ of $\mathcal S\cap \xi_1^\perp$ and write
\[
\Xi=[\xi_1,\Xi_\perp].
\]
Then there exists a unique nonsingular matrix $V=[V_1,\dots,V_K]\in\R^{K\times K}$ such that
\[
\Xi=M_0^{-1/2}AV,
\qquad
V_1=w.
\]
Here $V_\ell$ denotes the $\ell$th column of $V$, whereas $V_{k\ell}$ denotes the $(k,\ell)$ entry.
Thus the barycentric weights and simplex geometry are built from the pooled topic-weight mean, not from an arbitrary orthonormal basis of the common projector.

Define the oracle ratio matrix
\[
R(j,\ell)=\frac{\Xi_\perp(j,\ell)}{\xi_1(j)},
\qquad 1\le j\le p,
\quad 1\le \ell\le K-1.
\]
For $k\in[K]$, define
\[
v_k^\star(\ell):=\frac{V_{k,\ell+1}}{V_{k1}},
\qquad
1\le \ell\le K-1,
\]
and for each word $j$ define
\[
\pi_j(k):=\frac{A(j,k)V_{k1}}{\sum_{\ell=1}^K A(j,\ell)V_{\ell 1}}.
\]
Then each $\pi_j$ is a probability vector and
\[
r_j=\sum_{k=1}^K\pi_j(k)v_k^\star.
\]
If word $j$ is an anchor word for topic $k$, then $r_j=v_k^\star$. Consequently the rows of $R$ lie in a nondegenerate simplex with vertices $v_1^\star,\dots,v_K^\star$. Moreover,
\begin{equation}\label{eq:oracle_Astar_M0}
A\diag(w)=M_0^{1/2}\diag(\xi_1)\Pi,
\qquad
\Pi=[\pi_1,\dots,\pi_p]^\top.
\end{equation}

\begin{remark}[Why an arbitrary orthonormal basis is insufficient]\label{rem:arbitrary-basis}
Projector averaging recovers a subspace, not a distinguished first vector inside that subspace. If an arbitrary basis is rotated by a full $K\times K$ orthogonal matrix
\[
U=\begin{pmatrix}a&b^\top\\ c&D\end{pmatrix},
\]
then a row $\xi_1(j)(1,r_j)$ is transformed into
\[
\xi_1(j)(a+b^\top r_j,\ c+Dr_j),
\]
and the induced ratio map is
\[
r_j\mapsto \frac{c+Dr_j}{a+b^\top r_j},
\]
a projective transformation rather than a harmless $(K-1)$-dimensional rotation. This is why the federated procedure must first recover the canonical positive first direction from the pooled marginal.
\end{remark}

\subsubsection{Vertex hunting and barycentric recovery}\label{subsec:vh-barycentric}

Given empirical ratio points \(\hat r_j\), the Topic-SCORE simplex step first applies a vertex-hunting routine to the augmented points \(\hat y_j=(1,\hat r_j^\top)^\top\) and obtains vertices \(\hat v_1,\dots,\hat v_K\).  The corresponding unconstrained barycentric coordinates are defined by
\begin{equation}\label{eq:barycentric-appendix}
    \hat Q\hat\pi_j^\star=
    \binom{1}{\hat r_j},
    \qquad
    \hat Q=
    \begin{pmatrix}
        1&\cdots&1\\
        \hat v_1&\cdots&\hat v_K
    \end{pmatrix}.
\end{equation}
The simplex-normalized coordinate vector is
\[
    \hat\pi_j=\mathcal T(\hat\pi_j^\star),
    \qquad
    \mathcal T(b)=\frac{[b]_+}{\|[b]_+\|_1},
\]
with \(\mathcal T(b)=K^{-1}\mathbf 1\) if \([b]_+=0\).  In the federated analysis below, the empirical residual basis is identifiable only up to a \((K-1)\)-dimensional rotation; the perturbation proof therefore applies the same barycentric step after aligning the empirical ratio cloud by \(O_\perp\).

% =========================================================
\subsection{Federated Topic-SCORE via projector averaging}

\subsubsection{Protocol and communicated objects}\label{subsec:protocol}

The client-side spectral step uses the same SVD object as ordinary Topic-SCORE. After the server broadcasts the shared empirical whitening matrix $M_d$, client $m$ forms
\[
D_f^{(m)}:=M_d^{-1/2}D^{(m)}
\]
and computes its leading $K$ left singular vectors. Equivalently, for eigenvector purposes this is the same as diagonalizing the identity-centered Gram matrix
\begin{equation}\label{eq:svd-client-Gram-alg}
\hat G_m^{\mathrm{svd}}
:=
\frac1{n_m}M_d^{-1/2}D^{(m)}D^{(m)\top}M_d^{-1/2}-\frac1N I_p.
\end{equation}
The subtraction of $(1/N)I_p$ is only a scalar eigenvalue shift, so the eigenspace of \eqref{eq:svd-client-Gram-alg} is exactly the leading left singular subspace of $D_f^{(m)}$.

Let $\hat\Xi^{(m)}\in\R^{p\times K}$ contain the leading $K$ left singular vectors of $D_f^{(m)}$ and define the uploaded projector
\[
\hat P_m:=\hat\Xi^{(m)}\hat\Xi^{(m)\top}.
\]
The server averages
\[
\bar P:=\sum_{m=1}^M \alpha_m\hat P_m,
\qquad
\alpha_m\ge 0,
\qquad
\sum_{m=1}^M\alpha_m=1,
\]
and lets $\widehat P$ denote the top-$K$ spectral projector of $\bar P$.

The empirical canonical first direction is
\[
u_d:=M_d^{1/2}\1,
\qquad
\hat\xi_1:=\frac{\widehat P u_d}{\|\widehat P u_d\|_2}.
\]
Let
\[
\widehat P_\perp:=\widehat P-\hat\xi_1\hat\xi_1^\top.
\]
Choose any orthonormal basis $\hat\Xi_\perp$ of $\mathrm{range}(\widehat P_\perp)$, form the empirical ratio matrix
\[
\hat R(j,\ell)=\frac{\hat\Xi_\perp(j,\ell)}{\hat\xi_1(j)},
\]
apply vertex hunting, recover barycentric coordinates $\hat\Pi$, and define
\[
\tilde A_{\mathrm{fed}}:=M_d^{1/2}\diag(\hat\xi_1)\hat\Pi.
\]
Column-normalizing $\tilde A_{\mathrm{fed}}$ gives $\hat A_{\mathrm{fed}}$.

\begin{algorithm}[ht]
\caption{Practical federated Topic-SCORE via SVD upload and canonical direction recovery}
\label{alg:fed-topic-score}
\begin{algorithmic}[1]
\Require Local corpora $\{D^{(m)}\}_{m=1}^M$, topic number $K$, weights $\{\alpha_m\}_{m=1}^M$, a vertex-hunting routine
\Ensure Topic matrix estimate $\hat A_{\mathrm{fed}}$
\State Clients send $\bar d^{(m)}=n_m^{-1}D^{(m)}\1$ to the server
\State Server aggregates $\bar d=n_{\mathrm{tot}}^{-1}D\1$ and broadcasts $M_d=\diag(\bar d)$
\For{each client $m=1,\dots,M$ in parallel}
    \State Form $D_f^{(m)}=M_d^{-1/2}D^{(m)}$
    \State Compute the leading $K$ left singular vectors $\hat\Xi^{(m)}$ of $D_f^{(m)}$
    \State Upload $\hat\Xi^{(m)}$, equivalently $\hat P_m=\hat\Xi^{(m)}\hat\Xi^{(m)\top}$
\EndFor
\State Server forms $\bar P=\sum_{m=1}^M\alpha_m\hat P_m$ and extracts its top-$K$ projector $\widehat P$
\State Recover $\hat\xi_1=\widehat P M_d^{1/2}\1/\|\widehat P M_d^{1/2}\1\|_2$
\State Set $\widehat P_\perp=\widehat P-\hat\xi_1\hat\xi_1^\top$ and choose $\hat\Xi_\perp$ spanning $\mathrm{range}(\widehat P_\perp)$
\State Form $\hat R(j,\ell)=\hat\Xi_\perp(j,\ell)/\hat\xi_1(j)$
\State Apply vertex hunting to the rows of $\hat R$ and obtain $\hat v_1,\dots,\hat v_K$
\For{$j=1,\dots,p$}
    \State Solve $\begin{pmatrix}1&\cdots&1\\ \hat v_1&\cdots&\hat v_K\end{pmatrix}\hat\pi_j^\star=\begin{pmatrix}1\\ \hat r_j\end{pmatrix}$
    \State Truncate negative coordinates and $\ell_1$-normalize to obtain $\hat\pi_j$
\EndFor
\State Set $\hat\Pi=[\hat\pi_1,\dots,\hat\pi_p]^\top$, $\tilde A_{\mathrm{fed}}=M_d^{1/2}\diag(\hat\xi_1)\hat\Pi$
\State Column-normalize $\tilde A_{\mathrm{fed}}$ and output $\hat A_{\mathrm{fed}}$
\end{algorithmic}
\end{algorithm}
\begin{remark}[Computing the practical SVD upload]\label{rem:computational-centered-gram}
Lines 1--2 of Algorithm~\ref{alg:fed-topic-score} are a marginal-aggregation round used only to construct the shared whitening matrix. After $M_d$ is broadcast, the practical spectral upload is obtained by a standard truncated SVD of $D_f^{(m)}=M_d^{-1/2}D^{(m)}$. Equivalently, this is the eigenspace of the identity-centered Gram matrix \eqref{eq:svd-client-Gram-alg}; the identity shift does not affect eigenvectors. Thus the implementation can use the same sparse-matrix SVD routines as single-site Topic-SCORE.
% CAMERA_V1_BEGIN P-A04
Truncated SVD also provides the canonical best rank-(K) approximation of the local matrix and returns an orthonormal basis, making rotation-invariant projector aggregation straightforward. In contrast, methods that directly estimate local topics, such as LDA or NMF, typically require iterative nonconvex optimization and introduce permutation, scaling, and cross-site topic-alignment ambiguities. More robust or weighted subspace estimators could in principle replace SVD under alternative contamination or heteroscedastic noise assumptions, but they require additional modeling assumptions and tuning parameters. Under the noise model considered in this paper, SVD is therefore the simplest and most direct estimator of the model-implied shared subspace, while also enabling a single-round federated implementation.
% CAMERA_V1_END P-A04

\end{remark}

\begin{remark}[Optional centered local spectral upload]\label{rem:optional-debiased-upload}
The ordinary-SVD upload in Algorithm~\ref{alg:fed-topic-score} is computationally
convenient because each client can use a standard truncated SVD of
\(M_d^{-1/2}D^{(m)}\). In sparse or highly heterogeneous settings, however, its
fixed-whitening population target contains the deterministic diagonal term
\(D_{0,m}^{\rm svd}\). This term is an artifact of the ordinary-SVD identity shift,
not of the federated projector-averaging principle. An optional diagonal-debiased
variant lets client \(m\) form
\[
\widehat G_m^{\rm deb}
=
n_m^{-1}M_d^{-1/2}D^{(m)}D^{(m)\top}M_d^{-1/2}
-
N^{-1}M_d^{-1/2}M_d^{(m)}M_d^{-1/2},
\]
where \(M_d^{(m)}=\diag(n_m^{-1}D^{(m)}\mathbf 1)\), and upload the top-\(K\)
eigenspace of \(\widehat G_m^{\rm deb}\). The server-side projector averaging,
canonical direction recovery, ratio construction, vertex hunting, and column
normalization are unchanged. This variant preserves the same rank-\(K\) spectral
communication order but removes the ordinary-SVD diagonal-bias residue from the
perturbation radius, replacing it with the corresponding centered-Gram concentration
terms. The main text reports the ordinary-SVD implementation, so this centered upload
is presented as an optional extension rather than as the default algorithm.
\end{remark}

\begin{remark}[Communication and computation budget]\label{rem:communication-budget}
The non-iterative summary-sharing protocol first sends one \(p\)-vector
\(\bar d^{(m)}\) from each client, which is needed to form the common whitening
matrix. After the server broadcasts \(M_d\), each client sends a \(p\times K\)
orthonormal basis \(\hat\Xi^{(m)}\); the equivalent projector \(\hat P_m\) need not be
transmitted or stored as a dense \(p\times p\) matrix. The server can apply the
averaged projector to a vector by
\[
\bar P x=\sum_{m=1}^M\alpha_m\hat\Xi^{(m)}\{\hat\Xi^{(m)\top}x\},
\]
so extracting the top-$K$ eigenspace of $\bar P$ can also be done matrix-free. Thus the spectral communication after whitening is $O(pK)$ numbers per client, in addition to the one-time marginal vector.
% CAMERA_V1_BEGIN P-A01
The iterative-free and spectral properties of our method have significant advantages in reducing communication and computation costs over iterative approaches. For a representative iterative Federated LDA procedure with $R$ communication rounds and $E$ local inference iterations per round, site $m$ performs approximately $O(REpn_mK)$ local computation and communicates a $p\times K$ topic matrix or sufficient statistic over $R$ rounds, for $O(RpK)$ communication. In contrast, Federated Topic-SCORE performs only one rank-$K$ truncated SVD of the local $p\times n_m$ matrix, with cost $O(pn_mK)$ up to a constant number of matrix passes, and communicates a $p$-vector and a $p\times K$ spectral summary, for fixed $O(p+pK)$ communication per site.

When site $m$ receives $\Delta n_m$ new patients, keeping the current $\widehat A$ lets Equation (4) estimate their topic weights locally at $O(pK\Delta n_m)$ cost and no communication. Updating $A$ uses one non-iterative refresh of the enlarged matrix, costing $O\{p(n_m+\Delta n_m)K\}$ locally and $O(p+pK)$ in communication. An iterative update repeats this work over $R$ rounds, costing $O\{REp(n_m+\Delta n_m)K\}$ locally and $O(RpK)$ in communication. The non-iterative design therefore avoids repeated computation and communication as cohorts grow.
% CAMERA_V1_END P-A01

\end{remark}

\begin{remark}[Zero empirical frequencies and the definition of $M_d^{-1/2}$]\label{rem:zero-frequency}
The theoretical statements are made on the high-probability event on which $M_d(j,j)\asymp M_0(j,j)$ for every word $j$; on this event $M_d^{-1/2}$ is well defined. An implementation can use the harmless floored version
\[
M_{d,\tau}:=\diag(\bar d_1\vee \tau,\ldots,\bar d_p\vee\tau)
\]
with $\tau$ smaller than a sufficiently small constant multiple of $p^{-1}$.
Under Assumptions~\ref{assump:hmin}--\ref{assump:Sigma}, the floor is inactive
on the perturbative marginal event in Lemma~\ref{lem:Md_concentration}.
Fix measurable rules for all spectral and vertex-hunting ties; each local
fixed-whitening spectral rule depends only on that client's matrix.
If a required spectral matrix is undefined, assign the fixed rank-$K$
orthogonal projector $P_{\mathrm{fb}}=\sum_{k=1}^K e_ke_k^\top$.
To define the estimator on every outcome, if a required inverse does not
exist, a normalization denominator vanishes, a canonical coordinate is
nonpositive, or vertex hunting fails to return an invertible simplex, set
\[
\widehat A_{\mathrm{fed}}=p^{-1}\1_p\1_K^\top,
\qquad \widehat w_i=K^{-1}\1_K.
\]
Use the same fixed weight fallback if the weighted least-squares matrix is
singular or simplex normalization is undefined, and assign fixed finite
values to any still undefined intermediate coordinates.
These conventions affect only outcomes outside the good events used in the
proofs; on those events all original formulas apply.
\end{remark}

\subsubsection{Fixed-whitening matrices and the identity-centered SVD bridge}\label{subsec:Gdef}

The algorithmic projector is exactly the ordinary-SVD projector uploaded in Algorithm~\ref{alg:fed-topic-score}. Throughout the main text and the proof, set
\begin{equation}\label{eq:Ghatm_svd_def}
\hat G_m
:=
\frac1{n_m}M_d^{-1/2}D^{(m)}D^{(m)\top}M_d^{-1/2}-\frac1N I_p,
\end{equation}
so that \(\hat P_m\) is the top-\(K\) spectral projector of the same matrix whose eigenvectors are the leading left singular vectors uploaded by client \(m\). The fixed-whitening comparison matrix is
\begin{equation}\label{eq:Gtilde_svd_def}
\tilde G_m
:=
\frac1{n_m}M_0^{-1/2}D^{(m)}D^{(m)\top}M_0^{-1/2}-\frac1N I_p,
\end{equation}
and the rank-\(K\) common-space target is
\begin{equation}\label{eq:G0m_def}
G_{0,m}:=\Bigl(1-\frac1N\Bigr)\frac1{n_m}M_0^{-1/2}D_0^{(m)}D_0^{(m)\top}M_0^{-1/2}.
\end{equation}
Let \(\tilde P_m\) be the top-\(K\) projector of \(\tilde G_m\).

The identity shift \(N^{-1}I_p\) is algorithmically convenient and is exactly what ordinary SVD implements. It leaves an additional deterministic diagonal term in the fixed-whitening population mean:
\begin{equation}\label{eq:Dsvd_pop_def}
D_{0,m}^{\mathrm{svd}}
:=\E[\tilde G_m\mid W^{(1:M)}]-G_{0,m}
=\frac1N\left(M_0^{-1/2}M_{0,m}M_0^{-1/2}-I_p\right).
\end{equation}
This is not a centering correction; it is the explicit price of using the ordinary SVD identity shift rather than a local diagonal debiasing term. We measure its row-wise size by
\begin{equation}\label{eq:Delta_svd_def}
\Delta_{\mathrm{svd}}
:=\max_{1\le m\le M}\left\{
\|D_{0,m}^{\mathrm{svd}}\|+
\max_{1\le j\le p}h_j^{-1/2}\|D_{0,m}^{\mathrm{svd}}e_j\|_2
\right\}.
\end{equation}
Under Assumptions~\ref{assump:hmin}--\ref{assump:Sigma}, \(\Delta_{\mathrm{svd}}\lesssim \sqrt p/N\); this deterministic term is included in the final radius.

Because
\[
G_{0,m}=\Bigl(1-\frac1N\Bigr)B\Sigma_{W,m}B^\top,
\qquad
B:=M_0^{-1/2}A,
\]
all \(G_{0,m}\) share the same range \(\mathcal S=\col(B)\). Let \(P\) denote the orthogonal projector onto \(\mathcal S\).

\begin{lemma}[Identity-centered fixed-whitening mean and bridge decomposition]\label{lem:Md_bridge_decomp}
Fix \(m\in[M]\) and define
\[
\hat S_m:=\frac1{n_m}D^{(m)}D^{(m)\top},
\qquad
T_m:=M_0^{-1/2}\hat S_mM_0^{-1/2}=\tilde G_m+\frac1N I_p,
\]
\[
R_M:=M_d^{-1/2}M_0^{1/2}-I_p.
\]
Then
\begin{equation}\label{eq:fixed-mean-svd}
\E[\tilde G_m\mid W^{(1:M)}]=G_{0,m}+D_{0,m}^{\mathrm{svd}},
\end{equation}
and
\begin{equation}\label{eq:Md_bridge_decomp_eq}
\hat G_m-\tilde G_m
=R_MT_m+T_mR_M+R_MT_mR_M.
\end{equation}
Consequently,
\begin{equation}\label{eq:svd-main-decomp}
\hat G_m-G_{0,m}
=\underbrace{\{\tilde G_m-\E(\tilde G_m\mid W^{(1:M)})\}}_{\text{centered sampling fluctuation}}
+D_{0,m}^{\mathrm{svd}}
+R_MT_m+T_mR_M+R_MT_mR_M.
\end{equation}
\end{lemma}

The proof decomposition above follows the algorithmic identity-centered SVD matrix \(\hat G_m\) throughout. The term \(D_{0,m}^{\mathrm{svd}}\) is a deterministic diagonal perturbation of the common-space signal, and the three terms in \eqref{eq:Md_bridge_decomp_eq} are the shared-whitening bridge induced by replacing \(M_0\) with \(M_d\).

Define the pooled marginal deviation
\begin{equation}\label{eq:Delta1_def}
\Delta_1(D,D_0):=\max_{1\le j\le p}h_j^{-1}|M_d(j,j)-M_0(j,j)|.
\end{equation}

\subsubsection{Abstract projector averaging}\label{subsec:agg}

The next results hold for any independent rank-$K$ projectors $\{\mathcal P_m\}_{m=1}^M$ with common deterministic target $P$ and deterministic averaging weights $\alpha_m\ge0$ satisfying $\sum_{m=1}^M\alpha_m=1$. Set
\[
\bar{\mathcal P}:=\sum_{m=1}^M\alpha_m\mathcal P_m,
\qquad
\nu_m:=\E\|\mathcal P_m-P\|^2.
\]

\begin{assumption}[Client independence]\label{assump:ind}
The projectors being averaged are mutually independent.
\end{assumption}

\begin{lemma}[Davis--Kahan with a projector signal]\label{lem:dk-projector}
Let $P$ be a rank-$K$ projector and let $\bar{\mathcal P}$ be symmetric. If $\|\bar{\mathcal P}-P\|\le 1/4$, then the top-$K$ spectral projector of $\bar{\mathcal P}$ is uniquely defined and satisfies
\[
\|\widehat P-P\|\le 2\|\bar{\mathcal P}-P\|.
\]
More generally, the same conclusion holds with another absolute constant whenever $\|\bar{\mathcal P}-P\|\le c_{\mathrm{DK}}<1/2$.
\end{lemma}

\begin{lemma}[Variance-bias decomposition]\label{lem:vb-decomp}
For deterministic weights $\alpha_m$ satisfying $\sum_{m=1}^M\alpha_m=1$,
\[
\|\bar{\mathcal P}-P\|
\le
\Bigl\|\sum_{m=1}^M\alpha_m(\mathcal P_m-\E\mathcal P_m)\Bigr\|
+
\Bigl\|\sum_{m=1}^M\alpha_m(\E\mathcal P_m-P)\Bigr\|.
\]
No independence is needed for this deterministic decomposition; independence enters only in the concentration step below.
\end{lemma}

\begin{lemma}[Matrix Bernstein for projector averaging]\label{lem:bernstein-matrix}
Let $X_m:=\alpha_m(\mathcal P_m-\E\mathcal P_m)$ and $S:=\sum_mX_m$. Then $\|X_m\|\le 2\alpha_m$ and
\[
\Bigl\|\sum_{m=1}^M\E[X_m^2]\Bigr\|
\le 4\sum_{m=1}^M\alpha_m^2\nu_m.
\]
Hence, for any $\delta\in(0,1)$, with probability at least $1-\delta$,
\[
\|S\|
\le 4\sqrt{2\log\!\frac{2p}{\delta}}\Bigl(\sum_{m=1}^M\alpha_m^2\nu_m\Bigr)^{1/2}
+
\frac43\log\!\frac{2p}{\delta}\max_m\alpha_m.
\]
\end{lemma}

\begin{lemma}[Second-order spectral-projector bias]\label{lem:second-order-bias}
Suppose $G_m=G+E_m$ with $\E E_m=0$, and let $P$ and $P_m$ be the spectral projectors onto the same isolated size-$K$ spectral cluster with eigengap $\gamma>0$. Assume the target cluster has diameter at most $C_0\gamma$ for a fixed constant $C_0$; in the applications below, all nonzero population eigenvalues are bounded above and below by constants, so this condition is automatic. Assume further that $\|E_m\|\le \gamma/4$ almost surely. With $C$ depending only on $C_0$, we have
\[
\|\E P_m-P\|\le \frac{C}{\gamma^2}\E\|E_m\|^2.
\]
If, in addition, $G\succeq0$ has rank $K$, $P$ is its nonzero-eigenspace projector, and the remaining spectrum is exactly zero, then for every $j\in[p]$,
\begin{equation}\label{eq:second-order-col}
\|(\E P_m-P)e_j\|_2
\le
\frac C{\gamma^2}
\left[
(\E\|E_m\|^2)^{1/2}(\E\|E_me_j\|_2^2)^{1/2}
+\|Pe_j\|_2\,\E\|E_m\|^2
\right].
\end{equation}
\end{lemma}
\begin{remark}[High-probability versions of the second-order bias bound]
Lemma~\ref{lem:second-order-bias} is stated with an
almost-sure small-perturbation condition. When this
condition holds only on
\(\mathcal E_m=\{\|E_m\|\le\gamma/4\}\), apply the Riesz
expansion on \(\mathcal E_m\). Since
\[
\mathbb E[\mathcal L_G(E_m)\mathbf 1_{\mathcal E_m}]
=
-\mathbb E[\mathcal L_G(E_m)\mathbf 1_{\mathcal E_m^c}],
\]
the truncation argument also controls the linear tail
term, in addition to the projector contribution on
\(\mathcal E_m^c\). In the multinomial application, the
relevant tail terms are bounded by
Lemma~\ref{lem:operator-tail-truncation}, while the
deterministic ordinary-SVD bias is treated separately,
as in the proof of Lemma~\ref{lem:tau-bound-strong}. The columnwise display is intentionally stated only for the rank-$K$ signal case. This is the case used for the fixed-whitening population matrices $G_{0,m}$; for a general isolated cluster we use only the operator bias bound and the operator remainder in Lemma~\ref{lem:proj-expansion}.
\end{remark}

\begin{remark}[Why the columnwise bias bound contains $\|Pe_j\|_2$]\label{rem:second-order-bias-col}
The columnwise part of Lemma~\ref{lem:second-order-bias} is deliberately stated only for the rank-$K$ signal case used in Lemma~\ref{lem:tau-bound-strong}. In that case the non-target spectrum is zero, so the resolvent on $P^\perp$ is scalar and the columnwise remainder can retain the factor $\|E_me_j\|_2$. The term $\|Pe_j\|_2\,\E\|E_m\|^2$ is essential: a second-order resolvent remainder can act first on the signal component of $e_j$, and that component is controlled in this paper by the oracle row-scale bound $\|Pe_j\|_2\lesssim\sqrt{h_j}$.
\end{remark}

\begin{theorem}[Federated projector averaging bound]\label{thm:projector-avg}
Under Assumption~\ref{assump:ind} and deterministic weights $\alpha_m\ge0$ with $\sum_m\alpha_m=1$, for any $\delta\in(0,1)$, with probability at least $1-\delta$,
\begin{equation}\label{eq:projector-avg}
\|\bar{\mathcal P}-P\|
\le
\Bigl\|\sum_{m=1}^M\alpha_m(\E\mathcal P_m-P)\Bigr\|
+4\sqrt{2\log\!\frac{2p}{\delta}}\Bigl(\sum_{m=1}^M\alpha_m^2\nu_m\Bigr)^{1/2}
+\frac43\log\!\frac{2p}{\delta}\max_m\alpha_m.
\end{equation}
If the right-hand side is at most $1/4$, then the top-$K$ spectral projector $\widehat P$ of $\bar{\mathcal P}$ is well defined and obeys $\|\widehat P-P\|\le 2\|\bar{\mathcal P}-P\|$.
\end{theorem}

\begin{remark}[Which projectors are averaged in the theorem]
The practical projectors $\hat P_m$ are coupled through the shared random matrix $M_d$. The theorem is applied first to the fixed-whitening ordinary-SVD projectors $\tilde P_m$, which are independent conditional on $W^{(1:M)}$. Their population target differs from $G_{0,m}$ by the deterministic diagonal term $D_{0,m}^{\mathrm{svd}}$, and the empirical-whitening bridge $\hat P_m-\tilde P_m$ is handled deterministically.
\end{remark}

\begin{lemma}[Vector Bernstein for projector columns]\label{lem:vector-bernstein}
Fix $j\in[p]$ and define
\[
r_j:=\Bigl(\sum_{m=1}^M\alpha_m\mathcal P_m-P\Bigr)e_j.
\]
Let $\tau_{m,j}^2:=\E\| (\mathcal P_m-P)e_j\|_2^2$. Then, for any $\delta\in(0,1)$, with probability at least $1-\delta$,
\[
\|r_j-\E r_j\|_2
\le
\sqrt{2\log\!\frac2\delta}\Bigl(\sum_{m=1}^M\alpha_m^2\tau_{m,j}^2\Bigr)^{1/2}
+\frac43\log\!\frac2\delta\max_m\alpha_m.
\]
After a union bound over $j$, the row-normalized version of this display contributes the conservative range term $h_{\min}^{-1/2}\ell_{n,M}(\delta)/M\lesssim \sqrt p\,\ell_{n,M}(\delta)/M$ used in Theorem~\ref{thm:fed-rowwise}.
\end{lemma}

\subsubsection{Specialization to federated Topic-SCORE}\label{subsec:rates}

For readability we now specialize to equal client sizes and equal averaging weights,
\[
n_1=\cdots=n_M=n,
\qquad
\alpha_m=1/M.
\]
The abstract projector-averaging results above continue to hold for general $\{n_m,\alpha_m\}$; the same variance--bias decomposition then carries the weights $\alpha_m$ and the local sample sizes $n_m$ explicitly. We do not expand that notation below. Define
\[
\beta_n:=
\begin{cases}
1+\min\{p/N,\ p^2/N^{3/2}\}, & n\ge \max\{Np^2,\ p^3,\ N^2p^5\},\\[0.3em]
1+p^2/N^{3/2}, & n<\max\{Np^2,\ p^3,\ N^2p^5\}.
\end{cases}
\]
Also set
\[
\ell_{n,M}(\delta):=\log\!\frac{2pM}{\delta},
\]
\[
\begin{aligned}
\vartheta_n(\delta)
&:=
\beta_n\sqrt{\frac{p\ell_{n,M}(\delta)}{Nn}}
+\beta_n^2\frac{p\ell_{n,M}(\delta)}{Nn}
+\sqrt{\beta_n p}\,\frac{p\ell_{n,M}(\delta)}{Nn}
+\beta_n\sqrt{\frac pN}\,\frac{\ell_{n,M}(\delta)}{n},\\
\eta_{n,M}^{\mathrm{marg}}(\delta)
&:=
\sqrt{\frac{p\ell_{n,M}(\delta)}{NnM}}
+\frac{p\ell_{n,M}(\delta)}{NnM}.
\end{aligned}
\]
Define also the deterministic ordinary-SVD diagonal-bias radius
\[
\zeta_N^{\mathrm{svd}}:=\frac{\sqrt p}{N}.
\]
The linear-in-$\ell_{n,M}$ terms in $\vartheta_n(\delta)$ retain
the range contributions in the document-level Bernstein bounds.
The term $\sqrt{\beta_n p}\,p\ell/(Nn)$ controls higher moments of
the product term in the column decomposition. The term
$\beta_n\sqrt{p/N}\,\ell/n$ is used in both the Frobenius/operator
and column Bernstein bounds. These terms are kept explicitly in the
local concentration interface.

\begin{lemma}[Fixed-design heterogeneity envelope]\label{lem:beta-envelope}
Assume Assumptions~\ref{assump:hmin}--\ref{assump:Sigma}. For every client $m$ and document $i$, let
\[
\pi_i^{(m)}:=Aw_i^{(m)}.
\]
Then, with a constant depending only on the regularity constants and fixed $K$,
\[
\max_{m,i}\max_{1\le j\le p}\frac{\pi_i^{(m)}(j)}{m_0(j)}\le C,
\qquad
\max_{m,i}\sum_{j=1}^p\frac{\pi_i^{(m)}(j)}{m_0(j)}\le Cp,
\qquad
\max_{m,i}\sum_{j=1}^p\frac{\{\pi_i^{(m)}(j)\}^2}{m_0(j)}\le C.
\]
Since $\beta_n\ge1$, these bounds imply the envelope used in the multinomial concentration lemmas with $C\beta_n$, $C\beta_n p$, and $C\beta_n$ on the right-hand sides. Thus the envelope is not an additional assumption in the regular theorem class.
\end{lemma}

\begin{proof}
Let $\bar w=n_{\mathrm{tot}}^{-1}W\1$. From $\lambda_{\min}(\Sigma_W)\ge c_2$ and $0\le w_i(k)\le1$,
\[
\bar w_k=\frac1{n_{\mathrm{tot}}}\sum_i w_i(k)\ge \frac1{n_{\mathrm{tot}}}\sum_i w_i(k)^2=\Sigma_W(k,k)\ge c_2.
\]
Therefore $m_0(j)=a_j^\top\bar w\ge c_2 h_j$, while $\pi_i^{(m)}(j)=a_j^\top w_i^{(m)}\le h_j$. Hence $\pi_i^{(m)}(j)/m_0(j)\le c_2^{-1}$ uniformly. Summing this inequality over $j$ gives the second display. Finally,
\[
\sum_j\frac{\{\pi_i^{(m)}(j)\}^2}{m_0(j)}
\le \Bigl(\max_j\frac{\pi_i^{(m)}(j)}{m_0(j)}\Bigr)\sum_j\pi_i^{(m)}(j)
\le C,
\]
because each $\pi_i^{(m)}$ is a probability vector. This proves the lemma.
\end{proof}

% CAMERA_V1_BEGIN P-A02
\begin{remark}[Unequal sample sizes across sites]\label{rem:unequal-site-sizes}
We focused on a setting with equal site size only to simplify the main theorem. For unequal site sizes, a natural choice of weights is $\alpha_m=\frac{n_m}{n_{\mathrm{tot}}}$. With this choice, the leading variance term is $O\left(\sqrt{\frac{p\log(pM/\delta)}{Nn_{\mathrm{tot}}}}\right)$, up to the same sparsity factor as in the current theorem. This is the same order as the bound in Theorem 1, because when $n_m=n$ for all sites, $n_{\mathrm{tot}}=nM$ and the term becomes $O\left(\sqrt{\frac{p\log(pM/\delta)}{NnM}}\right)$. The corresponding proof requires only bookkeeping changes. We replace $1/M$ by $\alpha_m$, use $n_m$ in each site-specific concentration bound, and apply the weighted matrix and vector Bernstein inequalities. The local bias becomes a weighted sum of the site-specific bias terms, and the range term depends on $\max_m\alpha_m$. The remaining projector recovery and Topic-SCORE arguments are unchanged.
\end{remark}
% CAMERA_V1_END P-A02

\begin{remark}[Origin and interpretation of the heavy-tail coefficient]\label{rem:beta-envelope-origin}
The coefficient $\beta_n$ is inherited from the single-site SVD Topic-SCORE analysis of Ke and Wang's SVD Topic-SCORE paper. It packages the extra Bernstein/Freedman range terms that arise when multinomial word-count vectors are sparse or heavy-tailed after whitening. In the generic branch, $N\gtrsim p^{4/3}$ ensures $\beta_n=O(1)$. In the first branch, where the displayed additional condition on $n$ holds, $N\gtrsim p$ suffices. In either case the variance component has the familiar sample-size scaling. When $N\ll p$, rare word-level increments have larger bounded range after whitening and $\beta_n$ grows, reflecting the same heavy-tail correction as in the row-wise singular-vector theory for single-site Topic-SCORE. Lemma~\ref{lem:beta-envelope} is not an algorithmic requirement; it records the deterministic envelope under which the multinomial moment bounds in Appendix~\ref{app:multinomial} are uniform over clients and documents.
\end{remark}

\begin{lemma}[Single-client projector bounds for the practical SVD upload]\label{lem:tau-bound-strong}
Assume Assumptions~\ref{assump:hmin}--\ref{assump:local-richness} and
\[
\beta_n^2\frac{p^2}{Nn}\le c_0,\qquad \zeta_N^{\mathrm{svd}}\le c_0
\]
for a sufficiently small constant \(c_0>0\). For each client \(m\), let \(\gamma_m:=\lambda_K(G_{0,m})\). Then \(\gamma_m\ge c_{\mathrm{gap}}>0\) uniformly in \(m\). Moreover, for the fixed-whitening ordinary-SVD projector \(\tilde P_m\),
\[
\E\| (\tilde P_m-P)e_j\|_2^2\le C h_j\left(\beta_n^2\frac p{Nn}+(\zeta_N^{\mathrm{svd}})^2\right),
\qquad
\E\|\tilde P_m-P\|^2\le C\left(\beta_n^2\frac p{Nn}+(\zeta_N^{\mathrm{svd}})^2\right),
\]
and
\[
\|(\E\tilde P_m-P)e_j\|_2\le C\sqrt{h_j}\left(\beta_n^2\frac p{Nn}+\zeta_N^{\mathrm{svd}}\right),
\qquad
\|\E\tilde P_m-P\|\le C\left(\beta_n^2\frac p{Nn}+\zeta_N^{\mathrm{svd}}\right).
\]
For later use, define the client-level good events
\[
\mathcal E_{\mathrm{loc},m}:=\{\|\tilde G_m-G_{0,m}\|\le c_{\mathrm{gap}}/8\},
\]
\[
\mathcal E_{\mathrm{col},m}(\delta):=
\Bigl\{
\max_{1\le j\le p}h_j^{-1/2}\|(\tilde G_m-G_{0,m})e_j\|_2
\le C_{\mathrm{col}}\{\vartheta_n(\delta)+\zeta_N^{\mathrm{svd}}\}
\Bigr\},
\]
and
\[
\mathcal E_{\mathrm{bridge},m}:=
\{\Delta_1(D,D_0)\le c_{\mathrm{br}},\ \|\hat G_m-\tilde G_m\|\le C_0\Delta_1(D,D_0)\},
\]
where \(c_{\mathrm{br}}>0\) is chosen sufficiently small and \(C_{\mathrm{col}},C_0>0\) are absolute constants.

On \(\mathcal E_{\mathrm{loc},m}\cap \mathcal E_{\mathrm{bridge},m}\),
\[
\begin{aligned}
\|(\hat P_m-\tilde P_m)e_j\|_2
&\le C\sqrt{h_j}\Delta_1(D,D_0)
 +C\Delta_1(D,D_0)\|(\tilde G_m-G_{0,m})e_j\|_2\\
&\quad +C\Delta_1(D,D_0)^2\{1+\|\tilde G_m-G_{0,m}\|\}^2
+C\frac{\Delta_1(D,D_0)}{N}\|e_j\|_2,
\end{aligned}
\]
and
\[
\|\hat P_m-\tilde P_m\|
\le C\Delta_1(D,D_0)+C\Delta_1(D,D_0)\|\tilde G_m-G_{0,m}\|.
\]
Consequently, on \(\mathcal E_{\mathrm{loc},m}\cap\mathcal E_{\mathrm{col},m}(\delta)\cap\mathcal E_{\mathrm{bridge},m}\),
\[
\|\hat P_m-\tilde P_m\|\le C\Delta_1(D,D_0),
\]
and
\[
\max_{1\le j\le p}h_j^{-1/2}\|(\hat P_m-\tilde P_m)e_j\|_2
\le C\Delta_1(D,D_0)\{1+\vartheta_n(\delta)+\zeta_N^{\mathrm{svd}}\}
+C\sqrt p\,\Delta_1(D,D_0)^2+C\frac{\sqrt p}{N}\Delta_1(D,D_0).
\]
\end{lemma}

\begin{corollary}[Single-client projector operator-norm bound]
\label{cor:single-client-op}
Under the assumptions of Lemma~\ref{lem:tau-bound-strong},
for every client \(m\) and every \(\delta\in(0,1)\),
\[
\begin{aligned}
\|\hat P_m-P\|
&\le C\Biggl[
\beta_n\sqrt{\frac{p\log(2p/\delta)}{Nn}}
+\beta_n^2\frac{p\log(2p/\delta)}{Nn}
\\
&\qquad
+\sqrt{\beta_n p}\,\frac{p\log(2p/\delta)}{Nn}
+\beta_n\sqrt{\frac pN}\,
  \frac{\log(2p/\delta)}{n}
\\
&\qquad
+\Delta_1(D,D_0)
+\zeta_N^{\mathrm{svd}}
\Biggr]
\end{aligned}
\]
with probability at least \(1-\delta\).
\end{corollary}

\begin{lemma}[Projector-signal row extraction]\label{lem:proj-col-from-avg}
Let $P=\Xi\Xi^\top$ be a rank-$K$ projector satisfying $\|e_j^\top\Xi\|_2\le C\sqrt{h_j}$ for all $j$. Let $\bar P=P+E$ be symmetric, and let $\widehat P$ be the top-$K$ spectral projector of $\bar P$. If $\|E\|\le c$ for a sufficiently small constant $c>0$, then
\[
\|\widehat P-P\|+\max_{1\le j\le p}h_j^{-1/2}\|(\widehat P-P)e_j\|_2
\le
C\Bigl[\|E\|+\max_{1\le j\le p}h_j^{-1/2}\|Ee_j\|_2\Bigr].
\]
\end{lemma}

\begin{lemma}[Canonical first-direction recovery]\label{lem:canonical-first}
Assume
\[
\Delta_1(D,D_0)+\|\widehat P-P\|\le c
\]
for a sufficiently small constant $c>0$. Define $u_d=M_d^{1/2}\1$ and $\hat\xi_1=\widehat P u_d/\|\widehat P u_d\|_2$. Then
\[
\max_{1\le j\le p}h_j^{-1/2}|\hat\xi_1(j)-\xi_1(j)|
\le C\Biggl[\Delta_1(D,D_0)+
\max_{1\le j\le p}h_j^{-1/2}\|(\widehat P-P)e_j\|_2+
\|\widehat P-P\|\Biggr].
\]
\end{lemma}

\begin{lemma}[Residual basis recovery]\label{lem:residual-basis}
Let $P_\perp=P-\xi_1\xi_1^\top$ and $\widehat P_\perp=\widehat P-\hat\xi_1\hat\xi_1^\top$. Choose orthonormal bases $\Xi_\perp$ and $\hat\Xi_\perp$ of their ranges. If the right-hand side in Lemma~\ref{lem:canonical-first} is sufficiently small, then there exists an orthogonal matrix $O_\perp\in\R^{(K-1)\times(K-1)}$ such that
\[
\max_{1\le j\le p}h_j^{-1/2}\|e_j^\top(\hat\Xi_\perp O_\perp-\Xi_\perp)\|_2
\le C\Biggl[\Delta_1(D,D_0)+
\max_{1\le j\le p}h_j^{-1/2}\|(\widehat P-P)e_j\|_2+
\|\widehat P-P\|\Biggr].
\]
\end{lemma}

Define the deterministic perturbative radius

\begin{equation}\label{eq:epsilon-nM}
\begin{aligned}
\varepsilon_{n,M}(\delta):={}&
\beta_n\sqrt{\frac{p\ell_{n,M}(\delta)}{NnM}}
+\beta_n^2\frac{p\ell_{n,M}(\delta)}{Nn}\\
&+\eta_{n,M}^{\mathrm{marg}}(\delta)\{1+\vartheta_n(\delta)+\zeta_N^{\mathrm{svd}}\}
+\sqrt p\,\{\eta_{n,M}^{\mathrm{marg}}(\delta)\}^2
+\zeta_N^{\mathrm{svd}}\\
&+\frac{\sqrt p\,\ell_{n,M}(\delta)}{M}
+\Bigl(\beta_n^2\frac p{Nn}\Bigr)^2.
\end{aligned}
\end{equation}
The factor $\eta_{n,M}^{\mathrm{marg}}(\delta)\vartheta_n(\delta)$ is the row-wise bridge interaction created by the term $\Delta_1(D,D_0)\|(\tilde G_m-G_{0,m})e_j\|_2$ in Lemma~\ref{lem:tau-bound-strong}. The factor $\sqrt p\{\eta_{n,M}^{\mathrm{marg}}(\delta)\}^2$ is the row-normalized quadratic projector-remainder term in the bridge step, and the small term $\sqrt p\,\eta_{n,M}^{\mathrm{marg}}(\delta)/N$ is absorbed by the displayed bridge radius under the stated perturbative regime. The deterministic term $\zeta_N^{\mathrm{svd}}$ is the price of using the ordinary-SVD identity shift: it controls the population diagonal perturbation $D_{0,m}^{\mathrm{svd}}$ in \eqref{eq:Dsvd_pop_def}. The factor $\sqrt p\,\ell_{n,M}(\delta)/M$ is the conservative row-wise Bernstein range term obtained after dividing the vector-Bernstein bound by $\sqrt{h_j}$ and using $h_{\min}\asymp p^{-1}$. The final square term $(\beta_n^2p/(Nn))^2$ is retained only as a harmless higher-order Riesz-remainder term from the fixed-whitening perturbation expansion; under the perturbative scaling it is absorbed by the displayed first-order local bias term, but keeping it visible makes the expansion auditable. The range term can be improved to $\ell_{n,M}(\delta)/M$ if one proves a localized almost-sure projector-column range bound, but the displayed form is the unconditional row-wise consequence used here.

\begin{theorem}[Federated row-wise canonical basis error]\label{thm:fed-rowwise}
Assume Assumptions~\ref{assump:hmin}--\ref{assump:local-richness}, $n_m=n$, $\alpha_m=1/M$, and
\[
\beta_n^2\frac{p^2}{Nn}\le c_0.
\]
Suppose also that the local fixed-whitening perturbation radius and the final deterministic radius satisfy
\[
\vartheta_n(\delta)\le c_{\mathrm{pert}},
\qquad
\zeta_N^{\mathrm{svd}}\le c_{\mathrm{diag}},
\qquad
\varepsilon_{n,M}(\delta)\le c_*,
\]
for sufficiently small constants $c_{\mathrm{pert}},c_{\mathrm{diag}},c_*>0$. Then, with probability at least $1-\delta$, there exists an orthogonal matrix $O_\perp\in\R^{(K-1)\times(K-1)}$ such that
\begin{equation}\label{eq:fed-rowwise}
\max_{1\le j\le p}h_j^{-1/2}
\Bigl\|e_j^\top\Bigl([\hat\xi_1,\hat\Xi_\perp O_\perp]-[\xi_1,\Xi_\perp]\Bigr)\Bigr\|_2
\le C\varepsilon_{n,M}(\delta).
\end{equation}
Consequently,
\[
\max_{1\le j\le p}\|O_\perp^\top\hat r_j-r_j\|_2\le C\varepsilon_{n,M}(\delta).
\]
\end{theorem}

\begin{remark}[The perturbative assumption is deterministic]\label{rem:deterministic-eps}
The smallness conditions in Theorem~\ref{thm:fed-rowwise} are deterministic. The condition $\vartheta_n(\delta)\le c_{\mathrm{pert}}$ guarantees the clientwise local spectral perturbation event used before projector averaging, while $\zeta_N^{\mathrm{svd}}\le c_{\mathrm{diag}}$ keeps the population diagonal SVD-bias perturbation local and $\varepsilon_{n,M}(\delta)\le c_*$ keeps the averaged projector and subsequent SCORE ratios perturbative. The radius $\varepsilon_{n,M}$ is stated after replacing the random marginal deviation $\Delta_1(D,D_0)$ by the concentration bound from Lemma~\ref{lem:Md_concentration}, inserting the local Gram-column radius from Lemma~\ref{lem:Ecol_hp} into the bridge term, adding the deterministic ordinary-SVD diagonal-bias radius $\zeta_N^{\mathrm{svd}}$, adding the row-normalized quadratic bridge remainder $\sqrt p\{\eta_{n,M}^{\mathrm{marg}}(\delta)\}^2$, and retaining the conservative row-wise Bernstein range term $\sqrt p\,\ell_{n,M}(\delta)/M$. Thus the theorem does not assume a random event in its hypotheses.
\end{remark}

\begin{corollary}[Federated projector operator-norm bound]\label{cor:fed-op}
Under the assumptions of Theorem~\ref{thm:fed-rowwise}, with probability at least $1-\delta$,
\[
\|\widehat P-P\|+
\max_{1\le j\le p}h_j^{-1/2}\|(\widehat P-P)e_j\|_2
\le C\varepsilon_{n,M}(\delta).
\]
\end{corollary}

\subsubsection{From ratios to topics}\label{subsec:topic-error}
The loss $L(\hat A,A)$ is defined in \eqref{eq:L-loss-def}; all topic-matrix bounds below are understood after the minimizing common permutation of topic labels.

We use the pooled marginal deviation $\Delta_1(D,D_0)$ defined in \eqref{eq:Delta1_def}. Define
\begin{align}
\Delta_2&:=\max_{1\le j\le p}h_j^{-1/2}
\Bigl\|e_j^\top\Bigl([\hat\xi_1,\hat\Xi_\perp O_\perp]-[\xi_1,\Xi_\perp]\Bigr)\Bigr\|_2,\label{eq:Delta2_def}\\
\Err_{\mathrm{VH}}(O_\perp)&:=\min_{\kappa\in S_K}\max_{1\le k\le K}\|O_\perp^\top\hat v_k-v_{\kappa(k)}^\star\|_2.\label{eq:ErrVH_def}
\end{align}

\begin{condition}[Rotation-equivariant vertex-hunting stability]\label{cond:vh}
Suppose the population ratio cloud lies in a nondegenerate simplex with vertices $v_1^\star,\dots,v_K^\star$. If there exists an orthogonal matrix $O\in\R^{(K-1)\times(K-1)}$ and a perturbation radius $\varepsilon_{\mathrm{VH}}\le c_{\mathrm{VH}}$, for a sufficiently small constant depending only on the oracle simplex geometry and fixed $K$, such that the observed ratio points $\hat r_1,\dots,\hat r_p$ obey
\[
\max_{1\le j\le p}\|O^\top\hat r_j-r_j\|_2\le \varepsilon_{\mathrm{VH}},
\]
then the vertex-hunting routine returns vertices $\hat v_1,\dots,\hat v_K$ such that, up to a permutation of $[K]$,
\[
\max_{1\le k\le K}\|O^\top\hat v_k-v_k^\star\|_2\le C_{\mathrm{VH}}\varepsilon_{\mathrm{VH}}.
\]
Euclidean vertex-hunting methods such as successive projection and simplex vertex search satisfy this interface when their non-rotated stability theorem holds, because distances, affine independence, and simplex volumes are invariant under a common orthogonal transformation. Under Assumption~\ref{assump:anchor} and the oracle nondegeneracy in Lemma~\ref{lem:oracle-V-geometry}, this condition is the standard Topic-SCORE vertex-hunting stability requirement.

\end{condition}

\begin{lemma}[A deterministic margin for augmented simplex projection]
\label{lem:augmented-sp-margin}
Let $Q=[q_1,\ldots,q_K]\in\R^{K\times K}$ be the augmented
vertex matrix with $q_k=(1,(v_k^\star)^\top)^\top$.
Assume $\sigma_{\min}(Q)\ge c_Q>0$ and
$\max_k\|q_k\|_2\le C_Q$.
For any $S\subsetneq[K]$, let $\Pi_S$ be the orthogonal
projector onto $\mathrm{span}\{q_k:k\in S\}$, with
$\Pi_\varnothing=0$, and set
\[
 f_S(y):=\|(I-\Pi_S)y\|_2,
 \qquad
 R_S:=\max_{k\notin S}f_S(q_k),
 \qquad
 V_S:=\{q_k:k\notin S\}.
\]
Then, for every $y=Q\pi$ with $\pi\in\Delta^{K-1}$,
\[
 \mathrm{dist}(y,V_S)
 \le
 \frac{4C_Q^2}{c_Q^2}\bigl\{R_S-f_S(y)\bigr\}.
\]
In particular, for every $t\ge0$,
\[
 R_S-f_S(y)\le t
 \quad\Longrightarrow\quad
 \mathrm{dist}(y,V_S)\le Ct,
 \qquad C=\frac{4C_Q^2}{c_Q^2}.
\]
\end{lemma}

\begin{proof}
Fix $S\subsetneq[K]$, write $J=[K]\setminus S$, and set
$r_k=(I-\Pi_S)q_k$ for all $k\in[K]$.
Thus $r_k=0$ for $k\in S$.
For any $u\in\R^{|J|}$, the singular-value assumption gives
\[
 \|(I-\Pi_S)Q_Ju\|_2
 =
 \min_{v\in\R^{|S|}}\|Q_Ju-Q_Sv\|_2
 \ge c_Q\|u\|_2.
\]
Consequently,
\[
 \|r_k\|_2\ge c_Q \quad(k\in J),
 \qquad
 \|r_i-r_j\|_2\ge\sqrt{2}\,c_Q
 \quad(i,j\in J,\ i\ne j).
\]
In particular, $c_Q\le R_S\le C_Q$.

Fix $\pi\in\Delta^{K-1}$ and define
\[
 a:=\sum_{k\in S}\pi_k,
 \qquad
 s:=\sum_{k\in J}\pi_k=1-a,
 \qquad
 m:=\max_{k\in J}\pi_k.
\]
By convexity,
\[
 f_S(y)=\left\|\sum_{k=1}^K\pi_kr_k\right\|_2\le R_S.
\]
The squared-norm variance identity yields
\begin{align*}
 d
 &:=
 R_S^2-\left\|\sum_{k=1}^K\pi_kr_k\right\|_2^2\\
 &=
 \sum_{k=1}^K\pi_k\bigl(R_S^2-\|r_k\|_2^2\bigr)
 +
 \frac12\sum_{i,j=1}^K
 \pi_i\pi_j\|r_i-r_j\|_2^2\\
 &\ge
 aR_S^2+c_Q^2as
 +c_Q^2\left(s^2-\sum_{k\in J}\pi_k^2\right)\\
 &=
 aR_S^2+c_Q^2\left(s-\sum_{k\in J}\pi_k^2\right)\\
 &\ge
 c_Q^2\left(1-\sum_{k\in J}\pi_k^2\right)\\
 &\ge c_Q^2(1-m).
\end{align*}
The first inequality uses the contributions from $S$ in the
first sum, the $S$--$J$ cross terms, and the pairs of distinct
indices in $J$.
The last inequality follows from
$\sum_{k\in J}\pi_k^2\le ms\le m$.

Choose $k_*\in J$ with $\pi_{k_*}=m$.
Since $\|q_k-q_{k_*}\|_2\le2C_Q$, we obtain
\begin{align*}
 \mathrm{dist}(y,V_S)
 &\le \|y-q_{k_*}\|_2\\
 &=
 \left\|\sum_{k\ne k_*}\pi_k(q_k-q_{k_*})\right\|_2\\
 &\le 2C_Q(1-m)\\
 &\le \frac{2C_Q}{c_Q^2}d.
\end{align*}
Finally,
\[
 d
 =
 \bigl\{R_S-f_S(y)\bigr\}
 \bigl\{R_S+f_S(y)\bigr\}
 \le
 2C_Q\bigl\{R_S-f_S(y)\bigr\}.
\]
Combining the last two displays proves the claim.
\end{proof}

\begin{proposition}[A sufficient condition for successive projection]\label{prop:sp-vh}
Suppose the oracle ratio cloud is contained in a nondegenerate simplex $S(v_1^\star,\ldots,v_K^\star)$ and, for each vertex $v_k^\star$, at least one oracle point is exactly equal to that vertex. Let $\sigma_{\min}$ denote the smallest singular value of the augmented vertex matrix
\[
Q=\begin{pmatrix}1&\cdots&1\\ v_1^\star&\cdots&v_K^\star\end{pmatrix},
\]
and assume $\sigma_{\min}\ge c_Q>0$ and $\max_k\|v_k^\star\|\le C_Q$. If the observed points satisfy
\[
\max_j\|\hat r_j-r_j\|_2\le \delta
\]
with $\delta\le c(c_Q,C_Q,K)$, then the augmented successive-projection routine---that is, successive projection applied to
\[
\hat y_j=(1,\hat r_j^\top)^\top\in\R^K
\]
and returning the last $K-1$ coordinates of the selected augmented points---returns vertices $\hat v_1,\ldots,\hat v_K$ such that, after a permutation,
\[
\max_k\|\hat v_k-v_k^\star\|_2\le C(c_Q,C_Q,K)\delta.
\]
Consequently successive projection satisfies Condition~\ref{cond:vh}; the same conclusion after a common orthogonal rotation follows by applying the proposition to $O^\top\hat r_j$.
\end{proposition}

\begin{proof}
We give the deterministic perturbation argument in the standard robust-SP form.  The routine analyzed here is the augmented SP routine stated above; ordinary SP applied directly to the unaugmented ratio points is not the object used in this proposition.  Work with augmented points
\[
 y_j^\star=\binom{1}{r_j},\qquad \hat y_j=\binom{1}{\hat r_j},\qquad q_k=\binom{1}{v_k^\star}.
\]
Then \(y_j^\star=Q\pi_j\) with \(\pi_j\in\Delta^{K-1}\), and for every \(k\) there is an anchor index \(j_k\) such that \(y_{j_k}^\star=q_k\).  The assumptions imply that every submatrix of \(Q\) formed by a subset of columns is uniformly well-conditioned and that \(\|q_k\|\le C\).

For \(S\subset[K]\), let \(\Pi_S\) be the orthogonal projector onto \({\rm span}\{q_k:k\in S\}\), with \(\Pi_\varnothing=0\), and define \(f_S(y)=\|(I-\Pi_S)y\|_2\).  If \(S\ne[K]\), then for every \(y=Q\pi\) in the oracle cloud,
\[
 f_S(y)=\left\|(I-\Pi_S)\sum_{k\notin S}\pi_kq_k\right\|_2
 \le \max_{k\notin S} f_S(q_k)=:R_S,
\]
by convexity of the norm.  Also \(R_S\ge c>0\), since otherwise a column \(q_k\), \(k\notin S\), would be within distance \(o(1)\) of the span of the selected columns, contradicting the lower singular-value bound for \(Q\). Let $V_S:=\{q_k:k\notin S\}$ denote the set of unselected
oracle vertices.

By Lemma~\ref{lem:augmented-sp-margin}, for every $t\ge0$,
\begin{equation}\label{eq:sp-margin}
 R_S-f_S(y)\le t
 \quad\Longrightarrow\quad
 \mathrm{dist}(y,V_S)\le Ct,
\end{equation}
where $C$ depends only on the lower singular-value bound
and the upper bound on the augmented vertex norms.

We now prove the SP guarantee by induction.  Suppose that after \(t\) steps the selected noisy augmented columns are within \(C_t\delta\) of distinct oracle vertices indexed by a set \(S_t\) of size \(t\).  Uniform conditioning of the selected oracle columns implies that the projector \(\widehat\Pi_t\) onto the span of the selected noisy columns satisfies
\begin{equation}\label{eq:sp-proj-pert}
\|\widehat\Pi_t-\Pi_{S_t}\|\le C_t\delta .
\end{equation}
Hence, uniformly over all observed points,
\[
\big|\,\|(I-\widehat\Pi_t)\hat y_j\|_2-f_{S_t}(y_j^\star)\,\big|
\le \|\hat y_j-y_j^\star\|_2+\|\widehat\Pi_t-\Pi_{S_t}\|\,\|y_j^\star\|_2
\le C_t\delta .
\]
Since an anchor point for every remaining vertex is present, a point attaining \(R_{S_t}\) exists in the population cloud.  The point selected by SP at step \(t+1\) therefore satisfies
\[
 f_{S_t}(y_{j_{t+1}}^\star)\ge R_{S_t}-C_t\delta .
\]
By the margin implication \eqref{eq:sp-margin}, \(y_{j_{t+1}}^\star\) lies within \(C_t\delta\) of a new vertex \(q_{k_{t+1}}\) with \(k_{t+1}\notin S_t\).  Adding the observation error gives \(\|\hat y_{j_{t+1}}-q_{k_{t+1}}\|\le C_{t+1}\delta\).  This closes the induction; the base case is the same argument with \(S_0=\varnothing\).

After \(K\) steps, the selected augmented noisy points can be matched to \(q_1,\ldots,q_K\) within \(C\delta\).  Dropping the first coordinate gives the displayed vertex bound.  The argument uses only Euclidean distances and orthogonal projections, so applying it to the rotated cloud \(O^\top\hat r_j\) proves the rotation-equivariant form.
\end{proof}

\begin{remark}[Concrete vertex-hunting routines]\label{rem:vh-concrete}
Condition~\ref{cond:vh} is the same stability interface used in the single-site Topic-SCORE proof. The successive-projection routine satisfies this interface under the anchor-word condition and a nondegenerate simplex, and sketched vertex search satisfies it under the standard cluster-denoising regularity conditions; see the vertex-hunting appendix of the single-site Topic-SCORE analysis. The present topic-matrix theorem is conditional on this vertex-hunting stability: the federated contribution is the projector-to-canonical-basis step before vertex hunting, while the final simplex estimation module can be replaced by any routine satisfying Condition~\ref{cond:vh}.
\end{remark}

\begin{lemma}[Deterministic Topic-SCORE propagation]\label{lem:nonstochastic-fedA}
Assume the oracle simplex is nondegenerate. If
\[
\Delta_1(D,D_0)\le c,
\qquad
\Delta_2\le c,
\qquad
\Err_{\mathrm{VH}}(O_\perp)\le c,
\]
for a sufficiently small constant $c>0$, then, up to a common permutation of topic labels,
\begin{equation}\label{eq:nonstochastic_row_final}
\max_{1\le j\le p}\frac{\|\hat a_j-a_j\|_1}{h_j}
\le C\{\Delta_1(D,D_0)+\Delta_2+\Err_{\mathrm{VH}}(O_\perp)\}.
\end{equation}
Consequently,
\[
L(\hat A_{\mathrm{fed}},A)
\le C\{\Delta_1(D,D_0)+\Delta_2+\Err_{\mathrm{VH}}(O_\perp)\}.
\]
\end{lemma}

\begin{theorem}[Federated Topic-SCORE topic-matrix error, conditional on rotation-equivariant vertex hunting]\label{thm:fed-A}
Assume Assumption~\ref{assump:anchor}, the conditions of Theorem~\ref{thm:fed-rowwise}, and suppose the vertex-hunting routine satisfies the rotation-equivariant stability interface in Condition~\ref{cond:vh}. Choose the small constant $c_*$ in Theorem~\ref{thm:fed-rowwise} sufficiently small that the ratio perturbation bound $C\varepsilon_{n,M}(\delta)$ is below the vertex-hunting threshold $c_{\mathrm{VH}}$. In particular, successive projection satisfies this interface under the sufficient conditions in Proposition~\ref{prop:sp-vh}. Then, for every $\delta\in(0,1)$, with probability at least $1-\delta$, up to a common permutation of topic labels,
\begin{equation}\label{eq:fed-topic-error-row}
\|\hat a_j-a_j\|_1\le C h_j\varepsilon_{n,M}(\delta),
\qquad 1\le j\le p,
\end{equation}
and
\begin{equation}\label{eq:fed-topic-error-global}
L(\hat A_{\mathrm{fed}},A)\le C\varepsilon_{n,M}(\delta).
\end{equation}
In particular, after using the pooled-marginal concentration bound for $\Delta_1(D,D_0)$ and the clientwise Gram-column bound from Lemma~\ref{lem:Ecol_hp}, the shared-whitening bridge residues are
\[
\eta_{n,M}^{\mathrm{marg}}(\delta)\{1+\vartheta_n(\delta)+\zeta_N^{\mathrm{svd}}\}
\qquad\text{and}\qquad
\sqrt p\,\{\eta_{n,M}^{\mathrm{marg}}(\delta)\}^2,
\]
the practical-SVD diagonal-bias residue is $\zeta_N^{\mathrm{svd}}$, and the conservative row-wise Bernstein range residue is $\sqrt p\,\ell_{n,M}(\delta)/M$. Thus, when the SVD-bias and bridge terms are lower order, the dominant global rate delivered by the row-wise theorem is
\[
L(\hat A_{\mathrm{fed}},A)
\lesssim
\beta_n\sqrt{\frac{p\ell_{n,M}(\delta)}{NnM}}
+\beta_n^2\frac{p\ell_{n,M}(\delta)}{Nn}
+\frac{\sqrt p\,\ell_{n,M}(\delta)}{M},
\]
up to the displayed lower-order bridge terms, the practical-SVD diagonal-bias term, and the fixed-whitening quadratic bias term. If an additional localized projector-column range inequality is imposed or proved, the final range term can be replaced by $\ell_{n,M}(\delta)/M$.
\end{theorem}

\subsubsection{Regime comparison and the three-term balance}\label{subsec:three-term-balance}

\begin{corollary}[Regime comparison with the Bernstein range term]\label{cor:regimes}
Under the rate simplification after Theorem~\ref{thm:fed-A}, set $\delta=(nM)^{-3}$ and
\[
\ell_{n,M}:=\ell_{n,M}(\delta)=\log(2pn^3M^4).
\]
Define
\[
T_{\mathrm{var}}:=\beta_n\sqrt{\frac{p\ell_{n,M}}{NnM}},
\qquad
T_{\mathrm{bias}}:=\beta_n^2\frac{p\ell_{n,M}}{Nn},
\qquad
T_{\mathrm{range}}:=\frac{\sqrt p\,\ell_{n,M}}{M},
\]
and
\[
T_{\mathrm{br,row}}:=\eta_{n,M}^{\mathrm{marg}}(\delta)\{1+\vartheta_n(\delta)+\zeta_N^{\mathrm{svd}}\},
\qquad
T_{\mathrm{br,quad}}:=\sqrt p\,\{\eta_{n,M}^{\mathrm{marg}}(\delta)\}^2,
\qquad
T_{\mathrm{svd}}:=\zeta_N^{\mathrm{svd}}.
\]
Then, up to the fixed-whitening quadratic lower-order term $(\beta_n^2p/(Nn))^2$,
\begin{equation}\label{eq:cor-start}
L(\hat A_{\mathrm{fed}},A)\lesssim T_{\mathrm{var}}+T_{\mathrm{bias}}+T_{\mathrm{range}}+T_{\mathrm{br,row}}+T_{\mathrm{br,quad}}+T_{\mathrm{svd}}.
\end{equation}
The first three terms are the main variance/local-bias/range contributions. The extra bridge terms $T_{\mathrm{br,row}}$ and $T_{\mathrm{br,quad}}$ are inherited from the shared empirical whitening matrix, while $T_{\mathrm{svd}}$ is the diagonal-bias price of the practical ordinary-SVD upload. In regimes where these residues are lower order, the qualitative balance is governed by the three main terms.

More explicitly, the variance-bias crossover solves $T_{\mathrm{var}}\asymp T_{\mathrm{bias}}$, giving
\[
M\asymp \frac{Nn}{\beta_n^2p\ell_{n,M}}.
\]
The conservative row-wise Bernstein range term is no larger than the bias term once
\[
M\gtrsim \frac{Nn}{\beta_n^2\sqrt p},
\]
and is no larger than the variance term once
\[
M\gtrsim \frac{\ell_{n,M}Nn}{\beta_n^2}.
\]
Thus, unlike the operator-norm range residue, the unconditional row-wise range residue can be the limiting finite-sample term unless $M$ is fairly large. A sharper localized projector-column range inequality would replace $\sqrt p\,\ell_{n,M}/M$ by $\ell_{n,M}/M$ and recover the milder balance in the operator-norm theorem.
\end{corollary}

\begin{remark}[Why the range term remains in the statement]
The familiar two-term decomposition $T_{\mathrm{var}}+T_{\mathrm{bias}}$ (called $T_{\rm var}+T_{\rm loc}$ in the main text) is recovered only when $T_{\mathrm{range}}$, $T_{\mathrm{br,row}}$, $T_{\mathrm{br,quad}}$, and $T_{\mathrm{svd}}$ are lower order. In finite samples, the row-wise projector-average Bernstein inequality contributes the range term, the shared empirical whitening contributes the two bridge residues, and practical SVD upload contributes $T_{\mathrm{svd}}$. The cleanest nonasymptotic statement keeps all of them visible. The displayed row-wise theorem is most informative in many-client regimes; when $M$ is small, a direct pooled or single-site row-wise analysis can give sharper constants and avoid interpreting the conservative $\sqrt p/M$ range residue as a statistical limitation.
\end{remark}

\subsubsection{Heterogeneity-sensitive interpretation}\label{subsec:upturn}

Define the local signal gaps
\[
\rho_m:=\lambda_K(G_{0,m}),
\qquad 1\le m\le M.
\]
Since
\[
G_{0,m}=\Bigl(1-\frac1N\Bigr)B\Sigma_{W,m}B^\top,
\qquad B=M_0^{-1/2}A,
\]
we have
\begin{equation}\label{eq:rho-lower-upturn}
\rho_m\ge \Bigl(1-\frac1N\Bigr)\lambda_{\min}(B^\top B)\lambda_{\min}(\Sigma_{W,m}).
\end{equation}
Thus extreme client imbalance can shrink local gaps even when the pooled matrix $\Sigma_W$ is well behaved.

\begin{proposition}[Heterogeneity-sensitive fixed-whitening projector bound for ordinary SVD]\label{prop:upturn-gap}
Let
\[
\bar P^{(0)}:=\sum_{m=1}^M\alpha_m\tilde P_m,
\qquad
\rho_m:=\lambda_K(G_{0,m}).
\]
Assume $\rho_m>0$ for every $m$ and deterministic weights
$\alpha_m\ge0$ with $\sum_m\alpha_m=1$.
For the identity-centered fixed-whitening matrix, decompose
\[
\tilde G_m-G_{0,m}=F_m+B_m,
\qquad
F_m:=\tilde G_m-\E[\tilde G_m\mid W^{(1:M)}],
\qquad
B_m:=D_{0,m}^{\mathrm{svd}},
\]
so that \(\E[F_m\mid W^{(1:M)}]=0\). Define
\[
v_m:=\E\|F_m\|^2,
\qquad
b_m:=\|B_m\|.
\]
Let \(\widehat P^{(0)}\) denote the top-\(K\) spectral projector of \(\bar P^{(0)}\) whenever it is well defined, and let \(\hat\Xi^{(0)}\) be any orthonormal basis of \(\mathrm{range}(\widehat P^{(0)})\). Assume that the deterministic ordinary-SVD diagonal bias is locally perturbative, \(b_m\le \rho_m/8\) for all \(m\). For \(\delta\in(0,1)\) define
\[
\mathcal E_{\mathrm{loc}}(\delta):=\bigcap_{m=1}^M\{\|F_m\|\le \rho_m/8\}.
\]
Assume \(\Pbb(\mathcal E_{\mathrm{loc}}(\delta/2))\ge 1-\delta/2\). Then, with probability at least \(1-\delta\), on \(\mathcal E_{\mathrm{loc}}(\delta/2)\) and the unconditional averaging event supplied by Theorem~\ref{thm:projector-avg},
\begin{equation}\label{eq:upturn-main-bound}\makeatletter\let\camera@tagform\tagform@\def\tagform@#1{\raisebox{-3pt}[\height][\depth]{\camera@tagform{#1}}}\makeatother
\|\bar P^{(0)}-P\|
\le C\Biggl[
\left[\log\!\frac{4p}{\delta}\sum_{m=1}^M\alpha_m^2\frac{v_m+b_m^2}{\rho_m^2}\right]^{1/2}
+
\sum_{m=1}^M\alpha_m\left\{\frac{v_m}{\rho_m^2}+\frac{b_m}{\rho_m}+\frac{b_m^2}{\rho_m^2}\right\}
+
\log\!\frac{4p}{\delta}\max_m\alpha_m
\Biggr].
\end{equation}
If the right-hand side is at most \(1/4\), the extracted top-\(K\) subspace obeys the same bound up to an absolute constant; equivalently,
\begin{equation}\label{eq:upturn-main-bound-subspace}
\|\sinTheta(\hat\Xi^{(0)},\Xi)\|
\le C\Biggl[
\left[\log\!\frac{4p}{\delta}\sum_{m=1}^M\alpha_m^2\frac{v_m+b_m^2}{\rho_m^2}\right]^{1/2}
+
\sum_{m=1}^M\alpha_m\left\{\frac{v_m}{\rho_m^2}+\frac{b_m}{\rho_m}+\frac{b_m^2}{\rho_m^2}\right\}
+
\log\!\frac{4p}{\delta}\max_m\alpha_m
\Biggr].
\end{equation}
\end{proposition}

\begin{corollary}[Recovery of the regular regime]\label{cor:upturn-regular}
Assume the conditions of Proposition~\ref{prop:upturn-gap}, \(\alpha_m=1/M\),
\[
v_m\le C\beta_n^2\frac p{Nn},
\qquad
b_m\le C\zeta_N^{\mathrm{svd}},
\qquad
\zeta_N^{\mathrm{svd}}\le c\rho_*,
\qquad
\rho_m\ge \rho_*>0,
\qquad 1\le m\le M.
\]
Then, on the same event,
\begin{equation}\label{eq:upturn-regular-regime}\makeatletter\let\camera@tagform\tagform@\def\tagform@#1{\raisebox{-3pt}[\height][\depth]{\camera@tagform{#1}}}\makeatother
\|\sinTheta(\hat\Xi^{(0)},\Xi)\|
\le C\Biggl[
\frac{\beta_n}{\rho_*}\sqrt{\frac{p\log(4p/\delta)}{NnM}}
+
\frac{\zeta_N^{\mathrm{svd}}}{\rho_*}\sqrt{\frac{\log(4p/\delta)}{M}}
+
\frac{\beta_n^2}{\rho_*^2}\frac p{Nn}
+
\frac{\zeta_N^{\mathrm{svd}}}{\rho_*}
+
\frac{\log(4p/\delta)}{M}
\Biggr].
\end{equation}
\end{corollary}
\begin{remark}[Gap-explicit diagnostic for imbalance]
The mild upward bend sometimes seen in the federated curve is not caused by a failure of the anchor-word/simplex geometry. It comes from the local matrices $\Sigma_{W,m}$: if some of them are nearly singular, then the corresponding local gaps $\rho_m$ are small, and local spectral bias can dominate the variance reduction from averaging. This is a diagnostic statement outside the uniform regular theorem class when the gaps approach zero; the main consistency theorem assumes the regular regime $\lambda_{\min}(\Sigma_{W,m})\ge c_{\mathrm{loc}}$.
\end{remark}

\subsubsection{A lower bound}\label{subsec:lower}

\begin{theorem}[Minimax lower bound on the matched theorem class]\label{thm:lower}
Fix $K\ge2$. Assume $n_m=n$, $n\ge K$, and $p\ge p_0(K)$, where
$p_0(K)$ is a sufficiently large constant, and
\[
\log(nM)\le\min\{p,N\},
\qquad
p\log(nM)=o(NnM).
\]
There exists a constant $c_{\star}(K)\in(0,1/(2K)]$ such that the following holds for any fixed
$c_1,c_2,c_{\mathrm{loc}}\in(0,c_{\star}(K)]$.

Let $\Phi_{n,N,p}^{(M)}(K,c_1,c_2,c_{\mathrm{loc}})$ be the collection of parameters $(A,W^{(1)},\dots,W^{(M)})$ such that: each topic has at least one anchor word; $h_{\min}\ge c_1K/p$; $\lambda_{\min}(\Sigma_W)\ge c_2$, $\lambda_{\min}(\Sigma_A)\ge c_2$, and $\lambda_{\min}(\Sigma_{W,m})\ge c_{\mathrm{loc}}$ for every client $m$. Then there exist constants $c_0,\delta_0>0$ such that, along any sequence $(n,M,N,p)$ satisfying these conditions and for all sufficiently large problem sizes,
\[
\inf_{\hat A}\sup_{(A,W^{(1:M)})\in\Phi_{n,N,p}^{(M)}(K,c_1,c_2,c_{\mathrm{loc}})}
\Pbb\Bigl(L(\hat A,A)\ge c_0\sqrt{\frac p{NnM}}\Bigr)
\ge\delta_0.
\]
\end{theorem}
The least favorable subclass used in the proof is chosen so that every
client already satisfies local richness. Each client contains the same
balanced collection of pure-topic documents, with either
$\lfloor n/K\rfloor$ or $\lceil n/K\rceil$ documents at each topic
weight $e_k$. Consequently,
$\lambda_{\min}(\Sigma_{W,m})\ge1/(2K)$ for all $m$; when $K$ divides
$n$, this reduces to $\Sigma_{W,m}=K^{-1}I_K$.
The packing therefore lies inside the same theorem class used by the
upper bounds.

\begin{remark}[Interpretation of the lower bound]
\label{rem:lower-matches-leading}
Theorem~\ref{thm:lower} gives a minimax lower bound of order
$\sqrt{p/(NnM)}$ on the stated parameter class. The variance
contribution in Theorem~\ref{thm:fed-A} has the same dependence
on $p,N,n,M$, up to logarithmic and $\beta_n$ factors. The complete
finite-sample upper bound also includes the local spectral bias,
row-wise Bernstein range, shared-whitening, and ordinary-SVD
contributions.

With $\ell=\ell_{n,M}(\delta)$, the conservative terms displayed
in the main text satisfy
\[
T_{\rm loc}T_{\rm row}
=\sqrt p\,\ell\,T_{\rm var}^{\,2}.
\]
The comparison with the lower bound concerns the variance
component and does not assert a matching minimax rate for the
complete upper bound; every term
in~\eqref{eq:main-radius} is retained.
\end{remark}

\subsubsection{Auxiliary estimation of document weights and $K$}\label{subsec:WK}

\begin{theorem}[Document-level topic weights via sample splitting]\label{thm:fed-W}
Split the corpus into a training part and a held-out document $d_i^{\mathrm{te}}$. Suppose each client contributes $n_{\mathrm{tr}}$ training documents, and set $n_{\mathrm{tot}}^{\mathrm{tr}}:=Mn_{\mathrm{tr}}$. Construct $\hat A_{\mathrm{fed}}^{\mathrm{tr}}$ and $M_d^{\mathrm{tr}}$ from the training split only. Define the training-split population and empirical pooled marginals by
\[
D_0^{\mathrm{tr}}:=AW^{\mathrm{tr}},\qquad
M_0^{\mathrm{tr}}:=\diag\!\left((n_{\mathrm{tot}}^{\mathrm{tr}})^{-1}D_0^{\mathrm{tr}}\1\right),\qquad
M_d^{\mathrm{tr}}:=\diag\!\left((n_{\mathrm{tot}}^{\mathrm{tr}})^{-1}D^{\mathrm{tr}}\1\right),
\]
and define
\[
\Delta_1^{\mathrm{tr}}
:=
\max_{1\le j\le p}h_j^{-1}
\bigl|M_d^{\mathrm{tr}}(j,j)-M_0^{\mathrm{tr}}(j,j)\bigr|.
\]
Assume that all hypotheses of Theorem~\ref{thm:fed-A}, including
the anchor-word, minimum-row-mass, pooled and client-level richness,
vertex-hunting stability, and perturbative conditions, hold on the
training split with $n$ replaced by $n_{\mathrm{tr}}$.
For the error statements, relabel the training topic estimate by the common
permutation supplied by that theorem and express the estimated weights
in the same topic order; this permutation depends only on the training data.
Define
\[
\hat w_i^\star:=\arg\min_{b\in\R^K}\sum_{j=1}^p\frac1{M_d^{\mathrm{tr}}(j,j)}\Bigl(d_i^{\mathrm{te}}(j)-\sum_{k=1}^Kb(k)\hat A_{\mathrm{fed}}^{\mathrm{tr}}(j,k)\Bigr)^2.
\]
Truncate negative coordinates and renormalize to the simplex to obtain $\hat w_i$. Let $\hat a_j^{\mathrm{tr}}$ denote the $j$th row of $\hat A_{\mathrm{fed}}^{\mathrm{tr}}$, let $\varepsilon_{n_{\mathrm{tr}},M}^{\mathrm{tr}}(\delta_{\mathrm{tr}})$ denote the deterministic radius from Theorems~\ref{thm:fed-rowwise} and~\ref{thm:fed-A} applied to the training split, and define the training event
\[
\mathcal E_{\mathrm{tr}}(\delta_{\mathrm{tr}}):=
\left\{
\max_{1\le j\le p}\frac{\|\hat a_j^{\mathrm{tr}}-a_j\|_1}{h_j}\le C\varepsilon_{n_{\mathrm{tr}},M}^{\mathrm{tr}}(\delta_{\mathrm{tr}}),
\qquad
\Delta_1^{\mathrm{tr}}\le c
\right\},
\]
for a sufficiently small deterministic constant $c>0$. Conditional on the training split, on $\mathcal E_{\mathrm{tr}}(\delta_{\mathrm{tr}})$,
\[
\|\hat w_i-w_i\|_1
\le
2\wedge C\left\{\varepsilon_{n_{\mathrm{tr}},M}^{\mathrm{tr}}(\delta_{\mathrm{tr}})+\sqrt{\frac{\log(2K/\delta_{\mathrm{te}})}{N}}\right\}
\]
with probability at least $1-\delta_{\mathrm{te}}$ over the held-out document. Moreover, if $\mathcal E_{\mathrm{tr}}(\delta_{\mathrm{tr}})$ holds with probability at least $1-\delta_{\mathrm{tr}}$ over the training split, then the same bound holds unconditionally with probability at least $1-\delta_{\mathrm{tr}}-\delta_{\mathrm{te}}$. The event $\mathcal E_{\mathrm{tr}}(\delta_{\mathrm{tr}})$ is supplied by Theorem~\ref{thm:fed-A} together with the pooled-marginal concentration lemma on the training split under the training-split regularity conditions just stated.
\end{theorem}

This training/held-out module is auxiliary to the projector-only core:
one first estimates $\hat A_{\mathrm{fed}}^{\mathrm{tr}}$ and
$M_d^{\mathrm{tr}}$ on the training split and then estimates the held-out $w_i$.
For cross-fitting with a fixed number of folds, evaluate each document only
using fits trained without that document. If several such predictions are
averaged, first align topic columns to a reference fit by minimum total
columnwise $\ell_1$ matching. This matching is correct whenever every fit has
maximum column error at most $e<c_2/2$ after its oracle permutation: for
$k\ne l$, Assumption~\ref{assump:Sigma} gives
\[
\|A_{\cdot k}-A_{\cdot l}\|_1
\ge (e_k-e_l)^\top\Sigma_A(e_k-e_l)\ge2c_2,
\]
so matched columns are at distance at most $2e$ and mismatched columns
are at distance at least $2c_2-2e>2e$.
For simultaneous guarantees, allocate the desired failure probability
across all training events and evaluated held-out documents.

\begin{theorem}[Optional pilot-stage rule for $K$; outside the projector-only core]\label{thm:fed-K}
Suppose each client forms an auxiliary symmetric pilot sketch $\hat G_m^{\mathrm{pil}}$ on an independent pilot pass, with
\[
\E[\hat G_m^{\mathrm{pil}}\mid W^{(1:M)}]=G_{0,m},
\]
and the server averages
\[
G_{\mathrm{pilot}}:=\frac1M\sum_{m=1}^M\hat G_m^{\mathrm{pil}},
\qquad
G_0:=\frac1M\sum_{m=1}^MG_{0,m}.
\]
Assume $\lambda_K(G_0)\ge \rho_*>0$ and $\lambda_{K+1}(G_0)=0$. For any threshold $t\in(0,\rho_*)$, define
\[
\hat K_t^{\mathrm{pil}}:=\#\{k\in[p]:\lambda_k(G_{\mathrm{pilot}})\ge t\}.
\]
If
\[
\Pbb\{\|G_{\mathrm{pilot}}-G_0\|<\min(t,\rho_*-t)\}\ge 1-o((nM)^{-3}),
\]
then $\hat K_t^{\mathrm{pil}}=K$ with probability $1-o((nM)^{-3})$. For a pilot sketch and threshold satisfying these conditions, one may rerun Algorithm~\ref{alg:fed-topic-score} with $K=\hat K_t^{\mathrm{pil}}$.
\end{theorem}

\subsubsection{Positioning relative to single-site Topic-SCORE and federated subspace averaging}\label{subsec:positioning}

The federated analysis differs from single-site SVD Topic-SCORE in one essential respect. In the single-site proof, a global normalized corpus matrix has a Perron-selected first singular direction, and the remaining analysis proceeds by SCORE ratios, simplex recovery, and column normalization. Here the client-level matrices share only the column space $\mathcal S=\col(M_0^{-1/2}A)$; their eigenbases depend on the local second moments $\Sigma_{W,m}$. Projector averaging is therefore used only to recover the common subspace. The pooled marginal then supplies the missing positive direction $\xi_1=M_0^{1/2}\1$, after which the deterministic Topic-SCORE geometry is the same as in the single-site Topic-SCORE analysis.

The projector-averaging step is also related to distributed principal-subspace estimation, but the statistical target and the loss are different. Distributed PCA analyses typically estimate a principal eigenspace in operator or Frobenius norm. The present problem additionally requires row-wise control, a canonical direction inside the averaged subspace, multinomial centering, and propagation through a simplex map to an $\ell_1$ topic-matrix loss. This is why the proof keeps separate the operator variance term, the row-wise Bernstein range term, the local second-order bias, and the shared-whitening bridge terms.

\subsubsection{Proof roadmap}\label{subsec:proof-roadmap}

The final proof chain is
\[
\tilde G_m-G_{0,m}\to \tilde P_m-P\to \bar P^{(0)}-P\to \bar P-P\to \widehat P-P\to (\hat\xi_1,\hat\Xi_\perp)\to \hat R\to \hat\Pi\to \hat A.
\]
The first arrow is the local identity-centered SVD perturbation, including the deterministic diagonal bias \(D_{0,m}^{\mathrm{svd}}\). The second is the variance-bias projector-averaging step for the fixed-whitening projectors. The third is the bridge correction from the shared empirical whitening matrix \(M_d\). The fourth extracts the spectral projector of the averaged matrix. The fifth is the canonical step unique to the federated problem: the projected pooled marginal supplies \(\hat\xi_1\), and the residual basis is aligned only by an \((K-1)\)-dimensional orthogonal matrix. Once the canonical direction is inserted, the ratio, simplex, and topic-matrix propagation mirrors the single-site Topic-SCORE proof.

% =========================================================
\paragraph{Proof organization.} The remaining subsections give the detailed proofs and supporting calculations.

% =========================================================
\subsection{Auxiliary linear algebra}\label{app:lin-alg}

\subsubsection{Proof of Lemma~\ref{lem:Md_bridge_decomp}}
\begin{proof}
For each document,
\[
\E[d_i d_i^\top\mid W]=\left(1-\frac1N\right)\pi_i\pi_i^\top+\frac1N\diag(\pi_i).
\]
Averaging within client \(m\) and pre- and post-multiplying by \(M_0^{-1/2}\) gives
\[
\E[\tilde G_m\mid W^{(1:M)}]
=
G_{0,m}+\frac1N\left(M_0^{-1/2}M_{0,m}M_0^{-1/2}-I_p\right)
=
G_{0,m}+D_{0,m}^{\mathrm{svd}},
\]
which proves \eqref{eq:fixed-mean-svd}. For the bridge, since
\[
M_d^{-1/2}=(I_p+R_M)M_0^{-1/2}
\quad\text{and}\quad
T_m=M_0^{-1/2}\hat S_mM_0^{-1/2},
\]
we have
\[
\hat G_m+\frac1N I_p=(I_p+R_M)T_m(I_p+R_M),
\qquad
\tilde G_m+\frac1N I_p=T_m.
\]
Subtracting the two displays gives
\[
\hat G_m-\tilde G_m=R_MT_m+T_mR_M+R_MT_mR_M.
\]
Combining this bridge identity with the fixed-whitening mean decomposition yields
\eqref{eq:svd-main-decomp}.
\end{proof}

\subsubsection{Proof of Lemma~\ref{lem:dk-projector}}
\begin{proof}
The projector $P$ has eigenvalues $1$ with multiplicity $K$ and $0$ with multiplicity $p-K$. If $\norm{\bar{\mathcal P}-P}\le 1/4$, Weyl's inequality shows that $\bar{\mathcal P}$ has exactly $K$ eigenvalues in $[3/4,5/4]$ and the remaining eigenvalues in $[-1/4,1/4]$. Hence the top-$K$ eigenspace is uniquely defined. Applying Davis--Kahan to the gap-one cluster of $P$ gives
\[
\norm{\hat P-P}=\norm{\sinTheta(\hat\Xi,\Xi)}
\le 2\norm{\bar{\mathcal P}-P}.
\]
The same proof gives a constant depending only on $c_{\mathrm{DK}}$ whenever $\norm{\bar{\mathcal P}-P}\le c_{\mathrm{DK}}<1/2$.
\end{proof}

\subsubsection{Proof of Lemma~\ref{lem:vb-decomp}}
\begin{proof}
The only algebraic point is to use the normalization $\sum_{m=1}^M\alpha_m=1$. Indeed,
\[
\bar{\mathcal P}-P
=
\sum_{m=1}^M\alpha_m\mathcal P_m-
\Bigl(\sum_{m=1}^M\alpha_m\Bigr)P.
\]
Adding and subtracting $\sum_{m=1}^M \alpha_m\E\mathcal P_m$ gives the exact identity
\[
\bar{\mathcal P}-P
=
\sum_{m=1}^M \alpha_m(\mathcal P_m-\E\mathcal P_m)
+
\sum_{m=1}^M \alpha_m(\E\mathcal P_m-P).
\]
Taking operator norms and applying the triangle inequality proves the claim. This lemma itself does not use independence or nonnegativity of the weights. Nonnegativity is used in Lemmas~\ref{lem:bernstein-matrix} and~\ref{lem:vector-bernstein} to write the range terms with $\max_m\alpha_m$ rather than $\max_m|\alpha_m|$.
\end{proof}

\subsubsection{Proof of Lemma~\ref{lem:bernstein-matrix}}
\begin{proof}
Set $A_m:=\mathcal P_m-P$ and $B_m:=\E\mathcal P_m-P=\E[A_m]$. Then
\[
\mathcal P_m-\E\mathcal P_m=A_m-B_m,
\qquad
X_m=\alpha_m(A_m-B_m).
\]
Because $\mathcal P_m$ and $P$ are orthogonal projectors and $\E\mathcal P_m$ is a positive semidefinite contraction, $\|\mathcal P_m\|\le1$ and $\|\E\mathcal P_m\|\le1$. Since the averaging weights are nonnegative,
\[
\|X_m\|\le \alpha_m\|\mathcal P_m-\E\mathcal P_m\|
\le 2\alpha_m .
\]
For the variance proxy, we use the noncommutative variance identity directly. Since $B_m=\E A_m$ is deterministic and self-adjoint,
\[
\begin{aligned}
\E[(\mathcal P_m-\E\mathcal P_m)^2]
&=\E[(A_m-B_m)^2] \\
&=\E[A_m^2]-\E[A_m]B_m-B_m\E[A_m]+B_m^2 \\
&=\E[A_m^2]-B_m^2 .
\end{aligned}
\]
The left-hand side is positive semidefinite, because $(A_m-B_m)^2\succeq0$ pointwise. Since $B_m^2\succeq0$, the identity gives
\[
0\preceq \E[(\mathcal P_m-\E\mathcal P_m)^2]
\preceq \E[A_m^2].
\]
Consequently,
\[
\Bigl\|\E[(\mathcal P_m-\E\mathcal P_m)^2]\Bigr\|
\le \|\E[A_m^2]\|
\le \E\|A_m^2\|
=\E\|A_m\|^2
=\nu_m.
\]
In particular, the weaker bound by $4\nu_m$ used in the statement also holds.
Multiplying by $\alpha_m^2$ and summing over $m$ gives
\[
\Bigl\|\sum_{m=1}^M\E[X_m^2]\Bigr\|
\le
\sum_{m=1}^M\alpha_m^2
\Bigl\|\E[(\mathcal P_m-\E\mathcal P_m)^2]\Bigr\|
\le
4\sum_{m=1}^M\alpha_m^2\nu_m.
\]
Tropp's self-adjoint matrix Bernstein inequality applied to the independent centered matrices
$X_m$ now yields the displayed probability bound.
\end{proof}

\subsubsection{Proof of Lemma~\ref{lem:second-order-bias}}
\begin{proof}
Let $R(z)=(zI-G)^{-1}$. Choose a contour $\mathcal C$ (or a finite
union of contours) enclosing precisely the target cluster and staying at
distance at least $\gamma/2$ from the entire spectrum of $G$.
The cluster-diameter assumption permits total length at most $C(C_0)\gamma$,
and $\sup_{z\in\mathcal C}\|R(z)\|\le2/\gamma$. The Riesz formulas are
\[
P=\frac1{2\pi i}\oint_{\mathcal C}R(z)\,dz,
\qquad
P_m=\frac1{2\pi i}\oint_{\mathcal C}(zI-G-E_m)^{-1}\,dz.
\]
If $\|E_m\|\le \gamma/4$, then $\|E_mR(z)\|\le 1/2$ on $\mathcal C$, so the resolvent admits the Neumann expansion
\[
(zI-G-E_m)^{-1}=R(z)+R(z)E_mR(z)+\sum_{r\ge2}R(z)(E_mR(z))^r.
\]
Subtracting the Riesz formula for $P$ gives
\[
P_m-P=\mathcal L_G(E_m)+\mathcal R_G(E_m),
\qquad
\E[\mathcal L_G(E_m)]=0,
\]
because $\E E_m=0$. The operator remainder satisfies
\[
\|\mathcal R_G(E_m)\|\le \frac{C}{\gamma^2}\|E_m\|^2,
\]
and hence
\[
\|\E P_m-P\|=\|\E\mathcal R_G(E_m)\|\le \frac{C}{\gamma^2}\E\|E_m\|^2.
\]

We now prove the columnwise display under the additional rank-$K$ signal assumption in the lemma. In this case $R(z)(I-P)=z^{-1}(I-P)$ on the contour and $|z|\asymp\gamma$. Hence, for any vector $x$ and any $z\in\mathcal C$,
\[
\|E_mR(z)x\|_2
\le C\gamma^{-1}\{\|E_mx\|_2+\|E_m\|\,\|Px\|_2\}.
\]
Indeed,
\[
E_mR(z)x=E_mR(z)Px+z^{-1}E_m(I-P)x,
\]
and the second term is bounded by $C\gamma^{-1}\{\|E_mx\|_2+\|E_m\|\|Px\|_2\}$.
Using the Neumann-series expression of the remainder,
\[
\mathcal R_G(E_m)e_j
=
\frac1{2\pi i}\oint_{\mathcal C}\sum_{r\ge2}R(z)(E_mR(z))^{r-1}E_mR(z)e_j\,dz,
\]
we therefore obtain, since $\|E_mR(z)\|\le1/2$,
\[
\|\mathcal R_G(E_m)e_j\|_2
\le
\frac{C}{\gamma^2}\,
\|E_m\|\{\|E_me_j\|_2+\|E_m\|\,\|Pe_j\|_2\}.
\]
Taking expectations and applying Cauchy--Schwarz to the first product gives
\[
\| (\E P_m-P)e_j\|_2
\le
\frac{C}{\gamma^2}
\left[
(\E\|E_m\|^2)^{1/2}(\E\|E_me_j\|_2^2)^{1/2}
+\|Pe_j\|_2\,\E\|E_m\|^2
\right],
\]
which proves \eqref{eq:second-order-col}.
\end{proof}

\subsubsection{Projector perturbation expansion}\label{app:proj-expansion}

\begin{lemma}[Linear term plus quadratic remainder]\label{lem:proj-expansion}
Let $G$ be symmetric with a size-$K$ isolated spectral cluster separated from the rest of the spectrum by an eigengap $\gamma>0$, and let $P$ be its associated projector. Assume the target cluster has diameter at most $C_0\gamma$ for a
fixed constant $C_0$. If $\hat G=G+E$ and $\norm{E}\le\gamma/4$, then,
with $C$ depending only on $C_0$,
\[
\hat P-P=\mathcal L_G(E)+\mathcal R_G(E),
\qquad
\norm{\mathcal L_G(E)}\le \frac{C}{\gamma}\norm{E},
\qquad
\norm{\mathcal R_G(E)}\le \frac{C}{\gamma^2}\norm{E}^2.
\]
If, in addition, $G\succeq0$ has rank $K$ and $P$ projects onto its
nonzero eigenspace, then, for each $j\in[p]$,
\begin{equation}\label{eq:proj-col-corrected}
\norm{(\hat P-P)e_j}_2
\le
\frac{C}{\gamma}\{\norm{Ee_j}_2+\norm{E}\norm{Pe_j}_2\}
+\frac{C}{\gamma^2}\norm{E}^2 .
\end{equation}
Under this additional rank-$K$ signal assumption, the quadratic
remainder has the refined column bound
\begin{equation}\label{eq:proj-remainder-col-refined}
\norm{\mathcal R_G(E)e_j}_2
\le
\frac{C}{\gamma^2}\norm{E}\Bigl\{\norm{Ee_j}_2+\norm{E}\norm{Pe_j}_2\Bigr\}.
\end{equation}
In that rank-$K$ signal case, the crude final term in \eqref{eq:proj-col-corrected} can be replaced by the right-hand side of \eqref{eq:proj-remainder-col-refined}.
\end{lemma}

\begin{proof}
Let $R(z):=(zI-G)^{-1}$. Choose a contour $\mathcal C$ (or a finite
union of contours) enclosing precisely the target cluster and staying at
distance at least $\gamma/2$ from the entire spectrum of $G$.
Its total length is at most $C(C_0)\gamma$ and
$\sup_{z\in\mathcal C}\|R(z)\|\le2/\gamma$. The Riesz formulas give
\[
P=\frac1{2\pi i}\oint_{\mathcal C}R(z)\,dz,
\qquad
\hat P=\frac1{2\pi i}\oint_{\mathcal C}(zI-G-E)^{-1}\,dz.
\]
Because $\|E\|\le\gamma/4$, the Neumann expansion is valid on $\mathcal C$:
\[
(zI-G-E)^{-1}=R(z)+R(z)ER(z)+\sum_{r\ge2}R(z)(ER(z))^r.
\]
Thus
\[
\mathcal L_G(E):=\frac1{2\pi i}\oint_{\mathcal C}R(z)ER(z)\,dz,
\qquad
\mathcal R_G(E):=\frac1{2\pi i}\oint_{\mathcal C}\sum_{r\ge2}R(z)(ER(z))^r\,dz.
\]
The operator-norm bounds follow from $\|R(z)\|\le C/\gamma$ and the geometric series.

For the column statements, assume that $G\succeq0$ has rank $K$
and $P$ projects onto its nonzero eigenspace. Put $Q=I-P$.
Because the complementary spectrum is exactly zero, the residue
calculation gives
\[
\mathcal L_G(E)=G^\dagger E Q+QEG^\dagger,
\]
where $G^\dagger$ is the Moore--Penrose inverse and
$\|G^\dagger\|\le\gamma^{-1}$.
Since $G^\dagger=G^\dagger P$, we have
\[
\begin{aligned}
\|\mathcal L_G(E)e_j\|_2
&\le \gamma^{-1}\|EQe_j\|_2
+\gamma^{-1}\|E\|\|Pe_j\|_2\\
&\le \gamma^{-1}
\{\|Ee_j\|_2+2\|E\|\|Pe_j\|_2\}.
\end{aligned}
\]
Combining this bound with
$\|\mathcal R_G(E)e_j\|_2\le C\gamma^{-2}\|E\|^2$
proves \eqref{eq:proj-col-corrected} under the stated rank-$K$
signal assumption.

It remains to prove the refined columnwise remainder in the rank-$K$ signal case. Assume now that $G\succeq0$ has rank $K$ and that $P$ projects onto its nonzero eigenspace. Then the complement spectrum is the single point zero, and on the contour
\[
R(z)(I-P)=z^{-1}(I-P),\qquad |z|^{-1}\le C\gamma^{-1}.
\]
Set
\[
y_z:=\|E(zI-G-E)^{-1}e_j\|_2.
\]
The resolvent identity gives
\[
(zI-G-E)^{-1}=R(z)+R(z)E(zI-G-E)^{-1}.
\]
Therefore
\[
y_z\le \|ER(z)e_j\|_2+\|E\|\|R(z)\|y_z.
\]
Because $\|E\|\|R(z)\|\le c<1$ on the contour, the second term can be absorbed. For the first term, using $R(z)(I-P)=z^{-1}(I-P)$,
\[
\|ER(z)e_j\|_2
\le C\gamma^{-1}\{\|E(I-P)e_j\|_2+\|E\|\|Pe_j\|_2\}
\le C\gamma^{-1}\{\|Ee_j\|_2+\|E\|\|Pe_j\|_2\}.
\]
Thus
\[
y_z\le C\gamma^{-1}\{\|Ee_j\|_2+\|E\|\|Pe_j\|_2\}.
\]
Using the identity
\[
\mathcal R_G(E)=\frac1{2\pi i}\oint_{\mathcal C}R(z)ER(z)E(zI-G-E)^{-1}\,dz,
\]
valid under the same Neumann expansion, we get
\[
\|\mathcal R_G(E)e_j\|_2
\le C\gamma\cdot \frac1\gamma\|E\|\cdot \frac1\gamma\cdot
\frac1\gamma\{\|Ee_j\|_2+\|E\|\|Pe_j\|_2\}
\le \frac{C}{\gamma^2}\|E\|\{\|Ee_j\|_2+\|E\|\|Pe_j\|_2\},
\]
which proves \eqref{eq:proj-remainder-col-refined}.
\end{proof}

% =========================================================

\subsection{Row-wise perturbation: proofs}\label{app:proof-rowwise}

This appendix proves the deterministic perturbation statements that convert an averaged projector into the canonical Topic-SCORE basis. The first column is not chosen from an arbitrary top-$K$ eigenbasis: it is reconstructed by projecting the pooled empirical marginal. The only remaining alignment is the residual $(K-1)$-dimensional orthogonal matrix $O_\perp$.

\subsubsection{Proof of Lemma~\ref{lem:proj-col-from-avg}}
\begin{proof}
Write $E:=\bar P-P$ and let $\hat\Xi$ be any orthonormal basis of the range of $\widehat P$. Since $P$ is a rank-$K$ projector, its nonzero eigenvalues are all equal to one and the eigengap between the signal cluster and the null space is one. Applying Lemma~\ref{lem:rowwise-eig-general} to the pair $(P,P+E)$ gives an orthogonal matrix $O\in\R^{K\times K}$ such that
\[
\max_j h_j^{-1/2}\|e_j^\top(\hat\Xi O-\Xi)\|_2
\le
C\left[\|E\|+\max_j h_j^{-1/2}\|Ee_j\|_2\right],
\qquad
\|\hat\Xi O-\Xi\|\le C\|E\|.
\]
Because
\[
\widehat P-P=(\hat\Xi O-\Xi)(\hat\Xi O)^\top+\Xi(\hat\Xi O-\Xi)^\top,
\]
we have, for every $j$,
\[
\|(\widehat P-P)e_j\|_2
=\|e_j^\top(\widehat P-P)\|_2
\le
\|e_j^\top(\hat\Xi O-\Xi)\|_2
+\|e_j^\top\Xi\|_2\|\hat\Xi O-\Xi\|.
\]
The oracle row-scale bound $\|e_j^\top\Xi\|_2\le C\sqrt{h_j}$ then gives the displayed row bound. The operator-norm bound follows from Davis--Kahan for the projector signal. This proves the lemma.
\end{proof}

\subsubsection{Proof of Lemma~\ref{lem:canonical-first}}
\begin{proof}
Let
\[
z:=\widehat P u_d,
\qquad
\nu:=\Delta_1(D,D_0)+\max_jh_j^{-1/2}\|(\widehat P-P)e_j\|_2+\|\widehat P-P\|.
\]
Because $P\xi_1=\xi_1$ and $\|u_d\|_2=\|\xi_1\|_2=1$,
\[
z-\xi_1=(\widehat P-P)u_d+P(u_d-\xi_1).
\]
For the first term,
\[
|e_j^\top(\widehat P-P)u_d|
\le \|(\widehat P-P)e_j\|_2\|u_d\|_2
=\|(\widehat P-P)e_j\|_2.
\]
For the second term, write $P=\Xi\Xi^\top$. By Lemma~\ref{lem:oracle-row-geometry}, $\|e_j^\top\Xi\|_2\le C\sqrt{h_j}$. On the event $\Delta_1(D,D_0)\le c$, the diagonal entries of $M_d$ and $M_0$ are comparable and
\[
|u_d(j)-\xi_1(j)|=\left|\sqrt{M_d(j,j)}-\sqrt{M_0(j,j)}\right|
\le C\Delta_1(D,D_0)\sqrt{h_j}.
\]
Since $\sum_j h_j=K$, this gives $\|u_d-\xi_1\|_2\le C\Delta_1(D,D_0)$. Hence
\[
|e_j^\top P(u_d-\xi_1)|
\le \|e_j^\top\Xi\|_2\,\|\Xi^\top(u_d-\xi_1)\|_2
\le C\sqrt{h_j}\Delta_1(D,D_0).
\]
Therefore
\[
\max_j h_j^{-1/2}|z(j)-\xi_1(j)|
\le C\Delta_1(D,D_0)+\max_jh_j^{-1/2}\|(\widehat P-P)e_j\|_2.
\]
Also,
\[
\|z-\xi_1\|_2\le \|\widehat P-P\|+C\Delta_1(D,D_0).
\]
By the assumed smallness of
$\Delta_1(D,D_0)+\|\widehat P-P\|$ and the preceding bound,
$\|z\|_2\ge 1/2$. Since $\hat\xi_1=z/\|z\|_2$,
\[
\hat\xi_1-\xi_1
=\frac{z-\xi_1}{\|z\|_2}
+\left(\frac1{\|z\|_2}-1\right)\xi_1.
\]
Using $|\xi_1(j)|\le C\sqrt{h_j}$ and $|\|z\|_2-1|\le\|z-\xi_1\|_2$ yields
\[
\max_jh_j^{-1/2}|\hat\xi_1(j)-\xi_1(j)|
\le C\left[\Delta_1(D,D_0)+\max_jh_j^{-1/2}\|(\widehat P-P)e_j\|_2+\|\widehat P-P\|\right].
\]
\end{proof}

\subsubsection{Proof of Lemma~\ref{lem:residual-basis}}
\begin{proof}
Set
\[
\eta:=\Delta_1(D,D_0)+\max_jh_j^{-1/2}\|(\widehat P-P)e_j\|_2+\|\widehat P-P\|.
\]
Lemma~\ref{lem:canonical-first} gives
\[
\max_j h_j^{-1/2}|\hat\xi_1(j)-\xi_1(j)|\le C\eta,
\qquad
\|\hat\xi_1-\xi_1\|_2\le C\eta.
\]
Now
\[
\widehat P_\perp-P_\perp
=(\widehat P-P)-(\hat\xi_1\hat\xi_1^\top-\xi_1\xi_1^\top),
\]
and
\[
\hat\xi_1\hat\xi_1^\top-\xi_1\xi_1^\top
=(\hat\xi_1-\xi_1)\hat\xi_1^\top+\xi_1(\hat\xi_1-\xi_1)^\top.
\]
Consequently,
\[
\|\widehat P_\perp-P_\perp\|\le C\eta.
\]
For the row/column bound, using $\hat\xi_1(j)\asymp\sqrt{h_j}$ in the perturbative regime,
\[
\|(\widehat P_\perp-P_\perp)e_j\|_2
\le \|(\widehat P-P)e_j\|_2
+ |\hat\xi_1(j)|\,\|\hat\xi_1-\xi_1\|_2
+ |\hat\xi_1(j)-\xi_1(j)|,
\]
so
\[
\max_jh_j^{-1/2}\|(\widehat P_\perp-P_\perp)e_j\|_2\le C\eta.
\]
Apply the same projector-to-basis argument as in Lemma~\ref{lem:proj-col-from-avg} to the rank-$(K-1)$ projectors $P_\perp$ and $\widehat P_\perp$. Since $\|e_j^\top\Xi_\perp\|_2\le C\sqrt{h_j}$, there exists $O_\perp\in\R^{(K-1)\times(K-1)}$ orthogonal such that
\[
\max_jh_j^{-1/2}\|e_j^\top(\hat\Xi_\perp O_\perp-\Xi_\perp)\|_2\le C\eta.
\]
\end{proof}

\subsubsection{Proof of Lemma~\ref{lem:vector-bernstein}}
\begin{proof}
Fix $j$ and write
\[
r_j-\E r_j=\sum_{m=1}^M v_{m,j},
\qquad
v_{m,j}:=\alpha_m\Bigl((\mathcal P_m-P)e_j-\E[(\mathcal P_m-P)e_j]\Bigr).
\]
Then $\E v_{m,j}=0$, $\|v_{m,j}\|_2\le 4\alpha_m$, and
\[
\sum_{m=1}^M\E\|v_{m,j}\|_2^2\le \sum_{m=1}^M\alpha_m^2\tau_{m,j}^2.
\]
Put $L=4\max_m\alpha_m$ and
$V=\sum_m\E\|v_{m,j}\|_2^2$. For every integer $q\ge3$,
\[
\sum_m\E\|v_{m,j}\|_2^q
\le L^{q-2}V
\le\frac{q!}{2}V(L/3)^{q-2},
\]
because $q!/2\ge3^{q-2}$. The threshold $\sqrt{2Vu}+Bu$ established
in the proof of Lemma~\ref{lem:hilbert-bernstein-criterion}, with $B=L/3$,
therefore gives, with $u=\log(2/\delta)$,
\[
\|r_j-\E r_j\|_2
\le
\sqrt{2u}\left(\sum_{m=1}^M\alpha_m^2\tau_{m,j}^2\right)^{1/2}
+\frac43u\max_m\alpha_m
\]
with probability at least $1-\delta$.
\end{proof}

\subsection{Multinomial toolkit}\label{app:multinomial}

\subsubsection{Identity-centered Gram expectation}\label{app:G-expectation}

For a generic client corpus \(D=[d_1,\dots,d_n]\) with \(d_i=X_i/N\) and \(X_i\sim\mathrm{Multinomial}(N,\pi_i)\), let \(D_0=[\pi_1,\dots,\pi_n]\), let \(M_* \) be any deterministic positive diagonal matrix, and define the local population marginal
\[
M_{0,\mathrm{loc}}:=\diag\Bigl(\frac1nD_0\1\Bigr).
\]
Set
\[
\tilde G^{\mathrm{svd}}(M_*):=\frac1nM_*^{-1/2}DD^\top M_*^{-1/2}-\frac1N I_p,
\]
and
\[
G_0(M_*):=\Bigl(1-\frac1N\Bigr)\frac1nM_*^{-1/2}D_0D_0^\top M_*^{-1/2}.
\]

\begin{lemma}[Identity-centered fixed-whitening expectation]\label{lem:G_expectation}
Under common document length \(N\),
\[
\E[\tilde G^{\mathrm{svd}}(M_*)\mid W]
=
G_0(M_*)+\frac1N\{M_*^{-1/2}M_{0,\mathrm{loc}}M_*^{-1/2}-I_p\}.
\]
In particular, with \(M_*=M_0\) and \(D=D^{(m)}\), this gives
\[
\E[\tilde G_m\mid W^{(1:M)}]=G_{0,m}+D_{0,m}^{\mathrm{svd}}.
\]
\end{lemma}

\begin{proof}
For each \(i\),
\[
\E[d_i]=\pi_i,
\qquad
\E[d_id_i^\top]=\Bigl(1-\frac1N\Bigr)\pi_i\pi_i^\top+\frac1N\diag(\pi_i).
\]
Therefore
\[
\E\left[\frac1nDD^\top\Bigm|W\right]
=
\Bigl(1-\frac1N\Bigr)\frac1nD_0D_0^\top+\frac1N M_{0,\mathrm{loc}}.
\]
Pre- and post-multiplying by \(M_*^{-1/2}\) and subtracting \(N^{-1}I_p\) proves the claim.
\end{proof}

\begin{remark}[Proof-level note on unequal document lengths]\label{rem:unequal-length-proof-note}
The main text uses the common-length convention to keep notation readable. If document lengths are unequal, the identity-centered ordinary-SVD matrix remains the algorithmic object, while the fixed-whitening expectation gains the corresponding length-weighted diagonal mean.  The federated projector and canonical-direction arguments are unchanged; only the deterministic diagonal SVD-bias term and the effective document-length scale in the concentration bounds are altered.
\end{remark}

\begin{remark}[Why the identity shift creates a diagonal bias]\label{rem:identity-shift-bias}
The identity shift \(N^{-1}I_p\) is exactly the shift implemented by ordinary SVD, and it does not alter the empirical singular vectors.  It is not an exact local multinomial debiasing term under pooled whitening: unless the local population marginal equals the pooled marginal, the fixed-whitening mean contains the diagonal perturbation \(D_{0,m}^{\mathrm{svd}}\).  This term is kept explicitly in the theorem radius through \(\zeta_N^{\mathrm{svd}}\).
\end{remark}

\subsubsection{Moment bounds and bridge control}

The following bounds summarize the multinomial concentration ingredients used throughout the main proofs.

\begin{lemma}[Whitened Gram fluctuation bounds]\label{lem:gram-moments}
Under Assumptions~\ref{assump:hmin}--\ref{assump:Sigma}, for the centered fixed-whitening sampling fluctuation
\[
E:=\tilde G_m-\E[\tilde G_m\mid W^{(1:M)}],
\]
the column moment bound
\[
\E\norm{Ee_j}_2^2\le C h_j\beta_n^2\frac{p}{Nn}
\]
holds for every $j\in[p]$. If, in addition, the local perturbative scaling
\[
\beta_n^2\frac{p^2}{Nn}\le c_0
\]
holds for a sufficiently small absolute constant $c_0>0$, then
\[
\E\norm{E}^2\le C\beta_n^2\frac{p}{Nn},
\qquad
\E\norm{E}^4\le C\beta_n^4\Bigl(\frac{p}{Nn}\Bigr)^2.
\]
In fact, the direct moment argument in Lemma~\ref{lem:Eop} proves these two operator moment bounds without the additional local scaling. Moreover, with probability at least $1-\delta$,
\[
\Delta_1(D,D_0)
\le
C\sqrt{\frac{p\log(2p/\delta)}{NnM}}
+
C\frac{p\log(2p/\delta)}{NnM}.
\]
In the perturbative regime $p\log(2p/\delta)/(NnM)\le 1$, the second term is absorbed by the first.
\end{lemma}

\begin{remark}
Lemma~\ref{lem:gram-moments} is the federated counterpart of the moment bounds used in the single-site Topic-SCORE analysis of Ke and Wang's SVD Topic-SCORE paper. The first two inequalities control the centered local sampling fluctuation under fixed whitening, while the deterministic identity-shift diagonal bias is tracked separately through \(D_{0,m}^{\mathrm{svd}}\). The final display quantifies the extra error induced by the shared empirical whitening matrix.
\end{remark}

\subsubsection{Basic multinomial moments}

\begin{lemma}[Second moments of multinomial proportions]\label{lem:multi-second}
Condition on $\pi\in\R^p$ with $\pi\ge 0$ and $\1^\top\pi=1$, and let $d=X/N$ with $X\sim\mathrm{Multinomial}(N,\pi)$. Then for all $j\neq \ell$,
\[
\E[d(j)]=\pi(j),
\qquad
\E[d(j)^2]=\pi(j)^2+\frac{1}{N}\pi(j)(1-\pi(j))\le \pi(j)^2+\frac{1}{N}\pi(j),
\]
\[
\E[d(j)d(\ell)]=\pi(j)\pi(\ell)-\frac{1}{N}\pi(j)\pi(\ell).
\]
\end{lemma}

\begin{proof}
The identities are standard multinomial moment formulas. Indeed,
\[
\E[d(j)]=\frac{1}{N}\E[X_j]=\pi(j),
\qquad
\Var(d(j))=\frac{1}{N^2}\Var(X_j)=\frac{1}{N}\pi(j)(1-\pi(j)),
\]
so
\[
\E[d(j)^2]=\Var(d(j))+\E[d(j)]^2=\pi(j)^2+\frac{1}{N}\pi(j)(1-\pi(j)).
\]
Likewise, for $j\neq \ell$,
\[
\E[d(j)d(\ell)]
=\frac{1}{N^2}\E[X_jX_\ell]
=\frac{1}{N^2}\bigl(\Cov(X_j,X_\ell)+\E[X_j]\E[X_\ell]\bigr)
=\pi(j)\pi(\ell)-\frac{1}{N}\pi(j)\pi(\ell),
\]
because $\Cov(X_j,X_\ell)=-N\pi(j)\pi(\ell)$ for multinomial counts.
\end{proof}

\begin{lemma}[Fourth moments and mixed fourth moments]\label{lem:multi-fourth}
Condition on $\pi$ and let $d=X/N$ with $X\sim\mathrm{Multinomial}(N,\pi)$. Then for $j\neq \ell$,
\[
\E[d(j)^2d(\ell)^2]
\le
C\Bigl(
\pi(j)^2\pi(\ell)^2
+\frac{1}{N}\bigl(\pi(j)^2\pi(\ell)+\pi(j)\pi(\ell)^2\bigr)
+\frac{1}{N^2}\pi(j)\pi(\ell)
\Bigr),
\]
and for each $j$,
\[
\E[d(j)^4]
\le
C\Bigl(\pi(j)^4+\frac{1}{N}\pi(j)^3+\frac{1}{N^2}\pi(j)^2+\frac{1}{N^3}\pi(j)\Bigr),
\]
where $C>0$ is an absolute constant.
\end{lemma}

\begin{proof}
For $j\neq \ell$, write
\[
X_j^2X_\ell^2
=
\bigl(X_j(X_j-1)+X_j\bigr)\bigl(X_\ell(X_\ell-1)+X_\ell\bigr)
\]
and expand into four terms. Using the factorial-moment identities for multinomial counts,
\[
\E[X_j(X_j-1)X_\ell(X_\ell-1)]
=N(N-1)(N-2)(N-3)\pi(j)^2\pi(\ell)^2,
\]
\[
\E[X_j(X_j-1)X_\ell]
=N(N-1)(N-2)\pi(j)^2\pi(\ell),
\qquad
\E[X_jX_\ell(X_\ell-1)]
=N(N-1)(N-2)\pi(j)\pi(\ell)^2,
\]
and
\[
\E[X_jX_\ell]=N(N-1)\pi(j)\pi(\ell).
\]
Dividing by $N^4$ and upper-bounding $(N-r)/N\le 1$ yields the displayed inequality for $\E[d(j)^2d(\ell)^2]$. The bound for $\E[d(j)^4]$ follows similarly by expanding $X_j^4$ into factorial moments and dividing by $N^4$.
\end{proof}

\subsubsection{Column and operator bounds for the Gram fluctuation}

For the identity-centered SVD Gram, define for a generic document
\[
Q_i:=M_0^{-1/2}d_id_i^\top M_0^{-1/2},
\qquad d_i=X_i/N,
\]
and the centered sampling fluctuation matrix
\[
E:=\frac1n\sum_{i=1}^n\{Q_i-\E[Q_i\mid W]\}.
\]
When the generic corpus is client \(m\), this matrix equals
\[
E=\tilde G_m-\E[\tilde G_m\mid W^{(1:M)}]
=\tilde G_m-G_{0,m}-D_{0,m}^{\mathrm{svd}}.
\]
The deterministic diagonal term \(D_{0,m}^{\mathrm{svd}}\) is not included in the centered fluctuation bounds below; it is controlled separately by Lemma~\ref{lem:Dsvd_concentration}.

\begin{lemma}[Column variance bound for the centered whitened Gram fluctuation]\label{lem:Eej}
Under Assumptions~\ref{assump:hmin}--\ref{assump:Sigma}, for every $j\in[p]$,
\[
\E\norm{Ee_j}_2^2\le C h_j\beta_n^2\frac{p}{Nn}.
\]
\end{lemma}

\begin{proof}
All expectations below are conditional on the fixed design. Fix a document
with mean $\pi=Aw_i$ and define
\[
 \mu=M_0^{-1/2}\pi,\quad
 V_t=M_0^{-1/2}(e_{X_t}-\pi),\quad
 g=N^{-1}\sum_{t=1}^N V_t,\quad \zeta=g_j,
\]
where the $X_t$ are independent with distribution $\pi$.
Regularity gives
\[
 \|\mu\|_2\le C,\quad |\mu_j|^2\le Ch_j,\quad
 \|V_t\|_2\le C\sqrt p,\quad
 \mathbb E\|V_t\|_2^2\le Cp,\quad
 \mathbb E V_t(j)^2\le C.
\]
Set $H_t=V_t(j)$ and
\[
 a=\mathbb E\|V_t\|_2^2,\quad b=\mathbb E H_t^2,\quad
 c=\mathbb E(V_tH_t),\quad d=\mathbb E(\|V_t\|_2^2H_t^2).
\]
Then $a\le Cp$, $b\le C$, $\|c\|_2^2\le ab\le Cp$, and
$d\le Cp\,\mathbb E H_t^2\le Cp$. Expanding the four-index sum,
all singleton-index terms vanish by independence and centering. Thus
\begin{equation}\label{eq:repair-product-second}
 \mathbb E(\|g\|_2^2\zeta^2)
 =\frac{Nd+N(N-1)(ab+2\|c\|_2^2)}{N^4}
 \le \frac{Cp}{N^2}.
\end{equation}
The term $d/N^3$ accounts for coincident indices.
Writing $Z_i=Q_i-\mathbb E Q_i$, we have
\[
 Z_i e_j=\mu\zeta+\mu_j g+
                    \{g\zeta-\mathbb E(g\zeta)\}.
\]
Since $\mathbb E\zeta^2\le C/N$ and
$\mathbb E\|g\|_2^2\le Cp/N$, it follows that
\[
 \mathbb E\|Z_i e_j\|_2^2
 \le C\left\{\frac1N+\frac{h_jp}{N}+\frac p{N^2}\right\}
 \le Ch_j\beta_n^2\frac pN.
\]
For the last step, $h_jp\ge c$, and the definition of $\beta_n$ implies
$\beta_n\ge\max\{1,p/N\}$, hence $p/N\le C\beta_n^2$.
Indeed, if $N\ge p$ the claim is immediate; if $N<p$, then
$p^2/N^{3/2}\ge p/N$, so either branch in its definition is at least $p/N$.
Finally the document summands are independent and centered, and therefore
\[
 \mathbb E\|Ee_j\|_2^2
 =\frac1{n^2}\sum_{i=1}^n\mathbb E\|Z_i e_j\|_2^2
 \le Ch_j\beta_n^2\frac p{Nn}.
\]
\end{proof}

\begin{lemma}[Operator-norm concentration for the centered Gram fluctuation]
\label{lem:Eop}
Under Assumptions~\ref{assump:hmin}--\ref{assump:Sigma}, for every
$u\ge1$,
\[
 \mathbb P\left\{\|E\|>
 C\left[\beta_n\sqrt{\frac{pu}{Nn}}
          +\beta_n\sqrt{\frac pN}\frac un\right]\right\}
 \le2e^{-u}.
\]
Consequently, for every $\delta\in(0,1)$,
\[
 \|E\|\le C\left[
 \beta_n\sqrt{\frac{p\log(2p/\delta)}{Nn}}
 +\beta_n\sqrt{\frac pN}\frac{\log(2p/\delta)}n
 \right]
\]
with probability at least $1-\delta$. Moreover,
\[
 \mathbb E\|E\|^2\le C\beta_n^2\frac p{Nn},\qquad
 \mathbb E\|E\|^4\le C\beta_n^4\left(\frac p{Nn}\right)^2.
\]
\end{lemma}
\begin{proof}
Work in the Hilbert space of matrices with the Frobenius norm. For each
document put $\mu_i=M_0^{-1/2}\pi_i$,
$g_i=M_0^{-1/2}(d_i-\pi_i)$, and $Z_i=Q_i-\mathbb E Q_i$. Then
\[
 Z_i=\mu_i g_i^\top+g_i\mu_i^\top+
           \{g_i g_i^\top-\mathbb E(g_i g_i^\top)\}.
\]
Using $\|\mu_i\|_2\le C$ and
Lemma~\ref{lem:whitened-vector-moments}, for every integer $q\ge2$,
\[
 \bigl(\mathbb E\|Z_i\|_F^q\bigr)^{1/q}
 \le C\left\{\sqrt{\frac{qp}{N}}+\frac{qp}{N}\right\}
 \le Cq\beta_n\sqrt{\frac pN}.
\]
Here the definition of $\beta_n$ gives $\beta_n\ge\max\{1,p/N\}$:
the claim is immediate for $N\ge p$, while for $N<p$ one has
$p^2/N^{3/2}\ge p/N$, so both branches give the asserted bound.
Since $q^q\le e^q q!$, the preceding estimate yields the Bernstein
moment condition with per-document variance proxy $C\beta_n^2p/N$
and scale proxy $C\beta_n\sqrt{p/N}$. Apply
Lemma~\ref{lem:hilbert-bernstein-criterion} to the independent centered
matrices $Z_i$, divide by $n$, and use $\|E\|\le\|E\|_F$ to obtain
the tail statement. The logarithmic specialization follows by increasing
constants if necessary. Integration of the tail gives
\[
 \mathbb E\|E\|^2
 \le C\beta_n^2\frac pN(n^{-1}+n^{-2}),\qquad
 \mathbb E\|E\|^4
 \le C\beta_n^4\left(\frac pN\right)^2(n^{-2}+n^{-4}).
\]
Using $n\ge1$ proves the moment statements. No local perturbative
scaling is needed for these moment bounds.
\end{proof}

\begin{lemma}[Negligible operator-tail terms for fixed-whitening truncation]\label{lem:operator-tail-truncation}
Assume Assumptions~\ref{assump:hmin}--\ref{assump:Sigma}, the envelope bound in Lemma~\ref{lem:beta-envelope}, and the perturbative scaling $\beta_n^2p^2/(Nn)\le c_0$ with $c_0$ sufficiently small. Fix a constant $\gamma>0$ bounded away from zero, and let
\[
F:=\tilde G_m-\E[\tilde G_m\mid W^{(1:M)}],
\qquad
B:=D_{0,m}^{\mathrm{svd}},
\qquad
E^{(s)}:=F+B,
\]
with $\|B\|\le C/N$, $\max_jh_j^{-1/2}\|Be_j\|_2\le C\zeta_N^{\mathrm{svd}}$, and $\zeta_N^{\mathrm{svd}}\le c_0$. Set
\[
 a_n^{\mathrm{fw}}:=\beta_n^2\frac p{Nn}+\{\zeta_N^{\mathrm{svd}}\}^2,
 \qquad
 \mathcal T:=\{\|E^{(s)}\|\le \gamma/4\}.
\]
Then, after increasing constants if necessary, uniformly in $m$ and $j$,
\[
\Pbb(\mathcal T^c)\le C h_j a_n^{\mathrm{fw}},
\qquad
\Pbb(\mathcal T^c)\le C a_n^{\mathrm{fw}},
\]
and
\[
\E[\|F\|\mathbf 1_{\mathcal T^c}]\le C a_n^{\mathrm{fw}},
\qquad
\E[\|Fe_j\|_2\mathbf 1_{\mathcal T^c}]\le C\sqrt{h_j}\,a_n^{\mathrm{fw}}.
\]
\end{lemma}
\begin{proof}
Put
\[
 a_F:=\beta_n^2\frac p{Nn}.
\]
Choose the universal constant $c_0$ in the perturbative assumptions sufficiently small that
$\|B\|\le \gamma/16$. Then
\[
 \mathcal T^c\subseteq \{\|F\|>\gamma/8\}.
\]
The fourth-moment part of Lemma~\ref{lem:Eop} gives
\[
 \E\|F\|^4\le C a_F^2.
\]
Markov's inequality therefore gives
\[
 \Pbb(\mathcal T^c)\le C a_F^2.
\]
Since the perturbative scaling implies $a_F\le c_0/p$ and Assumption~\ref{assump:hmin} gives
$h_j\ge c/p$ for fixed $K$, we have
\[
 a_F^2\le C h_j a_F\le C h_j a_n^{\mathrm{fw}},
 \qquad
 a_F^2\le C a_n^{\mathrm{fw}}.
\]
This proves the two displayed probability bounds. Notice that no large-deviation tail at parameter
$u\asymp p^2$ is being used here; the fourth-moment estimate is sufficient and avoids any restriction
coming from bounded-increment terms in the operator tail.

For the truncated expectations, Cauchy--Schwarz and the second-moment bounds in
Lemmas~\ref{lem:Eop} and~\ref{lem:Eej} give
\[
\E[\|F\|\mathbf 1_{\mathcal T^c}]
\le (\E\|F\|^2)^{1/2}\Pbb(\mathcal T^c)^{1/2}
\le C a_F^{3/2}
\le C a_n^{\mathrm{fw}},
\]
because $a_F\le1$ after decreasing $c_0$ if needed. Similarly,
\[
\E[\|Fe_j\|_2\mathbf 1_{\mathcal T^c}]
\le (\E\|Fe_j\|_2^2)^{1/2}\Pbb(\mathcal T^c)^{1/2}
\le C\sqrt{h_j}\,a_F^{3/2}
\le C\sqrt{h_j}\,a_n^{\mathrm{fw}}.
\]
The deterministic diagonal term $B$ is included in $a_n^{\mathrm{fw}}$ through
$\{\zeta_N^{\mathrm{svd}}\}^2$, and its only role in the tail event is the preliminary inclusion
$\mathcal T^c\subseteq\{\|F\|>\gamma/8\}$.
\end{proof}

\begin{lemma}[Vector moments for whitened multinomial averages]
\label{lem:whitened-vector-moments}
Under Assumptions~\ref{assump:hmin}--\ref{assump:Sigma}, let
$d=N^{-1}\sum_{t=1}^Ne_{X_t}$ with independent $X_t\sim\pi$,
where $\pi=Aw$ for a fixed simplex weight $w$, and set
\[
 g=M_0^{-1/2}(d-\pi),\qquad
 \zeta_j=\frac{d(j)-\pi(j)}{\sqrt{m_0(j)}}.
\]
Then
\[
 \mathbb E\|g\|_2^2\le C\frac pN,\qquad
 \mathbb E|\zeta_j|^2\le\frac CN.
\]
For every real $q\ge2$,
\begin{equation}\label{eq:repair-g-moments}
 \bigl(\mathbb E\|g\|_2^q\bigr)^{1/q}
 \le C\sqrt{\frac{qp}{N}},\qquad
 \bigl(\mathbb E|\zeta_j|^q\bigr)^{1/q}
 \le C\left\{\sqrt{\frac qN}+\frac{q}{N\sqrt{h_j}}\right\}.
\end{equation}
\end{lemma}
\begin{proof}
Set $V_t=M_0^{-1/2}(e_{X_t}-\pi)$. The fixed-design envelope and
$m_0(j)\asymp h_j\gtrsim p^{-1}$ give
\[
 \mathbb EV_t=0,\quad \|V_t\|_2\le C\sqrt p,\quad
 \mathbb E\|V_t\|_2^2\le Cp,\quad
 \mathbb EV_t(j)^2\le C,\quad |V_t(j)|\le Ch_j^{-1/2}.
\]
Independence proves the second-moment assertions.
For $f(X_1,\ldots,X_N)=\|N^{-1}\sum_tV_t\|_2$, changing one coordinate
changes $f$ by at most $C\sqrt p/N$, while
$\mathbb Ef\le C\sqrt{p/N}$. The bounded-differences inequality gives
\[
 \mathbb P\{f>\mathbb Ef+t\}\le\exp(-cNt^2/p),\qquad t\ge0.
\]
For clarity, this inequality follows by exposing the word draws in order:
each increment of the Doob martingale for $f$ has conditional range at most
$C\sqrt p/N$; the conditional bounded-variable exponential-moment bound
and its product over $N$ draws give
$\mathbb E\exp\{\lambda(f-\mathbb Ef)\}\le\exp(C\lambda^2p/N)$.
Chernoff's inequality gives the displayed tail, and integration gives
the first inequality in~\eqref{eq:repair-g-moments}.

For $\zeta_j=N^{-1}\sum_tV_t(j)$, scalar Bernstein has total variance
at most $C/N$ and summand range at most $C/(N\sqrt{h_j})$. Its tail
integration yields
\[
 \|\zeta_j\|_{L^q}
 \le C\left\{\sqrt{q/N}+q/(N\sqrt{h_j})\right\},
\]
as required. This uses a trace-scale sub-Gaussian bound only for the
Euclidean norm of $g$; it does not assert a variance-only sub-Gaussian
bound for every scalar whitened direction.
\end{proof}

\begin{lemma}[A Hilbert-space Bernstein criterion]\label{lem:hilbert-bernstein-criterion}
Conditionally on $W$, let $Z_1,\ldots,Z_n$ be independent random
vectors in a finite-dimensional Hilbert space with $\E[Z_i\mid W]=0$.
Suppose there are numbers $V\ge0$ and $B>0$, deterministic given $W$, such that
\[
\sum_{i=1}^n\E[\|Z_i\|_2^2\mid W]\le V,
\qquad
\sum_{i=1}^n\E[\|Z_i\|_2^q\mid W]\le \frac{q!}{2}VB^{q-2},\qquad q\ge3.
\]
Then, for every \(u\ge0\),
\[
\Pbb\left(\left\|\sum_{i=1}^n Z_i\right\|_2>C\{\sqrt{Vu}+Bu\}\Bigm|W\right)\le 2e^{-u}.
\]
\end{lemma}

\begin{proof}
All expectations and probabilities in this proof are conditional on $W$.
For $0<\lambda<B^{-1}$, the moment assumptions and the exponential series give
\[
\begin{aligned}
\sum_i\E\bigl[e^{\lambda\|Z_i\|_2}-1-\lambda\|Z_i\|_2\mid W\bigr]
&\le \frac V2\sum_{q\ge2}\lambda^qB^{q-2}\\
&=\frac{V\lambda^2}{2(1-B\lambda)}.
\end{aligned}
\]
The Hilbert-space case ($D=1$) of
Pinelis (1994, Theorem~3.1)\footnote{I. Pinelis,
``Optimum Bounds for the Distributions of Martingales in Banach Spaces,''
\emph{The Annals of Probability}, 22(4):1679--1706, 1994.
\url{https://doi.org/10.1214/aop/1176988477}.} therefore gives
\[
\begin{aligned}
&\Pbb\!\left(\left\|\sum_iZ_i\right\|_2>t\Bigm|W\right)\\
&\qquad\le2\exp\!\left\{-\lambda t+
\frac{V\lambda^2}{2(1-B\lambda)}\right\}.
\end{aligned}
\]
For $V>0$ and $u>0$, take
$s=\sqrt{2u/V}$, $\lambda=s/(1+Bs)$, and
$t=\sqrt{2Vu}+Bu$. The exponent is $-u$, proving the stronger threshold
$\sqrt{2Vu}+Bu$ and hence the displayed claim.
If $V=0$, every $Z_i$ vanishes almost surely conditionally on $W$;
the case $u=0$ is trivial.
\end{proof}

\begin{lemma}[Single-column Euclidean-norm concentration]\label{lem:Ecol_single}
Under Assumptions~\ref{assump:hmin}--\ref{assump:Sigma}, fix a client $m$ and a word $j$.  Let
\[
Z_{i,j}^{(m)}:=Q_i^{(m)}e_j-\E[Q_i^{(m)}e_j\mid W^{(1:M)}],
\qquad i=1,\ldots,n,
\]
where $Q_i^{(m)}$ is the single-document whitened outer-product matrix defined at the beginning of Appendix~\ref{app:multinomial}. Thus $Q_i^{(m)}$ itself is not centered; the vector $Z_{i,j}^{(m)}$ is the centered column summand. Then, for every $u\ge1$,
\[
\Pbb\left(
 h_j^{-1/2}\left\|\frac1n\sum_{i=1}^n Z_{i,j}^{(m)}\right\|_2
> C\left[
\beta_n\sqrt{\frac{pu}{Nn}}
+\beta_n^2\frac{pu}{Nn}
+\sqrt{\beta_n p}\,\frac{pu}{Nn}
+\beta_n\sqrt{\frac pN}\frac{u}{n}
\right]
\Bigm|W^{(1:M)}\right)
\le 2e^{-u}.
\]
\end{lemma}

\begin{proof}
All expectations are conditional on the fixed design. For a generic
document with mean $\pi=Aw_i$, write
\[
\mu=M_0^{-1/2}\pi,\qquad
g=M_0^{-1/2}(d_i-\pi),\qquad \zeta=g_j.
\]
Abbreviate $h=h_j$ and $\beta=\beta_n$. Define
\[
 v=h\beta^2\frac pN,\qquad
 b=\sqrt h\left\{
      \beta\sqrt{\frac pN}+\beta^2\frac pN
                 +\sqrt{\beta p}\frac pN\right\}.
\]
We will prove for every integer $q\ge3$ that
\begin{equation}\label{eq:repair-column-bernstein-moment}
 \mathbb E\|(Q_i-\mathbb E Q_i)e_j\|_2^q
 \le \frac{q!}{2}(C v)(C b)^{q-2}.
\end{equation}

Write
\[
 (Q_i-\mathbb E Q_i)e_j
  =\mu\zeta+\mu_jg+\{g\zeta-\mathbb E(g\zeta)\}.
\]
The first two terms satisfy
\[
 \|\mu\zeta\|_{L^q(\ell_2)}
 \le C\left\{\sqrt{q/N}+q/(N\sqrt h)\right\}
 \le Cq\sqrt v,
 \qquad
 \|\mu_jg\|_{L^q(\ell_2)}\le C\sqrt{qhp/N}\le Cq\sqrt v.
\]
Indeed $hp\ge c$, $\beta\ge1$, and $\beta^2N\ge p$ imply
$N^{-1}\le Cv$ and $(N^2h)^{-1}\le Cv$.
Since $b\ge\sqrt v$ and $q^q\le e^q q!$, both linear terms have the
Bernstein moments in~\eqref{eq:repair-column-bernstein-moment}.

For the product $Y=g\zeta$, let
\[
 a=\frac{\sqrt p}{N},\qquad
 d=\frac{\sqrt p}{N^{3/2}\sqrt h},\qquad
 L=\frac{C\sqrt p}{\sqrt h}.
\]
H\"older's inequality and~\eqref{eq:repair-g-moments} yield
\[
 \|Y\|_{L^q(\ell_2)}
 \le \|g\|_{L^{2q}(\ell_2)}\|\zeta\|_{L^{2q}}
 \le C(qa+q^{3/2}d).
\]
Moreover $\|Y\|_2\le L$ almost surely, and
\eqref{eq:repair-product-second} gives $\mathbb E\|Y\|_2^2\le Cv$.
The following comparisons are all deterministic:
\[
 a^2\le Cv,\qquad d^2\le Cv,\qquad
 b\ge\sqrt v,\qquad
 \frac{b}{d}\ge c\sqrt{\beta N},\qquad
 \frac{L}{b}\le \frac{CN}{\sqrt\beta}.
\]
For example,
\[
 \frac{a^2}{v}=\frac1{h\beta^2N}\le C,
 \qquad
 \frac{d^2}{v}=\frac1{h^2\beta^2N^2}\le C,
\]
using $h\ge c/p$, $\beta^2N\ge p$, and $\beta N\ge p$.
The last two comparisons follow from the component
$\sqrt h\sqrt{\beta p}\,p/N$ of $b$ and $hp\ge c$.

If $3\le q\le\beta N$, then the $(qa)^q$ part is at most
$q!C^qv b^{q-2}$. For the other part, $q^q\le e^q q!$ gives
\[
 q^{3q/2}d^q
 \le q!C^q v b^{q-2}
       q\left(\frac{q}{\beta N}\right)^{(q-2)/2}
 \le q!C_1^q v b^{q-2},
\]
where the final factor $q$ is absorbed into $C_1^q$.
If $q>\beta N$, use instead the deterministic bound and the second moment:
\[
 \mathbb E\|Y\|_2^q
 \le L^{q-2}\mathbb E\|Y\|_2^2
 \le Cv(Cq b)^{q-2}
 \le q!C_2^qv b^{q-2}.
\]
Here $L/b\le CN/\sqrt\beta\le Cq$ since $q>\beta N$ and $\beta\ge1$.
Thus the required Bernstein moment bound holds for every $q\ge3$.
Centering $Y$ changes constants only, by Jensen's inequality and
$\|Y-\mathbb EY\|_{L^q}\le2\|Y\|_{L^q}$.

Combining the three terms establishes
\eqref{eq:repair-column-bernstein-moment}. The second-moment bound is
\[
 \mathbb E\|(Q_i-\mathbb E Q_i)e_j\|_2^2
 \le C\{N^{-1}+hp/N+p/N^2\}\le Cv,
\]
as follows from the same decomposition and
\eqref{eq:repair-product-second}.
The Hilbert-space Bernstein criterion for the independent document
summands therefore gives
\[
 \left\|\frac1n\sum_i(Q_i-\mathbb E Q_i)e_j\right\|_2
 \le C\left\{\sqrt{\frac{vu}{n}}+\frac{bu}{n}\right\}
\]
with probability at least $1-2e^{-u}$. Dividing by $\sqrt h$ gives
the claimed four-term bound.
\end{proof}

\begin{lemma}[Uniform clientwise Gram-column concentration]\label{lem:Ecol_hp}
Under Assumptions~\ref{assump:hmin}--\ref{assump:Sigma}, for every $\delta\in(0,1)$,
\[
\Pbb\Biggl(
\max_{1\le m\le M}\max_{1\le j\le p}
 h_j^{-1/2}\| (\tilde G_m-G_{0,m})e_j\|_2
\le C\{\vartheta_n(\delta)+\zeta_N^{\mathrm{svd}}\}
\Biggr)\ge1-\delta.
\]
\end{lemma}

\begin{proof}
For fixed $(m,j)$,
\[
(\tilde G_m-G_{0,m})e_j
=
\frac1n\sum_{i=1}^n Z_{i,j}^{(m)}
+
D_{0,m}^{\mathrm{svd}}e_j,
\]
where \(Z_{i,j}^{(m)}\) is the centered sampling column from Lemma~\ref{lem:Ecol_single}. Apply Lemma~\ref{lem:Ecol_single} with \(u=\ell_{n,M}(\delta)=\log(2pM/\delta)\) and take a union bound over all \((m,j)\in[M]\times[p]\). Lemma~\ref{lem:Dsvd_concentration} controls the deterministic diagonal term. The resulting right-hand side is \(C\{\vartheta_n(\delta)+\zeta_N^{\mathrm{svd}}\}\).
\end{proof}

\begin{lemma}[Concentration of the pooled marginal matrix]\label{lem:Md_concentration}
Under Assumptions~\ref{assump:hmin}--\ref{assump:Sigma} and the conditional fixed-design setup, suppose all documents have common length $N$ and $n_m=n$. Then for every $\delta\in(0,1)$, with probability at least $1-\delta$,
\[
\Delta_1(D,D_0)
\le
C\sqrt{\frac{p\log(2p/\delta)}{NnM}}
+C\frac{p\log(2p/\delta)}{NnM}.
\]
In the perturbative regime $p\log(2p/\delta)/(NnM)\le1$, the second term is absorbed by the first.
\end{lemma}

\begin{proof}
For each $j$, write $M_d(j,j)$ as the average of the $NnM$
independent word indicators in the corpus, conditionally on $W^{(1:M)}$.
The centered summands have range at most $1/(NnM)$, their total variance
is at most $M_0(j,j)/(NnM)$, and the mean is $M_0(j,j)$.
A scalar Bernstein inequality combined with a union bound over $j\in[p]$ yields
\[
|M_d(j,j)-M_0(j,j)|
\le C\sqrt{\frac{M_0(j,j)\log(2p/\delta)}{NnM}}
+C\frac{\log(2p/\delta)}{NnM}
\]
for all $j$ with probability at least $1-\delta$. Since $M_0(j,j)\asymp h_j$ and $h_j\gtrsim p^{-1}$, division by $h_j$ gives the claim.
\end{proof}

\begin{lemma}[Deterministic size of the ordinary-SVD diagonal bias]\label{lem:Dsvd_concentration}
Under Assumptions~\ref{assump:hmin}--\ref{assump:Sigma},
\[
\Delta_{\mathrm{svd}}\le C\zeta_N^{\mathrm{svd}}=C\frac{\sqrt p}{N}.
\]
\end{lemma}

\begin{proof}
The matrix \(D_{0,m}^{\mathrm{svd}}\) is diagonal, with \(j\)th diagonal entry
\[
\frac1N\left\{\frac{M_{0,m}(j,j)}{M_0(j,j)}-1\right\}.
\]
Since \(M_{0,m}(j,j)\le h_j\) and \(M_0(j,j)\asymp h_j\) under Assumptions~\ref{assump:hmin}--\ref{assump:Sigma}, the absolute value of the ratio in braces is uniformly bounded by a constant. Hence
\[
\|D_{0,m}^{\mathrm{svd}}\|\le \frac C N,
\qquad
h_j^{-1/2}\|D_{0,m}^{\mathrm{svd}}e_j\|_2
\le \frac{C}{N\sqrt{h_j}}
\le C\frac{\sqrt p}{N}.
\]
Taking the maximum over \(m\) and \(j\) proves the claim.
\end{proof}

\subsubsection{Proof of Lemma~\ref{lem:gram-moments}}
\begin{proof}
The first display is exactly Lemma~\ref{lem:Eej}. The operator-norm second- and fourth-moment bounds follow from Lemma~\ref{lem:Eop}. The final display is Lemma~\ref{lem:Md_concentration}. Collecting these ingredients proves the claim.
\end{proof}

\begin{remark}
Lemma~\ref{lem:gram-moments} follows from elementary multinomial moments and Bernstein concentration for independent documents. The pooled marginal concentration controls the shared empirical whitening matrix $M_d$ in the bridge terms.
\end{remark}

% =========================================================
\subsection{A row-wise eigenvector perturbation lemma}\label{app:rowwise-eig-general}

The next lemma is the row-wise perturbation device underlying the singular-vector analysis in Topic-SCORE. It is stated abstractly because it is useful beyond the multinomial model itself.

\begin{lemma}[A full-cluster row-wise perturbation bound]\label{lem:rowwise-eig-general}
Let $G_0$ and $G=G_0+Y$ be symmetric $p\times p$ matrices. Assume $G_0\succeq0$ has rank $K$, and let $U_0\in\R^{p\times K}$ be an orthonormal basis for its range. Let $U\in\R^{p\times K}$ be an orthonormal basis for the leading $K$-dimensional invariant subspace of $G$. Suppose that, for some fixed $c\in(0,1)$,
\[
\lambda_K(G_0)\ge c\|G_0\|,
\qquad
\|Y\|\le c\|G_0\|/8.
\]
Then there exists an orthogonal matrix $O\in\R^{K\times K}$ such that, for every $j\in[p]$,
\[
\|e_j^\top(UO-U_0)\|_2
\le
\frac{C}{\|G_0\|}
\left\{\|Y\|\,\|e_j^\top U_0\|_2+\|Ye_j\|_2\right\},
\]
where $C$ depends only on $c$ and fixed $K$.
\end{lemma}

\begin{proof}
This is the full-signal-cluster version of the row-wise perturbation lemma used in single-site Topic-SCORE; it is the only version needed in the present federated proof.  Let $P_0=U_0U_0^\top$ and $P=UU^\top$.  Since the nonzero spectrum of $G_0$ is separated from the null space by at least $c\|G_0\|$ and $\|Y\|\le c\|G_0\|/8$, the leading $K$-dimensional invariant subspace of $G$ is well defined and Davis--Kahan gives
\[
\|P-P_0\|\le C\frac{\|Y\|}{\|G_0\|}.
\]
Let $O$ be the orthogonal polar factor of $U^\top U_0$; this choice minimizes $\|UO-U_0\|_F$.  Standard projector-to-basis algebra gives
\begin{equation}\label{eq:proj-to-basis-row-special}
\|e_j^\top(UO-U_0)\|_2
\le C\left\{\|(P-P_0)e_j\|_2+\|P-P_0\|\,\|e_j^\top U_0\|_2\right\}.
\end{equation}
For completeness, one way to see \eqref{eq:proj-to-basis-row-special} is to write
$UO-U_0=(P-P_0)U_0+U\{O-U^\top U_0\}$ and use
$\|O-U^\top U_0\|\le C\|P-P_0\|^2$, which follows from the singular values of $U^\top U_0$ being within $C\|P-P_0\|^2$ of one.

It remains to control the projector column.  Apply the Riesz expansion for the spectral projector of the full nonzero cluster of $G_0$.  Because the cluster is separated from the zero eigenvalue by a gap of order $\|G_0\|$, the first-order term has the cross-cluster form
\[
\mathcal L(Y)=G_0^\dagger Y(I-P_0)+(I-P_0)YG_0^\dagger
\qquad\text{on the range/null decomposition},
\]
where $G_0^\dagger$ denotes the Moore--Penrose inverse on $\mathrm{range}(G_0)$ and $\|G_0^\dagger\|\le C\|G_0\|^{-1}$.  Hence
\[
\|\mathcal L(Y)e_j\|_2
\le \frac{C}{\|G_0\|}\{\|Ye_j\|_2+\|Y\|\|P_0e_j\|_2\}.
\]
The quadratic Riesz remainder satisfies
\[
\|\mathcal R(Y)e_j\|_2
\le C\frac{\|Y\|}{\|G_0\|}\,\frac{\|Ye_j\|_2+\|Y\|\|P_0e_j\|_2}{\|G_0\|},
\]
which is absorbed by the preceding display because $\|Y\|\le c\|G_0\|/8$. Therefore
\[
\|(P-P_0)e_j\|_2
\le
\frac{C}{\|G_0\|}\{\|Ye_j\|_2+\|Y\|\|P_0e_j\|_2\}.
\]
Since $\|P_0e_j\|_2=\|e_j^\top U_0\|_2$, substituting this projector-column bound into \eqref{eq:proj-to-basis-row-special} proves the lemma.
\end{proof}

\begin{remark}[No client-specific canonical SCORE basis]\label{rem:no-local-score-basis}
Lemma~\ref{lem:rowwise-eig-general} is a row-wise perturbation device for subspaces and projector signals; it should not be read as saying that every local fixed-whitening eigenbasis is aligned with the pooled canonical basis $[\xi_1,\Xi_\perp]$. In general,
\[
G_{0,m}=\left(1-\frac1N\right)B\Sigma_{W,m}B^\top,
\qquad B=M_0^{-1/2}A,
\]
has eigenvectors that depend on the client-specific matrix $\Sigma_{W,m}$. Thus the local population first eigenvector need not equal the pooled canonical direction $\xi_1=M_0^{1/2}\1$, and local SCORE ratios built directly from a local eigenbasis need not be comparable to the oracle ratio matrix $R$.

Lemma~\ref{lem:rowwise-eig-general} is used to obtain row-wise
control from the averaged-projector signal in
Lemma~\ref{lem:proj-col-from-avg}.
The empirical-whitening projector bridge is established directly
by the common-reference resolvent comparison in Step~3 of the
proof of Lemma~\ref{lem:tau-bound-strong}. The canonical SCORE denominator is recovered only after server-side projector averaging by projecting the pooled marginal direction. Consequently, we do not assert a local-ratio guarantee: clients compute local SVD eigenspaces only for communication, while topic recovery uses only the server-side canonical ratio cloud.
\end{remark}

% =========================================================
\subsection{Proofs of the main theorems}\label{app:main-proofs}\label{app:proof-fedA}

\subsubsection{Proof of Theorem~\ref{thm:projector-avg}}
\begin{proof}
Apply Lemma~\ref{lem:vb-decomp} to decompose $\bar{\mathcal P}-P$ into variance and bias terms. The variance term is controlled by Lemma~\ref{lem:bernstein-matrix}. This yields \eqref{eq:projector-avg}. The subspace conclusion follows from Lemma~\ref{lem:dk-projector}.
\end{proof}

\subsubsection{Proof of Lemma~\ref{lem:tau-bound-strong}}
\begin{proof}
Fix a client \(m\) and suppress the client index. Write
\[
F^{(s)}:=\tilde G_m-\E[\tilde G_m\mid W^{(1:M)}],
\qquad
B^{\mathrm{svd}}:=D_{0,m}^{\mathrm{svd}},
\qquad
E^{(s)}:=\tilde G_m-G_{0,m}=F^{(s)}+B^{\mathrm{svd}},
\]
and
\[
E^{(w)}:=\hat G_m-\tilde G_m .
\]
Let \(\gamma_m:=\lambda_K(G_{0,m})\). Lemma~\ref{lem:M0-h-comparable}, Assumption~\ref{assump:local-richness}, and the standing upper-bound assumption \(N\ge2\) imply \(\gamma_m\ge c_{\mathrm{gap}}>0\) uniformly in \(m\); the factor \(1-1/N\) is bounded below by \(1/2\) and is absorbed into \(c_{\mathrm{gap}}\).

By Lemma~\ref{lem:gram-moments} and Lemma~\ref{lem:Dsvd_concentration},
\[
\E\|F^{(s)}e_j\|_2^2\le C h_j\beta_n^2\frac p{Nn},
\qquad
\E\|F^{(s)}\|^2\le C\beta_n^2\frac p{Nn},
\qquad
\E\|F^{(s)}\|^4\le C\beta_n^4\left(\frac p{Nn}\right)^2,
\]
and
\[
\|B^{\mathrm{svd}}\|\le \frac{C}{N},
\qquad
\max_j h_j^{-1/2}\|B^{\mathrm{svd}}e_j\|_2\le C\zeta_N^{\mathrm{svd}}.
\]

\paragraph{Step 1: fixed-whitening second moments.}
Set
\[
 a_n^{\mathrm{fw}}:=\beta_n^2\frac p{Nn}+\{\zeta_N^{\mathrm{svd}}\}^2.
\]
Let \(\mathcal E_m:=\{\|E^{(s)}\|\le\gamma_m/4\}\). Lemma~\ref{lem:operator-tail-truncation}, applied with the constant gap \(\gamma_m\ge c_{\mathrm{gap}}\), gives
\[
\Pbb(\mathcal E_m^c)\le C h_j a_n^{\mathrm{fw}},
\qquad
\Pbb(\mathcal E_m^c)\le C a_n^{\mathrm{fw}}.
\]
On \(\mathcal E_m\), Lemma~\ref{lem:proj-expansion} gives
\[
\| (\tilde P_m-P)e_j\|_2
\le
C\{\|E^{(s)}e_j\|_2+\|E^{(s)}\|\,\|Pe_j\|_2\}+C\|E^{(s)}\|^2.
\]
The oracle row-scale bound gives \(\|Pe_j\|_2\le C\sqrt{h_j}\). On \(\mathcal E_m^c\), use the crude bound \(\|(\tilde P_m-P)e_j\|_2\le2\). Therefore
\[
\E\|(\tilde P_m-P)e_j\|_2^2
\le
C\E\Bigl[\{\|E^{(s)}e_j\|_2^2+h_j\|E^{(s)}\|^2+\|E^{(s)}\|^4\}\mathbf 1_{\mathcal E_m}\Bigr]
+4\Pbb(\mathcal E_m^c).
\]
The first two terms are bounded by
\[
\E\|E^{(s)}e_j\|_2^2+h_j\E\|E^{(s)}\|^2
\le C h_j\left(\beta_n^2\frac p{Nn}+(\zeta_N^{\mathrm{svd}})^2\right).
\]
For the quadratic projector-remainder term, the refined bookkeeping is
\[
\E\|E^{(s)}\|^4
\le C\left(\beta_n^2\frac p{Nn}\right)^2+C N^{-4}
\le C h_j\left(\beta_n^2\frac p{Nn}+(\zeta_N^{\mathrm{svd}})^2\right).
\]
The last inequality uses \(\beta_n^2p^2/(Nn)\le c_0\), Assumption~\ref{assump:hmin}, and \(\zeta_N^{\mathrm{svd}}=\sqrt p/N\) together with the sharper operator bound \(\|B^{\mathrm{svd}}\|\le C/N\).  Combining these estimates with Lemma~\ref{lem:operator-tail-truncation} gives
\[
\E\|(\tilde P_m-P)e_j\|_2^2
\le C h_j\left(\beta_n^2\frac p{Nn}+(\zeta_N^{\mathrm{svd}})^2\right).
\]
The same argument without the column scale gives
\[
\E\|\tilde P_m-P\|^2
\le C\left(\beta_n^2\frac p{Nn}+(\zeta_N^{\mathrm{svd}})^2\right).
\]

\paragraph{Step 2: fixed-whitening projector bias.}
On \(\mathcal E_m\), the Riesz expansion gives
\[
\tilde P_m-P=L_m(F^{(s)})+L_m(B^{\mathrm{svd}})+R_m(E^{(s)}).
\]
The full, untruncated linear term satisfies \(\E L_m(F^{(s)})=0\). Hence
\[
\E\{L_m(F^{(s)})\mathbf 1_{\mathcal E_m}\}
=-\E\{L_m(F^{(s)})\mathbf 1_{\mathcal E_m^c}\}.
\]
Lemma~\ref{lem:operator-tail-truncation} and the column form of the linear projector derivative yield
\[
\bigl\|\E\{L_m(F^{(s)})\mathbf 1_{\mathcal E_m^c}\}\bigr\|
\le C a_n^{\mathrm{fw}},
\]
and
\[
\bigl\|\E\{L_m(F^{(s)})e_j\mathbf 1_{\mathcal E_m^c}\}\bigr\|_2
\le C\sqrt{h_j}\,a_n^{\mathrm{fw}}.
\]
The linear deterministic diagonal-bias term is bounded by
\[
\|L_m(B^{\mathrm{svd}})\|+\max_j h_j^{-1/2}\|L_m(B^{\mathrm{svd}})e_j\|_2
\le C\zeta_N^{\mathrm{svd}}.
\]
The second-order remainder is controlled on \(\mathcal E_m\) by the rank-\(K\) signal version of the refined columnwise Riesz remainder bound \eqref{eq:proj-remainder-col-refined} in Lemma~\ref{lem:proj-expansion}. This specialized version applies here because \(G_{0,m}\succeq0\) has rank \(K\), its non-target spectrum is exactly \(\{0\}\), and \(P\) projects onto \(\operatorname{range}(G_{0,m})\). Equivalently, it is the rank-\(K\) column-remainder part of Lemma~\ref{lem:second-order-bias} without using that lemma's mean-zero hypothesis. The mean-zero condition is used only for the full linear fluctuation \(L_m(F^{(s)})\). Specifically,
\[
\E\{\|R_m(E^{(s)})e_j\|_2\mathbf 1_{\mathcal E_m}\}
\le C\E\bigl[\|E^{(s)}\|\{\|E^{(s)}e_j\|_2+\|E^{(s)}\|\|Pe_j\|_2\}\bigr]
\le C\sqrt{h_j}\left(\beta_n^2\frac p{Nn}+(\zeta_N^{\mathrm{svd}})^2\right),
\]
where the final step uses Cauchy--Schwarz, the column and operator second-moment bounds above, and \(\|Pe_j\|_2\le C\sqrt{h_j}\).  The corresponding operator remainder satisfies
\[
\E\{\|R_m(E^{(s)})\|\mathbf 1_{\mathcal E_m}\}
\le C\E\|E^{(s)}\|^2
\le C\left(\beta_n^2\frac p{Nn}+(\zeta_N^{\mathrm{svd}})^2\right).
\]
The complement contributes at most the negligible tail terms from Lemma~\ref{lem:operator-tail-truncation}. Since \(\zeta_N^{\mathrm{svd}}\le c_0\), the term \((\zeta_N^{\mathrm{svd}})^2\) is absorbed by \(\zeta_N^{\mathrm{svd}}\). Thus
\[
\|\E\tilde P_m-P\|
\le C\left(\beta_n^2\frac p{Nn}+\zeta_N^{\mathrm{svd}}\right),
\]
and
\[
\|(\E\tilde P_m-P)e_j\|_2
\le C\sqrt{h_j}\left(\beta_n^2\frac p{Nn}+\zeta_N^{\mathrm{svd}}\right).
\]

\paragraph{Step 3: empirical-whitening bridge for the ordinary-SVD projector.}
Work on \(\mathcal E_{\mathrm{loc},m}\cap\mathcal E_{\mathrm{bridge},m}\). Lemma~\ref{lem:Gbridge_bound} gives
\[
\|E^{(w)}\|\le C\Delta_1(D,D_0)\{1+\|E^{(s)}\|\},
\]
and
\[
\|E^{(w)}e_j\|_2
\le C\sqrt{h_j}\Delta_1(D,D_0)
+C\Delta_1(D,D_0)\|E^{(s)}e_j\|_2
+C\frac{\Delta_1(D,D_0)}{N}.
\]
Choosing $c_{\mathrm{br}}$ sufficiently small ensures
$\|E^{(s)}\|+\|E^{(w)}\|\le c_{\mathrm{gap}}/4$.
We compare both perturbed matrices through the common rank-$K$
reference $G_{0,m}$. Choose a contour $\mathcal C$ enclosing the
nonzero spectrum of $G_{0,m}$, at distance at least
$c_{\mathrm{gap}}/2$ from its spectrum, and excluding zero.
The population spectral bounds imply that the contour length and
$|z|^{-1}$ on $\mathcal C$ are bounded by fixed constants. Define
\[
\widetilde R(z)=(zI-\tilde G_m)^{-1},
\qquad
\widehat R(z)=(zI-\hat G_m)^{-1}.
\]
Their operator norms are uniformly bounded on $\mathcal C$.
The Riesz formula and the resolvent identity give
\[
\hat P_m-\tilde P_m
=\frac1{2\pi i}\oint_{\mathcal C}
\widehat R(z)E^{(w)}\widetilde R(z)\,dz.
\]
For each column, the exact identity
\[
\widetilde R(z)e_j
=z^{-1}e_j+
z^{-1}\widetilde R(z)(G_{0,m}+E^{(s)})e_j
\]
implies
\[
\|E^{(w)}\widetilde R(z)e_j\|_2
\le C\left[
\|E^{(w)}e_j\|_2+
\|E^{(w)}\|
\{\|G_{0,m}e_j\|_2+\|E^{(s)}e_j\|_2\}
\right].
\]
Integrating and using $\|G_{0,m}e_j\|_2\le C\sqrt{h_j}$
therefore gives
\[
\|(\hat P_m-\tilde P_m)e_j\|_2
\le C\left[
\|E^{(w)}e_j\|_2+
\|E^{(w)}\|
\{\sqrt{h_j}+\|E^{(s)}e_j\|_2\}
\right].
\]
Substituting the preceding bridge bounds and using
$\|E^{(s)}\|\le c_{\mathrm{gap}}/8$ yields
\[
\|(\hat P_m-\tilde P_m)e_j\|_2
\le C\sqrt{h_j}\Delta_1(D,D_0)
+C\Delta_1(D,D_0)\|E^{(s)}e_j\|_2
+C\frac{\Delta_1(D,D_0)}{N}.
\]
This is stronger than the detailed column bound in the statement,
which retains an additional nonnegative quadratic remainder.
The same contour integral gives
$\|\hat P_m-\tilde P_m\|\le C\|E^{(w)}\|
\le C\Delta_1(D,D_0)$.
If $\mathcal E_{\mathrm{col},m}(\delta)$ also holds, divide by
$\sqrt{h_j}$, take the maximum over $j$, and use
$h_{\min}^{-1/2}\lesssim\sqrt p$ to obtain
\[
\max_jh_j^{-1/2}\|(\hat P_m-\tilde P_m)e_j\|_2
\le C\Delta_1(D,D_0)
\{1+\vartheta_n(\delta)+\zeta_N^{\mathrm{svd}}\}.
\]
Here the identity-shift contribution
$\sqrt p\,\Delta_1(D,D_0)/N$
equals $\zeta_N^{\mathrm{svd}}\Delta_1(D,D_0)$.
The simplified bridge bound in the statement follows.
\end{proof}

\subsubsection{Proof of Corollary~\ref{cor:single-client-op}}
\begin{proof}
Put $u=\log(2p/\delta)$ and
\[
a_\delta=
\beta_n\sqrt{\frac{pu}{Nn}}
+\beta_n^2\frac{pu}{Nn}
+\sqrt{\beta_n p}\,\frac{pu}{Nn}
+\beta_n\sqrt{\frac pN}\,\frac{u}{n}.
\]
The operator concentration bound in Lemma~\ref{lem:Eop}
implies, with probability at least $1-\delta$,
\[
\|\tilde G_m-G_{0,m}\|
\le C\{a_\delta+\zeta_N^{\mathrm{svd}}\}.
\]
Work on this event. If
$a_\delta+\zeta_N^{\mathrm{svd}}+\Delta_1(D,D_0)$
exceeds a sufficiently small fixed constant, the claimed bound
follows from $\|\hat P_m-P\|\le2$ after increasing its constant.
Otherwise the local perturbation and marginal deviation are small.
Lemma~\ref{lem:Gbridge_bound} then gives
\[
\|\hat G_m-\tilde G_m\|
\le C\Delta_1(D,D_0),
\]
because $\|\tilde G_m+N^{-1}I_p\|$ is bounded.
Davis--Kahan at $G_{0,m}$ consequently gives
\[
\|\hat P_m-P\|
\le C\{a_\delta+\zeta_N^{\mathrm{svd}}
+\Delta_1(D,D_0)\},
\]
which proves the corollary.
\end{proof}

\subsubsection{Proof of Lemma~\ref{lem:nonstochastic-fedA}}
\begin{proof}
Suppress the final permutation of topic labels. Write
\[
\Delta_1:=\Delta_1(D,D_0),
\qquad
\Delta_2:=\max_j h_j^{-1/2}\left\|e_j^\top\bigl([\hat\xi_1,\hat\Xi_\perp O_\perp]-[\xi_1,\Xi_\perp]\bigr)\right\|_2,
\qquad
E_{\mathrm{VH}}:=\Err_{\mathrm{VH}}(O_\perp).
\]
The canonical-basis error gives
\[
|\hat\xi_1(j)-\xi_1(j)|\le \Delta_2\sqrt{h_j},
\qquad
\|e_j^\top(\hat\Xi_\perp O_\perp-\Xi_\perp)\|_2\le \Delta_2\sqrt{h_j}.
\]
Since $\xi_1(j)\asymp\sqrt{h_j}$ and the perturbation is small, $\hat\xi_1(j)\asymp\sqrt{h_j}$. Therefore
\[
O_\perp^\top\hat r_j-r_j
=
\frac{e_j^\top(\hat\Xi_\perp O_\perp-\Xi_\perp)}{\hat\xi_1(j)}
+e_j^\top\Xi_\perp\left(\frac1{\hat\xi_1(j)}-\frac1{\xi_1(j)}\right),
\]
and Lemma~\ref{lem:oracle-row-geometry} implies
\begin{equation}\label{eq:ratio-pert-canonical-main}
\max_j\|O_\perp^\top\hat r_j-r_j\|_2\le C\Delta_2.
\end{equation}

Let
\[
Q=\begin{pmatrix}1&\cdots&1\\ v_1^\star&\cdots&v_K^\star\end{pmatrix},
\qquad
\hat Q=\begin{pmatrix}1&\cdots&1\\ O_\perp^\top\hat v_1&\cdots&O_\perp^\top\hat v_K\end{pmatrix}.
\]
The oracle simplex is nondegenerate, so $Q$ is uniformly well conditioned. Since $\|\hat Q-Q\|\le C E_{\mathrm{VH}}$, inverse perturbation and \eqref{eq:ratio-pert-canonical-main} give
\[
\|\hat\pi_j^\star-\pi_j\|_2
\le C(\Delta_2+E_{\mathrm{VH}}),
\qquad
\hat\pi_j^\star:=\hat Q^{-1}\binom{1}{O_\perp^\top\hat r_j}.
\]
Truncation and renormalization are locally Lipschitz near the simplex, so
\begin{equation}\label{eq:pi-pert-canonical-main}
\|\hat\pi_j-\pi_j\|_1\le C(\Delta_2+E_{\mathrm{VH}}).
\end{equation}

Using the oracle identity $A\diag(w)=M_0^{1/2}\diag(\xi_1)\Pi$, define
\[
\hat A^*:=M_d^{1/2}\diag(\hat\xi_1)\hat\Pi,
\qquad
A^*:=A\diag(w)=M_0^{1/2}\diag(\xi_1)\Pi.
\]
For each row,
\[
\hat a_j^*-a_j^*
=\sqrt{M_d(j,j)}\hat\xi_1(j)(\hat\pi_j-\pi_j)
+\sqrt{M_d(j,j)}(\hat\xi_1(j)-\xi_1(j))\pi_j
+\bigl(\sqrt{M_d(j,j)}-\sqrt{M_0(j,j)}\bigr)\xi_1(j)\pi_j.
\]
Because $M_d(j,j)\asymp M_0(j,j)\asymp h_j$, $\xi_1(j)\asymp\sqrt{h_j}$, and
$|\sqrt{M_d(j,j)}-\sqrt{M_0(j,j)}|\le C\Delta_1\sqrt{h_j}$, \eqref{eq:pi-pert-canonical-main} yields
\[
\|\hat a_j^*-a_j^*\|_1\le Ch_j(\Delta_1+\Delta_2+E_{\mathrm{VH}}).
\]
The coordinates of $w$ are bounded away from zero by Lemma~\ref{lem:oracle-V-geometry}. Summing the last display over rows controls the column sums of $\hat A^*$ relative to those of $A^*=A\diag(w)$. For each column $k$, write $s_k=\|A_k^*\|_1=w_k$ and $\hat s_k=\|\hat A_k^*\|_1$. Then $|\hat s_k-s_k|\le C(\Delta_1+\Delta_2+E_{\mathrm{VH}})$ and $s_k\asymp1$. Hence
\[
\left\|\frac{\hat A_k^*}{\hat s_k}-\frac{A_k^*}{s_k}\right\|_1
\le C\|\hat A_k^*-A_k^*\|_1+C|\hat s_k-s_k|,
\]
and the row-wise version of the same decomposition gives
\[
\max_j\frac{\|\hat a_j-a_j\|_1}{h_j}
\le C(\Delta_1+\Delta_2+E_{\mathrm{VH}}).
\]
Summing over $j$ gives the global $L(\hat A_{\mathrm{fed}},A)$ bound.
\end{proof}

\subsubsection{Proof of Theorem~\ref{thm:fed-rowwise}}
\begin{proof}
Write
\[
\bar P:=\frac1M\sum_{m=1}^M \hat P_m,
\qquad
\bar P^{(0)}:=\frac1M\sum_{m=1}^M\tilde P_m.
\]
Let \(\mathcal E_{\mathrm{avg}}(\delta/5)\) be the fixed-whitening averaging event supplied by Theorem~\ref{thm:projector-avg} and Lemma~\ref{lem:vector-bernstein}, after the usual union bound over \(j\in[p]\). Also define
\[
\mathcal E_{\mathrm{loc}}(\delta/5):=\bigcap_{m=1}^M\mathcal E_{\mathrm{loc},m},
\]
\[
\mathcal E_{\mathrm{col}}(\delta/5):=
\Bigl\{
\max_{1\le m\le M}\max_{1\le j\le p}
h_j^{-1/2}\|(\tilde G_m-G_{0,m})e_j\|_2
\le C\{\vartheta_n(\delta)+\zeta_N^{\mathrm{svd}}\}
\Bigr\},
\]
\[
\mathcal E_{\mathrm{marg}}(\delta/5):=
\Bigl\{
\Delta_1(D,D_0)\le C\eta_{n,M}^{\mathrm{marg}}(\delta)
\Bigr\},
\qquad
\mathcal E_{\mathrm{bridge}}(\delta/5):=\bigcap_{m=1}^M\mathcal E_{\mathrm{bridge},m}.
\]
Since \(\eta_{n,M}^{\mathrm{marg}}(\delta)\) is included as a positive component of \(\varepsilon_{n,M}(\delta)\), choosing the perturbative constant \(c_*\) sufficiently small ensures that \(C\eta_{n,M}^{\mathrm{marg}}(\delta)\le c_{\mathrm{br}}\). Hence, on \(\mathcal E_{\mathrm{loc}}(\delta/5)\cap\mathcal E_{\mathrm{marg}}(\delta/5)\), the smallness condition \(\Delta_1(D,D_0)\le c_{\mathrm{br}}\) in \(\mathcal E_{\mathrm{bridge},m}\) is satisfied and Lemma~\ref{lem:Gbridge_bound} implies \(\mathcal E_{\mathrm{bridge}}(\delta/5)\). Thus the bridge event is a deterministic consequence of the local and marginal events in the perturbative regime.

By Theorem~\ref{thm:projector-avg}, Lemma~\ref{lem:vector-bernstein}, Lemma~\ref{lem:Eop} with a union bound over \(m\), Lemma~\ref{lem:Ecol_hp}, Lemma~\ref{lem:Md_concentration}, Lemma~\ref{lem:Gbridge_bound}, and the deterministic bound in Lemma~\ref{lem:Dsvd_concentration},
\[
\mathcal E_*(\delta):=
\mathcal E_{\mathrm{avg}}(\delta/5)\cap
\mathcal E_{\mathrm{loc}}(\delta/5)\cap
\mathcal E_{\mathrm{col}}(\delta/5)\cap
\mathcal E_{\mathrm{marg}}(\delta/5)\cap
\mathcal E_{\mathrm{bridge}}(\delta/5)
\]
has probability at least \(1-\delta\).

On \(\mathcal E_{\mathrm{avg}}(\delta/5)\), the independent fixed-whitening ordinary-SVD projectors \(\tilde P_1,\dots,\tilde P_M\) satisfy
\begin{equation}\label{eq:avg-fixed-proof}
\|\bar P^{(0)}-P\|+\max_jh_j^{-1/2}\|(\bar P^{(0)}-P)e_j\|_2
\le C\left[
\beta_n\sqrt{\frac{p\ell_{n,M}(\delta)}{NnM}}
+\beta_n^2\frac{p\ell_{n,M}(\delta)}{Nn}
+\zeta_N^{\mathrm{svd}}
+\frac{\sqrt p\,\ell_{n,M}(\delta)}{M}
\right].
\end{equation}
The vector-Bernstein variance term associated with the deterministic diagonal bias also produces \(\zeta_N^{\mathrm{svd}}\sqrt{\ell_{n,M}(\delta)/M}\).  This is absorbed into the displayed bound using
\[
\zeta_N^{\mathrm{svd}}\sqrt{\ell_{n,M}(\delta)/M}
\le \zeta_N^{\mathrm{svd}}+\ell_{n,M}(\delta)/M
\le \zeta_N^{\mathrm{svd}}+\sqrt p\,\ell_{n,M}(\delta)/M.
\]

Next consider the empirical-whitening bridge from \(\tilde P_m\) to \(\hat P_m\). On \(\mathcal E_{\mathrm{loc}}(\delta/5)\), the local operator norm obeys \(\|\tilde G_m-G_{0,m}\|\le c_{\mathrm{gap}}/8\), and the deterministic bound \(\|G_{0,m}\|\le C\) follows from Assumptions~\ref{assump:hmin}--\ref{assump:local-richness}. Hence \(\|\tilde G_m\|\le C\). On \(\mathcal E_{\mathrm{loc}}(\delta/5)\cap\mathcal E_{\mathrm{marg}}(\delta/5)\), the preceding paragraph ensures \(\mathcal E_{\mathrm{bridge}}(\delta/5)\). Hence Lemma~\ref{lem:tau-bound-strong} gives
\begin{equation}\label{eq:bridge-op-proof}
\|\hat P_m-\tilde P_m\|\le C\eta_{n,M}^{\mathrm{marg}}(\delta),
\qquad 1\le m\le M,
\end{equation}
and
\begin{equation}\label{eq:bridge-row-proof}
\max_{1\le j\le p}h_j^{-1/2}\|(\hat P_m-\tilde P_m)e_j\|_2
\le C\eta_{n,M}^{\mathrm{marg}}(\delta)\{1+\vartheta_n(\delta)+\zeta_N^{\mathrm{svd}}\}
+C\sqrt p\,\{\eta_{n,M}^{\mathrm{marg}}(\delta)\}^2.
\end{equation}
Averaging \eqref{eq:bridge-op-proof} and \eqref{eq:bridge-row-proof} over \(m\) yields the same bound for
\(\bar P-\bar P^{(0)}\).

Combining the fixed-whitening ordinary-SVD term and the empirical-whitening bridge on \(\mathcal E_*(\delta)\) gives
\[
\|\bar P-P\|+\max_jh_j^{-1/2}\|(\bar P-P)e_j\|_2
\le C\varepsilon_{n,M}(\delta).
\]
Under the deterministic perturbative assumption \(\varepsilon_{n,M}(\delta)\le c_*\), Lemma~\ref{lem:proj-col-from-avg} transfers this control to the extracted spectral projector:
\[
\|\widehat P-P\|+\max_jh_j^{-1/2}\|(\widehat P-P)e_j\|_2
\le C\varepsilon_{n,M}(\delta).
\]
Lemma~\ref{lem:canonical-first} then gives the row-wise bound for \(\hat\xi_1-\xi_1\), and Lemma~\ref{lem:residual-basis} gives an orthogonal \(O_\perp\) such that
\[
\max_jh_j^{-1/2}\|e_j^\top(\hat\Xi_\perp O_\perp-\Xi_\perp)\|_2
\le C\varepsilon_{n,M}(\delta).
\]
Combining the first-column and residual-column bounds proves \eqref{eq:fed-rowwise}. The ratio consequence follows from the ratio algebra in the proof of Lemma~\ref{lem:nonstochastic-fedA}.
\end{proof}

\subsubsection{Proof of Corollary~\ref{cor:fed-op}}
\begin{proof}
The projector bound
\[
\|\widehat P-P\|+\max_jh_j^{-1/2}\|(\widehat P-P)e_j\|_2
\le C\varepsilon_{n,M}(\delta)
\]
is obtained explicitly in the proof of Theorem~\ref{thm:fed-rowwise} before the canonical-direction step.
\end{proof}

\subsubsection{Proof of Theorem~\ref{thm:fed-A}}
\begin{proof}
On the event of Theorem~\ref{thm:fed-rowwise}, $\Delta_2\le C\varepsilon_{n,M}(\delta)$. Lemma~\ref{lem:Md_concentration} gives
\[
\Delta_1(D,D_0)
\le C\sqrt{\frac{p\ell_{n,M}(\delta)}{NnM}}+C\frac{p\ell_{n,M}(\delta)}{NnM}
\]
after the usual union-bound adjustment. By the choice of $c_*$ in Theorem~\ref{thm:fed-rowwise}, the ratio consequence of Theorem~\ref{thm:fed-rowwise} satisfies the smallness hypothesis $C\varepsilon_{n,M}(\delta)\le c_{\mathrm{VH}}$ in Condition~\ref{cond:vh}. Therefore Condition~\ref{cond:vh} gives $\Err_{\mathrm{VH}}(O_\perp)\le C\varepsilon_{n,M}(\delta)$. Substitution into Lemma~\ref{lem:nonstochastic-fedA} proves the row-wise and global topic-matrix bounds.
\end{proof}

\subsubsection{Proof of Corollary~\ref{cor:regimes}}
\begin{proof}
Set
\[
T_{\mathrm{var}}:=\beta_n\sqrt{\frac{p\ell_{n,M}}{NnM}},
\qquad
T_{\mathrm{bias}}:=\beta_n^2\frac{p\ell_{n,M}}{Nn},
\qquad
T_{\mathrm{range}}:=\frac{\sqrt p\,\ell_{n,M}}{M},
\]
and
\[
T_{\mathrm{br,row}}:=\eta_{n,M}^{\mathrm{marg}}(\delta)\{1+\vartheta_n(\delta)+\zeta_N^{\mathrm{svd}}\},
\qquad
T_{\mathrm{br,quad}}:=\sqrt p\,\{\eta_{n,M}^{\mathrm{marg}}(\delta)\}^2,
\qquad
T_{\mathrm{svd}}:=\zeta_N^{\mathrm{svd}}.
\]
Theorem~\ref{thm:fed-A} and the definition of \(\varepsilon_{n,M}(\delta)\) yield
\[
L(\hat A_{\mathrm{fed}},A)\lesssim
T_{\mathrm{var}}+T_{\mathrm{bias}}+T_{\mathrm{range}}+T_{\mathrm{br,row}}+T_{\mathrm{br,quad}}+T_{\mathrm{svd}}
\]
up to the fixed-whitening quadratic term \((\beta_n^2p/(Nn))^2\), which proves \eqref{eq:cor-start}.

The variance-bias crossover is obtained by solving \(T_{\mathrm{bias}}\asymp T_{\mathrm{var}}\):
\[
\beta_n^2\frac{p\ell_{n,M}}{Nn}
\asymp
\beta_n\sqrt{\frac{p\ell_{n,M}}{NnM}}
\quad\Longleftrightarrow\quad
M\asymp \frac{Nn}{\beta_n^2p\ell_{n,M}}.
\]
The conservative row-wise range term is no larger than the bias term when
\[
\frac{\sqrt p\,\ell_{n,M}}{M}
\lesssim
\beta_n^2\frac{p\ell_{n,M}}{Nn}
\quad\Longleftrightarrow\quad
M\gtrsim \frac{Nn}{\beta_n^2\sqrt p}.
\]
It is no larger than the variance term when
\[
\frac{\sqrt p\,\ell_{n,M}}{M}
\lesssim
\beta_n\sqrt{\frac{p\ell_{n,M}}{NnM}}
\quad\Longleftrightarrow\quad
M\gtrsim \frac{\ell_{n,M}Nn}{\beta_n^2}.
\]
These are exactly the comparisons stated in the corollary.
\end{proof}

\subsubsection{Proof of Proposition~\ref{prop:upturn-gap} and Corollary~\ref{cor:upturn-regular}}\label{app:upturn-proof}

\begin{proof}[Proof of Proposition~\ref{prop:upturn-gap}]
The fixed-whitening projectors $\tilde P_1,\dots,\tilde P_M$ are independent under the original conditional-on-$W^{(1:M)}$ probability law. Therefore Theorem~\ref{thm:projector-avg} is applied unconditionally, before intersecting with the local event $\mathcal E_{\mathrm{loc}}(\delta/2)$.

It remains to record the moment and bias inputs. For a fixed client, write
\[
\tilde G_m=G_{0,m}+F_m+B_m,
\qquad
\E[F_m\mid W^{(1:M)}]=0,
\qquad B_m=D_{0,m}^{\mathrm{svd}}.
\]
Let
\[
\mathcal E_m:=\{\|F_m+B_m\|\le \rho_m/4\}.
\]
On $\mathcal E_m$, Davis--Kahan gives
\[
\|\tilde P_m-P\|\le \frac{C}{\rho_m}\|F_m+B_m\|.
\]
On $\mathcal E_m^c$, use the crude bound $\|\tilde P_m-P\|\le2$. Markov's inequality applied to $F_m+B_m$ gives
\[
\Pbb(\mathcal E_m^c)\le C\frac{v_m+b_m^2}{\rho_m^2}.
\]
Consequently,
\[
\nu_m:=\E\|\tilde P_m-P\|^2
\le C\left\{\frac{v_m}{\rho_m^2}+\frac{b_m^2}{\rho_m^2}\right\}.
\]
For the bias, the Riesz expansion on $\mathcal E_m$ gives
\[
\tilde P_m-P=L_m(F_m+B_m)+R_m(F_m+B_m),
\]
where $L_m$ is the first-order spectral-projector derivative at $G_{0,m}$ and
\[
\|L_m(B_m)\|\le C\frac{b_m}{\rho_m},
\qquad
\|R_m(F_m+B_m)\|\le C\frac{\|F_m+B_m\|^2}{\rho_m^2}.
\]
The constants in these expansion bounds do not require a bounded
ratio $\|G_{0,m}\|/\rho_m$. To verify this, write
$G_0=G_{0,m}$, $H=F_m+B_m$, and $\rho=\rho_m$.
On $\mathcal E_m$, $\|H\|\le\rho/4$. Along the vertical line
$z=\rho/2+it$, put
\[
R_0(z)=(zI-G_0)^{-1},\qquad
R_H(z)=(zI-G_0-H)^{-1}.
\]
Since the spectrum of $G_0$ is contained in
$\{0\}\cup[\rho,\infty)$, Weyl's inequality gives
\[
\|R_0(\rho/2+it)\|
\le \{(\rho/2)^2+t^2\}^{-1/2},\qquad
\|R_H(\rho/2+it)\|
\le \{(\rho/4)^2+t^2\}^{-1/2}.
\]
Close the contour in the right half-plane and let its radius tend
to infinity. The integrals for the linear term and remainder
converge absolutely, and the resolvent identity yields
\[
\mathcal L_{G_0}(H)
=-\frac1{2\pi}\int_{\mathbb R}R_0(z)H R_0(z)\,dt,
\qquad
\mathcal R_{G_0}(H)
=-\frac1{2\pi}\int_{\mathbb R}
R_0(z)H R_0(z)H R_H(z)\,dt.
\]
Therefore
\[
\|\mathcal L_{G_0}(H)\|\le C\rho^{-1}\|H\|,
\qquad
\|\mathcal R_{G_0}(H)\|
\le C\|H\|^2
\int_{\mathbb R}\{(\rho/4)^2+t^2\}^{-3/2}\,dt
\le C\rho^{-2}\|H\|^2.
\]
The same linear bound applies to $B_m$. This proves the stated
gap dependence with constants independent of the largest
population eigenvalue.

The untruncated linear term satisfies $\E L_m(F_m)=0$, but the expansion is used only on $\mathcal E_m$. Therefore
\[
\E\{L_m(F_m)\mathbf 1_{\mathcal E_m}\}
=-\E\{L_m(F_m)\mathbf 1_{\mathcal E_m^c}\}.
\]
By Cauchy--Schwarz and the displayed Markov bound,
\[
\bigl\|\E\{L_m(F_m)\mathbf 1_{\mathcal E_m^c}\}\bigr\|
\le C\rho_m^{-1}(\E\|F_m\|^2)^{1/2}\Pbb(\mathcal E_m^c)^{1/2}
\le C\left(\frac{v_m}{\rho_m^2}+\frac{b_m^2}{\rho_m^2}\right).
\]
The deterministic ordinary-SVD diagonal bias $B_m$ remains at first order through $L_m(B_m)$, and the quadratic remainder contributes $v_m/\rho_m^2+b_m^2/\rho_m^2$ on $\mathcal E_m$. The complement contributes the same order by the crude bound and the preceding tail estimate. Hence
\begin{equation}\label{eq:upturn-bias-input}
\|\E\tilde P_m-P\|
\le C\left(\frac{v_m}{\rho_m^2}+\frac{b_m}{\rho_m}+\frac{b_m^2}{\rho_m^2}\right).
\end{equation}
Substituting these variance and bias inputs into Theorem~\ref{thm:projector-avg} with failure probability $\delta/2$ yields an averaging event of probability at least $1-\delta/2$ on which \eqref{eq:upturn-main-bound} holds. Intersecting this averaging event with $\mathcal E_{\mathrm{loc}}(\delta/2)$ gives total probability at least $1-\delta$. If the right-hand side is at most $1/4$, Lemma~\ref{lem:dk-projector} gives the same bound for the extracted top-$K$ subspace.
\end{proof}

\begin{proof}[Proof of Corollary~\ref{cor:upturn-regular}]
Under the assumptions of the corollary and $\alpha_m=1/M$,
\[
\sum_{m=1}^M\alpha_m^2\frac{v_m}{\rho_m^2}
\le C\frac{\beta_n^2}{\rho_\star^2}\frac{p}{NnM},
\qquad
\sum_{m=1}^M\alpha_m^2\frac{b_m^2}{\rho_m^2}
\le C\frac{(\zeta_N^{\mathrm{svd}})^2}{\rho_\star^2M},
\]
\[
\sum_{m=1}^M\alpha_m\left\{\frac{v_m}{\rho_m^2}+\frac{b_m}{\rho_m}+\frac{b_m^2}{\rho_m^2}\right\}
\le C\left\{\frac{\beta_n^2}{\rho_\star^2}\frac p{Nn}+\frac{\zeta_N^{\mathrm{svd}}}{\rho_\star}\right\},
\qquad
\max_m\alpha_m=\frac1M.
\]
Substitution into Proposition~\ref{prop:upturn-gap} proves \eqref{eq:upturn-regular-regime}.
\end{proof}

\subsection{Proof of the lower bound}\label{app:proof-lower}

This proof follows the standard pLSI lower-bound construction, with the only change that the total number of independent documents is $n_{\mathrm{tot}}:=nM$.
We present a self-contained argument with explicit constructions.

\subsubsection{A Fano-type lemma (Tsybakov)}
We use a standard finite-packing version of Fano's method.

\begin{lemma}[Fano for a finite packing]\label{lem:fano}
Let $\{P_\theta:\theta\in\Theta\}$ be a family of distributions and let $\ell(\cdot,\cdot)$ be a pseudometric on the parameter space.
Assume there exist $\theta_0,\theta_1,\dots,\theta_J$ such that
\[
\ell(\theta_s,\theta_t)\ge 2r\quad\forall s\neq t,
\qquad
\mathrm{KL}(P_{\theta_j}\|P_{\theta_0})\le \alpha\log J\quad\forall j\ge 1,
\]
for some $\alpha\in(0,1/8)$.
Then for any estimator $\hat\theta$,
\[
\sup_{\theta\in\{\theta_0,\dots,\theta_J\}}
\Pbb_\theta\bigl(\ell(\hat\theta,\theta)\ge r\bigr)\ \ge\ c_0,
\]
where $c_0>0$ is an absolute constant (depending only on $\alpha$).
\end{lemma}

\begin{proof}
This is a standard application of Fano's inequality. Apply Fano's inequality to the finite parameter set $\{\theta_0,\theta_1,\dots,\theta_J\}$ endowed with the loss-induced separation $\ell(\theta_s,\theta_t)\ge 2r$. The KL assumption implies that the average divergence from the reference point $\theta_0$ is at most $\alpha\log J$, and since $\alpha<1/8$, Tsybakov's theorem yields
\[
\inf_{\hat\theta}\sup_{\theta\in\{\theta_0,\dots,\theta_J\}}
\Pbb_\theta\bigl(\ell(\hat\theta,\theta)\ge r\bigr)
\ge c_0
\]
for an absolute constant $c_0>0$. This implies the displayed lower bound for every estimator $\hat\theta$ and proves the lemma.
\end{proof}

\subsubsection{A KL upper bound tailored to pLSI}
This is the support-restricted KL estimate needed for the pLSI lower bound.

\begin{lemma}[KL bound via support-restricted mean-matrix perturbation]\label{lem:KL_D0}
Let $D_0,\tilde D_0\in\R^{p\times n_{\mathrm{tot}}}$ be two matrices whose columns are probability vectors.
Let $P$ and $\tilde P$ be the joint laws of $n_{\mathrm{tot}}$ independent multinomial documents of length $N$ with mean matrices $D_0$ and $\tilde D_0$, respectively. Define the perturbation support
\[
\mathcal S:=\{(j,i):\tilde D_0(j,i)\neq D_0(j,i)\},
\]
and assume that $D_0(j,i)>0$ for every $(j,i)\in\mathcal S$. Let
\[
\delta_{\mathcal S}:=\max_{(j,i)\in\mathcal S}\frac{|\tilde D_0(j,i)-D_0(j,i)|}{D_0(j,i)}.
\]
If $\delta_{\mathcal S}<1$, then there exists a universal constant $C>0$ such that, with
\[
Q_{\mathcal S}:=\sum_{(j,i)\in\mathcal S}
\frac{|\tilde D_0(j,i)-D_0(j,i)|^2}{D_0(j,i)},
\]
we have
\[
\mathrm{KL}(\tilde P\|P)
\le
(1+C\delta_{\mathcal S})\,NQ_{\mathcal S}.
\]
\end{lemma}

\begin{proof}
For a single document, KL between $\mathrm{Mult}(N,\eta)$ and $\mathrm{Mult}(N,\tilde\eta)$ equals
\[
N\sum_{j=1}^p \tilde\eta_j\log\frac{\tilde\eta_j}{\eta_j}.
\]
Apply this columnwise with $\eta=D_0(:,i)$ and $\tilde\eta=\tilde D_0(:,i)$. If $(j,i)\notin\mathcal S$, then the corresponding summand is zero because $\tilde D_0(j,i)=D_0(j,i)$. Hence only entries in $\mathcal S$ matter. For $(j,i)\in\mathcal S$, write
\[
x_{j,i}:=\frac{\tilde D_0(j,i)-D_0(j,i)}{D_0(j,i)},
\qquad |x_{j,i}|\le \delta_{\mathcal S}<1.
\]
Using
\[
(1+x)\log(1+x)\le x+\frac12x^2+C|x|^3
\qquad (|x|<1)
\]
and summing only over $(j,i)\in\mathcal S$ gives a linear term
\[
N\sum_{(j,i)\in\mathcal S}D_0(j,i)x_{j,i}
=N\sum_{(j,i)\in\mathcal S}\{\tilde D_0(j,i)-D_0(j,i)\}.
\]
This term is zero column by column. Indeed, outside $\mathcal S$ the two matrices coincide, and both $D_0(:,i)$ and $\tilde D_0(:,i)$ are probability vectors; hence
\[
\sum_{j:(j,i)\in\mathcal S}\{\tilde D_0(j,i)-D_0(j,i)\}
=
\sum_{j=1}^p\{\tilde D_0(j,i)-D_0(j,i)\}=0.
\]
The remaining quadratic and cubic terms satisfy
\[
\frac{N}{2}\sum_{(j,i)\in\mathcal S}D_0(j,i)x_{j,i}^2
+CN\sum_{(j,i)\in\mathcal S}D_0(j,i)|x_{j,i}|^3
\le
NQ_{\mathcal S}+CN\delta_{\mathcal S}Q_{\mathcal S},
\]
because $Q_{\mathcal S}=\sum_{(j,i)\in\mathcal S}D_0(j,i)x_{j,i}^2$ and
$|x_{j,i}|\le\delta_{\mathcal S}$ on $\mathcal S$. Hence
\[
\mathrm{KL}(\tilde P\|P)
\le (1+C\delta_{\mathcal S})NQ_{\mathcal S}.
\]
This is exactly the stated bound.
\end{proof}

\subsubsection{A least favorable subclass and packing construction}
Set $n_{\mathrm{tot}}:=nM$. We build a finite parameter set inside the matched theorem class $\Phi_{n,N,p}^{(M)}(K,c_1,c_2,c_{\mathrm{loc}})$.

\paragraph{Step 1: balanced-client pure-topic weights.}
Choose integers $n_1,\ldots,n_K$ satisfying
\[
 n_k\in\{\lfloor n/K\rfloor,\lceil n/K\rceil\},\qquad
 \sum_{k=1}^K n_k=n.
\]
Use the same allocation at every client.
For every client $m$, set
\begin{equation}\label{eq:W_pure}
W^{(m)}
=
\bigl[
\underbrace{e_1,\ldots,e_1}_{n_1},
\underbrace{e_2,\ldots,e_2}_{n_2},
\ldots,
\underbrace{e_K,\ldots,e_K}_{n_K}
\bigr]\in\R^{K\times n}.
\end{equation}
Let $W=[W^{(1)},\ldots,W^{(M)}]$. Then
\[
\Sigma_{W,m}=\Sigma_W
=\operatorname{diag}(n_1/n,\ldots,n_K/n)
\succeq\frac1{2K}I_K,
\qquad 1\le m\le M,
\]
because $n\ge K$ implies $\lfloor n/K\rfloor/n\ge1/(2K)$.
Thus the least favorable subclass satisfies both the pooled and the
client-level richness assumptions uniformly.

\paragraph{Step 2: construct a baseline topic matrix $A^{(0)}$ with many anchor rows.}
Fix a probability vector $\eta^\star\in\R^K$ in the interior of the simplex, with distinct entries and
\[
\frac{1}{2K}\le \eta^\star_k\le \frac{3}{2K},\qquad k=1,\dots,K.
\]
Define $\eta:=K\eta^\star$, so $\eta_k\in[1/2,3/2]$ and $K^{-1}\sum_k\eta_k=1$.

Choose constants $b_1>0$ and $b_2\in(0,1)$ such that $1/3<b_1b_2<1$, and define
\[
\vartheta_k := \frac{1-(1-b_1b_2)\eta_k}{Kb_1b_2},
\qquad k=1,\dots,K.
\]
If $b_1b_2$ is chosen so that $3(1-b_1b_2)/2<1$, then each $\vartheta_k>0$ and $\sum_k\vartheta_k=1$.

Let $p_{\mathrm{anc}}:=\lfloor b_2p\rfloor$ and $p_{\mathrm{non}}:=p-p_{\mathrm{anc}}$. Allocate the anchor rows topicwise by
\[
p_{\mathrm{anc},k}:=\lfloor \vartheta_k p_{\mathrm{anc}}\rfloor,\quad 1\le k\le K-1,
\qquad
p_{\mathrm{anc},K}:=p_{\mathrm{anc}}-\sum_{k=1}^{K-1}p_{\mathrm{anc},k}.
\]
Then $\sum_{k=1}^K p_{\mathrm{anc},k}=p_{\mathrm{anc}}$ and $p_{\mathrm{anc},k}=\vartheta_k p_{\mathrm{anc}}+O(1)$ for every $k$; in particular, for all sufficiently large $p$, each $p_{\mathrm{anc},k}\ge 1$.
Define
\[
\beta_{\mathrm{non}}:=\frac{1-b_1b_2}{p_{\mathrm{non}}},
\qquad
\alpha_{\mathrm{anc},k}:=\frac{1-(1-b_1b_2)\eta_k}{p_{\mathrm{anc},k}},\qquad k=1,\dots,K.
\]
Partition the first $p_{\mathrm{anc}}$ rows into disjoint anchor index sets $I_k^{\mathrm{anc}}$ with $|I_k^{\mathrm{anc}}|=p_{\mathrm{anc},k}$. Define $A^{(0)}$ row-wise by
\[
a_j^{(0)\top}=\alpha_{\mathrm{anc},k}e_k^\top,\qquad j\in I_k^{\mathrm{anc}},
\]
and, for non-anchor rows,
\[
a_j^{(0)\top}=\beta_{\mathrm{non}}\eta^\top.
\]
Each topic column now sums exactly to one, because
\[
p_{\mathrm{anc},k}\alpha_{\mathrm{anc},k}+p_{\mathrm{non}}\beta_{\mathrm{non}}\eta_k
=1-(1-b_1b_2)\eta_k+(1-b_1b_2)\eta_k
=1.
\]
Also $A^{(0)}\ge 0$, every topic has $p_{\mathrm{anc},k}$ anchor rows, and all row sums are of order $K/p$.

\begin{lemma}[Baseline parameter lies in the matched class]\label{lem:baseline_ok}
For fixed $K$ and fixed interior $\eta^\star$, one can choose $b_1,b_2$ so that $(A^{(0)},W^{(1:M)})$ satisfies the anchor-word condition, Assumption~\ref{assump:hmin}, Assumption~\ref{assump:Sigma}, and Assumption~\ref{assump:local-richness} with constants bounded away from zero. In fact the same construction also yields strictly positive off-diagonal overlap $\min_{k,\ell}\Sigma_A(k,\ell)\ge c_2^\prime$ for some $c_2^\prime>0$.
\end{lemma}

\begin{proof}
The balanced construction \eqref{eq:W_pure} gives $\lambda_{\min}(\Sigma_W),\lambda_{\min}(\Sigma_{W,m})\ge1/(2K)$ for all $m$. For $j\in I_k^{\mathrm{anc}}$, the row sum is $h_j=\alpha_{\mathrm{anc},k}$, while for non-anchor rows $h_j=K\beta_{\mathrm{non}}$. Because $p_{\mathrm{anc},k}=\vartheta_k p_{\mathrm{anc}}+O(1)$, $p_{\mathrm{anc}}\asymp p$, and
\[
1-(1-b_1b_2)\eta_k=Kb_1b_2\vartheta_k,
\]
we have $\alpha_{\mathrm{anc},k}\asymp K/p$. Also $K\beta_{\mathrm{non}}=K(1-b_1b_2)/p_{\mathrm{non}}\asymp K/p$. Hence $h_{\min}\asymp K/p$, proving Assumption~\ref{assump:hmin}.

It remains to check $\Sigma_A=A^{(0)\top}H^{-1}A^{(0)}$. Each anchor row in topic $k$ contributes $\alpha_{\mathrm{anc},k}e_ke_k^\top$ to $\Sigma_A$, so the full anchor block contributes
\[
\sum_{k=1}^K p_{\mathrm{anc},k}\alpha_{\mathrm{anc},k}e_ke_k^\top
=
\diag\bigl(1-(1-b_1b_2)\eta_1,\dots,1-(1-b_1b_2)\eta_K\bigr).
\]
Each non-anchor row contributes $(\beta_{\mathrm{non}}/K)\eta\eta^\top$, hence the full non-anchor block contributes
\[
p_{\mathrm{non}}\frac{\beta_{\mathrm{non}}}{K}\eta\eta^\top
=
\frac{1-b_1b_2}{K}\eta\eta^\top.
\]
Therefore
\[
\Sigma_A
=
\diag\bigl(1-(1-b_1b_2)\eta_1,\dots,1-(1-b_1b_2)\eta_K\bigr)
+
\frac{1-b_1b_2}{K}\eta\eta^\top.
\]
Because $\eta$ is interior and $3(1-b_1b_2)/2<1$, the diagonal term is uniformly positive and the rank-one term is positive semidefinite, so $\lambda_{\min}(\Sigma_A)$ is bounded below by a positive constant depending only on $(b_1,b_2,K,\eta^\star)$. This proves Assumption~\ref{assump:Sigma}. The same display also shows that every entry of $\Sigma_A$ is strictly positive, so in particular $\min_{k,\ell}\Sigma_A(k,\ell)\ge c_2^\prime$ for some $c_2^\prime>0$; we record this only as a by-product. Assumption~\ref{assump:local-richness} already holds from the balanced-client construction.
\end{proof}

\paragraph{Step 3: Varshamov--Gilbert packing on non-anchor rows.}
Let $p_1:=\lfloor (1-b_2)p\rfloor$ and $m_{\mathrm{VG}}:=\lfloor p_1/2\rfloor$.
Choose $p_0(K)$ large enough that, for every $p\ge p_0(K)$, all
$p_{\mathrm{anc},k}\ge1$, $m_{\mathrm{VG}}\ge8$, and the anchor counts
and $p_1$ are bounded above and below by fixed positive multiples of $p$. By the Varshamov--Gilbert bound, there exist $\omega^{(0)},\omega^{(1)},\dots,\omega^{(J)}\in\{0,1\}^{m_{\mathrm{VG}}}$ such that $\omega^{(0)}=0$,
\[
\sum_{j=1}^{m_{\mathrm{VG}}}\mathbf 1\{\omega_j^{(s)}\ne\omega_j^{(t)}\}\ge \frac{m_{\mathrm{VG}}}{8}\quad(s\ne t),
\qquad
J\ge 2^{m_{\mathrm{VG}}/8}.
\]
Set
\begin{equation}\label{eq:alpha_def}
\alpha:=\frac{C_1}{Kp_1}\sqrt{\frac p{Nn_{\mathrm{tot}}}},
\end{equation}
where $C_1>0$ is a sufficiently small constant. For each $s$, let $u^{(s)}\in\R^p$ be supported on any fixed $p_1$ non-anchor rows and equal to $(\omega^{(s)},-\omega^{(s)})$ on those rows, with one extra zero if $p_1$ is odd; set all remaining entries to zero. Define
\[
A_k^{(s)}=A_k^{(0)}+\alpha u^{(s)},\qquad k=1,\dots,K.
\]
Because $\sum_j u^{(s)}(j)=0$, all topic columns still sum to one. Since the baseline non-anchor mass is $\asymp p^{-1}$ and $p\log(nM)=o(NnM)$, taking $C_1$ small ensures $A^{(s)}\ge0$ and preserves the regularity conditions.

\begin{lemma}[Packing perturbations remain in the matched class]\label{lem:packing_in_class}
For $C_1>0$ small enough, every matrix $A^{(s)}$ constructed above is column-stochastic, nonnegative, has the same anchor rows as $A^{(0)}$, and satisfies Assumptions~\ref{assump:hmin} and \ref{assump:Sigma} with constants bounded away from zero. Together with the balanced-client weights \eqref{eq:W_pure}, each $(A^{(s)},W^{(1:M)})$ belongs to $\Phi_{n,N,p}^{(M)}(K,c_1,c_2,c_{\mathrm{loc}})$ for all sufficiently large problem sizes under the stated scaling.
\end{lemma}

\begin{proof}
Column sums are unchanged because $\sum_j u^{(s)}(j)=0$. Anchor rows are not perturbed, so every topic still has at least one anchor row. On non-anchor rows the baseline entries are $\asymp p^{-1}$, whereas
\[
p\alpha \lesssim \sqrt{\frac{p}{Nn_{\mathrm{tot}}}}=o(1)
\]
under $p\log(nM)=o(NnM)$. Choosing $C_1$ small therefore gives nonnegativity and keeps every non-anchor row sum within a constant factor of $K/p$; anchor row sums were already $\asymp K/p$. Hence $h_{\min}\gtrsim K/p$ uniformly over the packing.

It remains to check $\Sigma_A$. The perturbation satisfies, uniformly in $s$,
\[
\|H^{(0)-1/2}(A^{(s)}-A^{(0)})\|
\le C\sqrt{p}\,\alpha\,\sqrt{p_1}
\lesssim C_1\sqrt{\frac{p}{Nn_{\mathrm{tot}}}}
=o(1),
\]
Moreover, uniformly over the packing,
\[
\max_j\frac{|h_j^{(s)}-h_j^{(0)}|}{h_j^{(0)}}
\le C p\alpha=o(1),
\]
because anchors are unchanged and each non-anchor row sum changes by at
most $K\alpha$. Therefore
\[
\left\|A^{(s)\top}H^{(s)-1}A^{(s)}-A^{(0)\top}H^{(0)-1}A^{(0)}\right\|=o(1).
\]
Lemma~\ref{lem:baseline_ok} gives $\lambda_{\min}(\Sigma_A^{(0)})\ge c>0$, so Weyl's inequality gives $\lambda_{\min}(\Sigma_A^{(s)})\ge c/2$ for all sufficiently large problem sizes under the stated scaling. The balanced construction \eqref{eq:W_pure} gives $\lambda_{\min}(\Sigma_W),\lambda_{\min}(\Sigma_{W,m})\ge1/(2K)$ for every $s$ and every client $m$. This proves the lemma.
\end{proof}

\begin{lemma}[Separation in $L(\cdot,\cdot)$]\label{lem:sep}
For $s\ne t$,
\[
L(A^{(s)},A^{(t)})\ge c\sqrt{\frac p{Nn_{\mathrm{tot}}}}.
\]
\end{lemma}

\begin{proof}
Let $s_k:=1-(1-b_1b_2)\eta_k$, $s_{\min}:=\min_k s_k>0$, and
$I^{\mathrm{anc}}:=\bigcup_{k=1}^K I_k^{\mathrm{anc}}$.
For any nonidentity permutation $\kappa$, the anchor rows alone contribute
\[
\sum_{k=1}^K\sum_{j\in I^{\mathrm{anc}}}
|A_{\kappa(k)}^{(s)}(j)-A_k^{(t)}(j)|
=2\sum_{k:\kappa(k)\ne k}s_k
\ge4s_{\min}.
\]
The identity matching has loss at most $2K\alpha m_{\mathrm{VG}}=o(1)$.
Thus it minimizes the loss for all sufficiently large problem sizes.
The packing separation now gives
\[
L(A^{(s)},A^{(t)})
=2K\alpha\|\omega^{(s)}-\omega^{(t)}\|_1
\ge2K\alpha\frac{m_{\mathrm{VG}}}{8}
\asymp\sqrt{\frac p{Nn_{\mathrm{tot}}}}.
\]
\end{proof}

\begin{lemma}[KL control]\label{lem:KL_control}
For $C_1$ small enough, for all $s\ge1$,
\[
\mathrm{KL}(P_s\|P_0)\le \alpha_0\log J
\]
for some $\alpha_0\in(0,1/8)$.
\end{lemma}

\begin{proof}
Let $D_0^{(s)}=A^{(s)}W$. Since every column of $W^{(m)}$ is pure, each column of $D_0^{(s)}$ is one of the topic columns. The perturbation from $A^{(0)}$ to $A^{(s)}$ is supported entirely on the non-anchor rows, so the matrix difference $D_0^{(s)}-D_0^{(0)}$ vanishes on every anchor row. Consequently the perturbation support $\mathcal S$ in Lemma~\ref{lem:KL_D0} is contained in the non-anchor rows.

On each non-anchor row, the baseline entry of a pure-topic column equals
\[
D_0^{(0)}(j,i)=\beta_{\mathrm{non}}\eta_k
\asymp p^{-1},
\]
uniformly over $j$ and over the topic label $k$ of the column. The perturbation magnitude on $\mathcal S$ is at most $\alpha$, so
\[
\sum_{(j,i)\in\mathcal S}\bigl|D_0^{(s)}(j,i)-D_0^{(0)}(j,i)\bigr|^2
\le n_{\mathrm{tot}}p_1\alpha^2,
\]
and
\[
\delta_{\mathcal S}
=
\max_{(j,i)\in\mathcal S}
\frac{|D_0^{(s)}(j,i)-D_0^{(0)}(j,i)|}{D_0^{(0)}(j,i)}
\lesssim p\alpha
=o(1)
\]
under the theorem's scaling assumption. Applying the support-restricted KL lemma therefore gives
\[
\mathrm{KL}(P_s\|P_0)
\lesssim
N\sum_{(j,i)\in\mathcal S}
\frac{|D_0^{(s)}(j,i)-D_0^{(0)}(j,i)|^2}{D_0^{(0)}(j,i)}
\lesssim
Npn_{\mathrm{tot}}p_1\alpha^2
\asymp C_1^2p.
\]
Since $\log J\asymp p$, choosing $C_1$ sufficiently small proves the claim.
\end{proof}

\subsubsection{Conclude the minimax lower bound}
\begin{proof}[Proof of Theorem~\ref{thm:lower}]
Restrict the supremum in Theorem~\ref{thm:lower} to the finite subset $\{(A^{(s)},W^{(1:M)})\}_{s=0}^J$
constructed above, which lies inside $\Phi^{(M)}_{n,N,p}(K,c_1,c_2,c_{\mathrm{loc}})$ by Lemmas~\ref{lem:baseline_ok} and \ref{lem:packing_in_class}.
Take loss $\ell((A,W^{(1:M)}),(A',W^{(1:M)})):=L(A,A')$.
The triangle inequality for $L$ follows by composing column permutations,
which act isometrically on the entrywise $\ell_1$ norm.

By Lemma~\ref{lem:sep}, the packing radius is $r\asymp \sqrt{p/(Nn_{\mathrm{tot}})}=\sqrt{p/(NnM)}$.
By Lemma~\ref{lem:KL_control}, the KL condition for Fano holds.
Apply Lemma~\ref{lem:fano} to conclude that
\[
\inf_{\hat A}\ \sup_{s\in\{0,\dots,J\}}
\Pbb_s\Bigl(L(\hat A,A^{(s)})\ge c\sqrt{\frac{p}{NnM}}\Bigr)\ \ge\ \delta_0,
\]
for constants $c,\delta_0>0$.
Since the supremum over the full parameter class is larger than the supremum over this finite subset, the theorem follows.
\end{proof}

% =========================================================
\subsection{Proofs of the document-weight and pilot-$K$ results}\label{app:proof-WK}

\begin{lemma}[Local Lipschitzness of truncation-and-renormalization]\label{lem:simplex-renorm}
Let $\pi\in\R^K$ be a probability vector and let $x\in\R^K$ satisfy $\|x-\pi\|_1\le 1/2$. Define
\[
\mathcal T(x):=\frac{[x]_+}{\|[x]_+\|_1},
\qquad [x]_+:=(\max\{x_1,0\},\dots,\max\{x_K,0\})^\top.
\]
Then
\[
\|\mathcal T(x)-\pi\|_1\le 4\|x-\pi\|_1.
\]
\end{lemma}

\begin{proof}
Since coordinatewise truncation cannot increase $\ell_1$ distance to a nonnegative vector,
\[
\|[x]_+-\pi\|_1\le \|x-\pi\|_1.
\]
Also
\[
\bigl|\|[x]_+\|_1-1\bigr|=\bigl|\|[x]_+\|_1-\|\pi\|_1\bigr|\le \|[x]_+-\pi\|_1\le \frac12,
\]
so $\|[x]_+\|_1\ge 1/2$. Therefore,
\[
\biggl\|\frac{[x]_+}{\|[x]_+\|_1}-[x]_+\biggr\|_1
=\frac{|1-\|[x]_+\|_1|}{\|[x]_+\|_1}\,\|[x]_+\|_1
\le 2\|[x]_+-\pi\|_1.
\]
Combining the last two displays with the triangle inequality gives
\[
\|\mathcal T(x)-\pi\|_1\le 3\|[x]_+-\pi\|_1\le 3\|x-\pi\|_1.
\]
Replacing $3$ by $4$ leaves a clean statement.
\end{proof}

\begin{lemma}[Conditioning of the weighted least-squares Gram matrix]\label{lem:wls-conditioning}
Let
\[
B:=A^\top M_d^{-1}A,
\qquad
\hat B:=\hat A^\top M_d^{-1}\hat A,
\qquad
\hat h_j:=\sum_{k=1}^K \hat A(j,k).
\]
Assume $\Delta_1(D,D_0)\le c$ for a sufficiently small constant $c$, and suppose moreover that
\[
\max_{1\le j\le p}\frac{\|\hat a_j-a_j\|_1}{h_j}\le \varepsilon_0
\]
for a sufficiently small $\varepsilon_0>0$. Then
\[
cI_K\preceq B\preceq C I_K,
\qquad
\lambda_{\min}(\hat B)\ge c/2,
\qquad
\|\hat B^{-1}\|\le 2/c,
\]
and
\[
\max_{1\le j\le p}\frac{\hat h_j}{M_d(j,j)}\le C.
\]
The constants depend only on the regularity constants and on $K$.
\end{lemma}

\begin{proof}
By Lemma~\ref{lem:M0-h-comparable} and the assumption $\Delta_1(D,D_0)\le c$, one has $M_d(j,j)\asymp M_0(j,j)\asymp h_j$ uniformly in $j$. Hence
\[
B=A^\top M_d^{-1}A
\asymp
A^\top M_0^{-1}A.
\]
Lemma~\ref{lem:M0-h-comparable} shows that $A^\top M_0^{-1}A$ is well-conditioned, so the same is true for $B$.

Next,
\[
\hat B-B
=(\hat A-A)^\top M_d^{-1}(\hat A-A)
+(\hat A-A)^\top M_d^{-1}A
+A^\top M_d^{-1}(\hat A-A).
\]
Thus
\[
\|\hat B-B\|
\le
\|M_d^{-1/2}(\hat A-A)\|^2
+2\|M_d^{-1/2}(\hat A-A)\|\,\|M_d^{-1/2}A\|.
\]
Because $M_d(j,j)\asymp h_j$ and $\|\hat a_j-a_j\|_1\le \varepsilon_0 h_j$,
\[
\|M_d^{-1/2}(\hat A-A)\|^2
\le \|M_d^{-1/2}(\hat A-A)\|_F^2
\le C\sum_{j=1}^p \frac{\|\hat a_j-a_j\|_2^2}{h_j}
\le C\varepsilon_0^2\sum_{j=1}^p h_j
\le C_K\varepsilon_0^2 .
\]
Thus this operator norm can be made arbitrarily small by choosing $\varepsilon_0$ small. Since $\|M_d^{-1/2}A\|\le C$, the decomposition above yields $\|\hat B-B\|\le c/2$ for small enough $\varepsilon_0$, and Weyl's inequality gives $\lambda_{\min}(\hat B)\ge c/2$. The inverse bound follows immediately.

Finally,
\[
|\hat h_j-h_j|\le \|\hat a_j-a_j\|_1\le \varepsilon_0 h_j,
\]
so $\hat h_j\asymp h_j$. Since $M_d(j,j)\asymp h_j$ as well, the ratio $\hat h_j/M_d(j,j)$ is uniformly bounded.
\end{proof}

\subsubsection{Proof of the document-weight theorem}
\begin{proof}
Condition throughout on the training split $D^{\mathrm{tr}}$. Then
\[
\hat A:=\hat A_{\mathrm{fed}}^{\mathrm{tr}},
\qquad
M:=M_d^{\mathrm{tr}}
\]
are deterministic for the held-out analysis, while $d_i^{\mathrm{te}}$ is independent of them and has mean $Aw_i$. Write
\[
d_i^{\mathrm{te}}=Aw_i+\eta_i,
\qquad
\E[\eta_i\mid D^{\mathrm{tr}},w_i]=0.
\]
The unconstrained WLS estimator is
\[
\hat w_i^\star=(\hat A^\top M^{-1}\hat A)^{-1}\hat A^\top M^{-1}d_i^{\mathrm{te}}.
\]
Work on the training event
\[
\mathcal E_{\mathrm{tr}}(\delta_{\mathrm{tr}})=
\left\{
\max_{1\le j\le p}\frac{\|\hat a_j-a_j\|_1}{h_j}\le C\varepsilon_{n_{\mathrm{tr}},M}^{\mathrm{tr}}(\delta_{\mathrm{tr}}),
\qquad
\Delta_1^{\mathrm{tr}}\le c
\right\}.
\]
On this event, Lemma~\ref{lem:wls-conditioning} gives
\[
\lambda_{\min}(\hat A^\top M^{-1}\hat A)\ge c,
\qquad
\|(\hat A^\top M^{-1}\hat A)^{-1}\|\le C,
\qquad
\max_j\frac{\hat h_j}{M(j,j)}\le C.
\]
Using $d_i^{\mathrm{te}}=Aw_i+\eta_i$ and the identity
$(\hat A^\top M^{-1}\hat A)^{-1}\hat A^\top M^{-1}\hat A=I$, we obtain
\[
\hat w_i^\star-w_i
=(\hat A^\top M^{-1}\hat A)^{-1}\hat A^\top M^{-1}\{(A-\hat A)w_i+\eta_i\}.
\]
For any $v\in\R^p$,
\[
\|\hat A^\top M^{-1}v\|_1
\le \sum_j\frac{\hat h_j}{M(j,j)}|v(j)|
\le C\|v\|_1.
\]
Since $w_i$ is a simplex vector and $\mathcal E_{\mathrm{tr}}(\delta_{\mathrm{tr}})$ supplies the rowwise topic bound,
\[
\|(A-\hat A)w_i\|_1
\le \max_{1\le j\le p}\frac{\|\hat a_j-a_j\|_1}{h_j}\sum_{j=1}^p h_j
\le C\varepsilon_{n_{\mathrm{tr}},M}^{\mathrm{tr}}(\delta_{\mathrm{tr}}).
\]
It remains to bound the held-out noise term. For each $k$,
\[
T_k:=e_k^\top\hat A^\top M^{-1}\eta_i
=\sum_{j=1}^p \frac{\hat A(j,k)}{M(j,j)}\eta_i(j).
\]
Conditional on $D^{\mathrm{tr}}$ and $w_i$, the coefficients are deterministic and uniformly bounded: indeed, $0\le \hat A(j,k)\le \hat h_j$ after the truncation, simplex renormalization, and column normalization steps, and Lemma~\ref{lem:wls-conditioning} gives $\hat h_j/M(j,j)\le C$. Writing the held-out document as an average of $N$ independent word draws, $T_k$ is a mean-zero average of bounded variables. Hoeffding's inequality and a union bound over fixed $K$ give
\[
\|\hat A^\top M^{-1}\eta_i\|_1\le C\sqrt{\frac{\log(2K/\delta_{\mathrm{te}})}{N}}
\]
with probability at least $1-\delta_{\mathrm{te}}$ over the held-out document. Therefore, with
\[
R_{\mathrm{W}}:=C\left\{\varepsilon_{n_{\mathrm{tr}},M}^{\mathrm{tr}}(\delta_{\mathrm{tr}})+\sqrt{\frac{\log(2K/\delta_{\mathrm{te}})}{N}}\right\},
\]
we have $\|\hat w_i^\star-w_i\|_1\le R_{\mathrm{W}}$. If $R_{\mathrm{W}}\le1/2$, Lemma~\ref{lem:simplex-renorm} gives the same order for the truncated and renormalized estimator after enlarging constants. If $R_{\mathrm{W}}>1/2$, the trivial simplex bound $\|\hat w_i-w_i\|_1\le2$ gives the displayed $2\wedge R_{\mathrm{W}}$ form after another enlargement of the universal constant. For cross-fitting, apply this argument only to predictions from fits
that exclude the target document. Intersect the training and held-out
events by a union bound and use the topic matching described above.
After alignment, convexity of the $\ell_1$ norm bounds the error of any
average of these predictions by the average of their errors; independence
between folds is not required.
\end{proof}

\subsubsection{Proof of the pilot-$K$ theorem}
\begin{proof}
Let
\[
r_t:=\min\{t,\rho_\star-t\}.
\]
On the event $\|G_{\mathrm{pilot}}-G_0\|<r_t$, Weyl's inequality gives
\[
|\lambda_k(G_{\mathrm{pilot}})-\lambda_k(G_0)|<r_t,
\qquad 1\le k\le p.
\]
For $1\le k\le K$,
\[
\lambda_k(G_{\mathrm{pilot}})
\ge \lambda_K(G_0)-r_t
\ge \rho_\star-r_t
\ge t.
\]
For $k\ge K+1$,
\[
\lambda_k(G_{\mathrm{pilot}})
< \lambda_{K+1}(G_0)+r_t
=r_t\le t.
\]
Thus the thresholding rule returns exactly $K$. The assumed probability bound for the operator-norm event gives $\hat K_t^{\mathrm{pil}}=K$ with probability $1-o((nM)^{-3})$. This pilot step is separate from Algorithm~\ref{alg:fed-topic-score}; after selecting $\hat K_t^{\mathrm{pil}}$, the main projector-only protocol is rerun with that value.
\end{proof}

% =========================================================
\subsection{Probability space and conditioning convention}\label{app:setup-notes}

Because the use of the shared empirical whitening matrix $M_d$ can obscure what is random and what is deterministic, we record the underlying probability space explicitly.
For each client $m$, the matrix $W^{(m)}$ is regarded as fixed.
Conditional on $W^{(1:M)}$, the document columns of $D^{(m)}$ are independent and satisfy
\[
N d_i^{(m)} \sim \mathrm{Multinomial}\bigl(N,Aw_i^{(m)}\bigr).
\]
Therefore the product probability measure is
\[
\Pbb_{A,W^{(1:M)}}
=
\bigotimes_{m=1}^M\bigotimes_{i=1}^{n_m}
\mathrm{Law}\bigl(d_i^{(m)}\mid A,w_i^{(m)}\bigr).
\]
All expectations and probabilities in the main text are taken with respect to this conditional law.
In particular,
\[
D_0^{(m)}=AW^{(m)},
\qquad
D_0=AW,
\qquad
M_0=\diag\Bigl(\frac{1}{n_{\mathrm{tot}}}D_0\1\Bigr)
\]
are deterministic once $W^{(1:M)}$ is fixed.
This is the fixed-design pLSI setup used throughout the analysis.

The only genuinely random object appearing in the whitening step is $M_d$.
Because the same $M_d$ is shared by all clients, the local matrices $\hat G_m$ are not independent.
This is why the fixed-whitening matrices \(\tilde G_m\) are essential:
\[
\hat G_m
\xleftrightarrow[\text{bridge}]{\ M_d \leftrightarrow M_0\ }
\tilde G_m
\xrightarrow{\ \E[\cdot\mid W^{(1:M)}]\ }
G_{0,m}+D_{0,m}^{\mathrm{svd}}.
\]
The fixed-whitening matrices \(\tilde G_m\) remain independent across clients, so the averaging argument can still be driven by matrix Bernstein inequalities.  The dependence induced by the shared \(M_d\) is absorbed through \(\hat P_m-\tilde P_m\), while the deterministic ordinary-SVD diagonal bias \(D_{0,m}^{\mathrm{svd}}\) is included in the fixed-whitening perturbation radius.
This separation is the key probabilistic simplification in the federated analysis.

If one instead postulates a hierarchical prior on
\(W^{(1:M)}\), the results may be applied conditionally
on the realized design whenever the stated assumptions
hold. Let \(\mathcal G_W\) denote the event that all
required design assumptions hold with fixed regularity
constants. Integrating the conditional guarantee yields
an unconditional failure probability at most
\[
\delta+\mathbb P(\mathcal G_W^c).
\]
The same \(1-\delta\) guarantee therefore follows when
the prior is supported on the regular class; otherwise,
the probability of \(\mathcal G_W^c\) must be controlled
separately.

A second subtle point concerns the choice of target subspace.
Because each client may have a different $W^{(m)}$, the population matrices $G_{0,m}$ are not numerically identical.
Nevertheless, as long as each $\Sigma_{W,m}$ is full rank, all $G_{0,m}$ share the same leading eigenspace:
\[
\mathrm{col}(G_{0,m})=\mathrm{col}(M_0^{-1/2}AW^{(m)})=\mathrm{col}(M_0^{-1/2}A).
\]
Hence there is still a common deterministic projector $P$.
This is the correct object to average around.
The bridge decomposition therefore solves two issues simultaneously:
\begin{enumerate}[leftmargin=1.5em]
\item it isolates the conditional sampling fluctuation from the deterministic ordinary-SVD diagonal bias \(D_{0,m}^{\mathrm{svd}}\);
\item it reveals a common deterministic target projector even though the client-wise population Gram matrices have different eigenvalues.
\end{enumerate}

The main text uses this convention uniformly.  Formulas involving conditional expectations are statements about the conditional law of the document noise with the entire fixed design, and hence the pooled matrix \(M_0\), held fixed. No contradiction arises from the fact that one may also place a higher-level probabilistic model on \(W^{(1:M)}\).

% =========================================================

\subsection{Oracle geometry under common whitening}\label{app:oracle-geometry}

We record the oracle geometry under the common whitening $M_0$. The first Topic-SCORE direction is fixed canonically by the pooled marginal, not by an arbitrary eigenbasis.

\begin{lemma}[The pooled marginal matrix is comparable to $H$]\label{lem:M0-h-comparable}
Recall $M_0=\diag(m_0)$ with $m_0=n_{\mathrm{tot}}^{-1}D_0\1$ and $D_0=AW$. Under Assumption~\ref{assump:Sigma},
\[
c h_j\le M_0(j,j)\le h_j,
\qquad 1\le j\le p.
\]
Consequently,
\[
cI_K\preceq A^\top M_0^{-1}A\preceq CI_K.
\]
\end{lemma}

\begin{proof}
Write $w=n_{\mathrm{tot}}^{-1}W\1$. Then $M_0(j,j)=a_j^\top w$. Since every coordinate of $w$ is at most one, $M_0(j,j)\le \|a_j\|_1=h_j$. Also,
\[
w(k)=\frac1{n_{\mathrm{tot}}}\sum_{i=1}^{n_{\mathrm{tot}}}w_i(k)
\ge \frac1{n_{\mathrm{tot}}}\sum_{i=1}^{n_{\mathrm{tot}}}w_i(k)^2
=\Sigma_W(k,k)\ge \lambda_{\min}(\Sigma_W)\ge c.
\]
Therefore $M_0(j,j)\ge ch_j$. The matrix inequalities follow by comparing $M_0$ with $H$ entrywise and using $\Sigma_A=A^\top H^{-1}A$.
\end{proof}

Set
\[
\xi_1:=M_0^{1/2}\1.
\]
Because $\1^\top m_0=1$, this vector has Euclidean norm one. Choose any orthonormal basis $\Xi_\perp$ of $\col(M_0^{-1/2}A)\cap\xi_1^\perp$ and define
\[
\Xi=[\xi_1,\Xi_\perp].
\]

\begin{lemma}[Canonical basis and change of coordinates]\label{lem:oracle-V-geometry}
There exists a unique nonsingular matrix $V=[V_1,\dots,V_K]\in\R^{K\times K}$ such that
\[
\Xi=M_0^{-1/2}AV.
\]
Its first column is exactly the pooled mean topic-weight vector
\[
V_1=w:=\frac1{n_{\mathrm{tot}}}W\1.
\]
Moreover,
\[
VV^\top=(A^\top M_0^{-1}A)^{-1}.
\]
In particular, every coordinate of $V_1$ is bounded away from zero and infinity.
\end{lemma}

\begin{proof}
Since $M_0^{-1/2}A$ has full column rank and the same column space as $\Xi$, the matrix $V$ exists uniquely. Also
\[
M_0^{-1/2}Aw=M_0^{-1/2}m_0=M_0^{1/2}\1=\xi_1,
\]
so uniqueness forces $V_1=w$. Multiplying $\Xi=M_0^{-1/2}AV$ on the left by $\Xi^\top$ yields
\[
I_K=\Xi^\top\Xi=V^\top A^\top M_0^{-1}AV,
\]
equivalently $VV^\top=(A^\top M_0^{-1}A)^{-1}$. By Lemma~\ref{lem:M0-h-comparable}, $A^\top M_0^{-1}A$ is uniformly well conditioned. Also each coordinate of $w$ is at least $\lambda_{\min}(\Sigma_W)$, proving strict positivity and boundedness.
\end{proof}

For each topic $k$, define
\[
v_k^\star(\ell)=\frac{V_{k,\ell+1}}{V_{k1}},
\qquad 1\le \ell\le K-1,
\]
and for each word $j$, define
\[
\pi_j(k)=\frac{A(j,k)V_{k1}}{\sum_{\ell=1}^K A(j,\ell)V_{\ell1}}.
\]

\begin{lemma}[Oracle simplex geometry]\label{lem:oracle-identity}
Writing the oracle ratio matrix as
\[
R(j,\ell)=\frac{\Xi_\perp(j,\ell)}{\xi_1(j)},
\]
one has
\[
r_j=\sum_{k=1}^K\pi_j(k)v_k^\star,
\qquad
A\diag(V_1)=M_0^{1/2}\diag(\xi_1)\Pi,
\qquad \Pi=[\pi_1,\dots,\pi_p]^\top.
\]
If word $j$ is an anchor word for topic $k$, then $r_j=v_k^\star$. Hence the oracle ratio cloud lies in a nondegenerate simplex with vertices $v_1^\star,\dots,v_K^\star$.
\end{lemma}

\begin{proof}
Because $\Xi=M_0^{-1/2}AV$,
\[
[1,R]=\diag(\xi_1)^{-1}M_0^{-1/2}AV.
\]
Write $V=\diag(V_1)[1,V^\star]$, where $V^\star(k,\ell)=V_{k,\ell+1}/V_{k1}$. Then
\[
[1,R]=\Pi[1,V^\star],
\qquad
\Pi:=\diag(\xi_1)^{-1}M_0^{-1/2}A\diag(V_1).
\]
This proves the convex-combination formula and the identity $A\diag(V_1)=M_0^{1/2}\diag(\xi_1)\Pi$. If row $j$ of $A$ has only one nonzero entry at topic $k$, then $\pi_j=e_k$ and $r_j=v_k^\star$. Finally,
\[
\begin{pmatrix}
1&\cdots&1\\
v_1^\star&\cdots&v_K^\star
\end{pmatrix}^\top
=\diag(V_1)^{-1}V,
\]
so the simplex is nondegenerate because both factors on the right are nonsingular.
\end{proof}

\begin{lemma}[Row scales]\label{lem:oracle-row-geometry}
For every $j\in[p]$,
\[
\xi_1(j)=\sqrt{M_0(j,j)}\asymp \sqrt{h_j},
\qquad
\|e_j^\top\Xi\|_2\le C\sqrt{h_j},
\qquad
\|r_j\|_2\le C.
\]
\end{lemma}

\begin{proof}
The first statement is immediate from Lemma~\ref{lem:M0-h-comparable}. For the second,
\[
e_j^\top\Xi=M_0(j,j)^{-1/2}a_j^\top V,
\]
so $\|e_j^\top\Xi\|_2\le M_0(j,j)^{-1/2}\|a_j\|_1\|V\|\le C\sqrt{h_j}$. The last bound follows because each ratio point lies in the convex hull of the bounded vertices $v_1^\star,\dots,v_K^\star$.
\end{proof}

\paragraph{Interface to the deterministic propagation.}
Appendix~\ref{app:granular-fedA} uses exactly three oracle inputs: the identity $A\diag(w)=M_0^{1/2}\diag(\xi_1)\Pi$; the row-scale bounds $\xi_1(j)\asymp\sqrt{h_j}$ and $\|e_j^\top\Xi\|_2\le C\sqrt{h_j}$; and the bounded geometry of the simplex matrix. Once these facts are in place, the remainder of the Topic-SCORE propagation is deterministic.

\subsection{Additional technical details for topic-matrix recovery}\label{app:granular-fedA}

This appendix records the deterministic Topic-SCORE propagation from the canonical basis error of Theorem~\ref{thm:fed-rowwise} to the topic-matrix error of Theorem~\ref{thm:fed-A}. Throughout, the empirical spectral basis is the canonical pair $[\hat\xi_1,\hat\Xi_\perp O_\perp]$, not an arbitrary top-$K$ eigenbasis. We use the shorthand
\[
\Delta_2:=\max_{1\le j\le p}h_j^{-1/2}\Bigl\|e_j^\top\bigl([\hat\xi_1,\hat\Xi_\perp O_\perp]-[\xi_1,\Xi_\perp]\bigr)\Bigr\|_2.
\]

\begin{lemma}[Scale of the canonical positive direction]\label{lem:granular-xi1}
Under Assumptions~\ref{assump:hmin}--\ref{assump:Sigma}, there exist constants $c_-,c_+>0$ such that
\[
c_-\sqrt{h_j}\le \xi_1(j)\le c_+\sqrt{h_j},
\qquad 1\le j\le p.
\]
Moreover, if $\Delta_1(D,D_0)$ and $\Delta_2$ are sufficiently small, then $\hat\xi_1(j)\asymp \sqrt{h_j}$ uniformly in $j$.
\end{lemma}

\begin{proof}
The deterministic bounds are Lemma~\ref{lem:oracle-row-geometry}. For the empirical vector, use $|\hat\xi_1(j)-\xi_1(j)|\le \Delta_2\sqrt{h_j}$ and the smallness of $\Delta_2$.
\end{proof}

\begin{lemma}[Ratio perturbation]\label{lem:granular-ratio}
Under the same assumptions,
\[
\max_{1\le j\le p}\|O_\perp^\top\hat r_j-r_j\|_2\le C\Delta_2.
\]
\end{lemma}

\begin{proof}
By definition,
\[
O_\perp^\top\hat r_j-r_j
=\frac{e_j^\top(\hat\Xi_\perp O_\perp-\Xi_\perp)}{\hat\xi_1(j)}
+e_j^\top\Xi_\perp\Bigl(\frac1{\hat\xi_1(j)}-\frac1{\xi_1(j)}\Bigr).
\]
Use $\hat\xi_1(j)\asymp\xi_1(j)\asymp\sqrt{h_j}$, $\|e_j^\top\Xi_\perp\|_2\le C\sqrt{h_j}$, and the definition of $\Delta_2$.
\end{proof}

\begin{lemma}[Barycentric perturbation]\label{lem:granular-bary}
Let
\[
Q=\begin{pmatrix}1&\cdots&1\\ v_1^\star&\cdots&v_K^\star\end{pmatrix},
\qquad
\hat Q=\begin{pmatrix}1&\cdots&1\\ O_\perp^\top\hat v_1&\cdots&O_\perp^\top\hat v_K\end{pmatrix}.
\]
Then for all $j$,
\[
\|\hat\pi_j-\pi_j\|_1\le C\{\Delta_2+\Err_{\mathrm{VH}}(O_\perp)\}.
\]
\end{lemma}

\begin{proof}
The unconstrained barycentric coordinates satisfy
\[
\hat\pi_j^\star=\hat Q^{-1}\binom{1}{O_\perp^\top\hat r_j},
\qquad
\pi_j=Q^{-1}\binom{1}{r_j}.
\]
The simplex matrix $Q$ is uniformly well conditioned by Lemma~\ref{lem:oracle-identity}, and $\|\hat Q-Q\|\le C\Err_{\mathrm{VH}}(O_\perp)$. Therefore inverse perturbation plus Lemma~\ref{lem:granular-ratio} gives $\|\hat\pi_j^\star-\pi_j\|_2\le C(\Delta_2+\Err_{\mathrm{VH}})$. Truncation and renormalization are locally Lipschitz near the simplex, yielding the same order in $\ell_1$.
\end{proof}

\begin{lemma}[Pre-normalized row perturbation]\label{lem:granular-preA}\label{lem:granular-Astar}
Let
\[
\hat A^*:=M_d^{1/2}\diag(\hat\xi_1)\hat\Pi,
\qquad
A^*:=A\diag(w)=M_0^{1/2}\diag(\xi_1)\Pi.
\]
Then
\[
\|\hat a_j^*-a_j^*\|_1
\le Ch_j\{\Delta_1(D,D_0)+\Delta_2+\Err_{\mathrm{VH}}(O_\perp)\}.
\]
\end{lemma}

\begin{proof}
Write
\[
\hat a_j^*-a_j^*
=\sqrt{M_d(j,j)}\hat\xi_1(j)(\hat\pi_j-\pi_j)
+\sqrt{M_d(j,j)}(\hat\xi_1(j)-\xi_1(j))\pi_j
+(\sqrt{M_d(j,j)}-\sqrt{M_0(j,j)})\xi_1(j)\pi_j.
\]
Use $M_0(j,j)\asymp M_d(j,j)\asymp h_j$, $\xi_1(j)\asymp\hat\xi_1(j)\asymp\sqrt{h_j}$, and Lemma~\ref{lem:granular-bary}.
\end{proof}

\begin{lemma}[Column normalization perturbation]\label{lem:granular-colnorm}
Let $\hat A_{\mathrm{fed}}$ be obtained by column-normalizing $\hat A^*$, and let $A$ be obtained by column-normalizing $A^*=A\diag(w)$. Then
\[
\|\hat a_j-a_j\|_1
\le Ch_j\{\Delta_1(D,D_0)+\Delta_2+\Err_{\mathrm{VH}}(O_\perp)\}.
\]
\end{lemma}

\begin{proof}
Let $\hat s_k=\|\hat A_k^*\|_1$ and $s_k=\|A_k^*\|_1=w_k$. Summing Lemma~\ref{lem:granular-preA} over rows gives $|\hat s_k-s_k|\le C(\Delta_1+\Delta_2+\Err_{\mathrm{VH}})$. By Lemma~\ref{lem:oracle-V-geometry}, $w_k$ is bounded away from zero and infinity, and so is $\hat s_k$ for small perturbations. Then
\[
\hat a_j(k)-a_j(k)=\frac{\hat a_j^*(k)-a_j^*(k)}{\hat s_k}+\biggl(\frac1{\hat s_k}-\frac1{s_k}\biggr)a_j^*(k),
\]
and summing over $k$ proves the claim.
\end{proof}

Together, the preceding lemmas prove Lemma~\ref{lem:nonstochastic-fedA}: first control ratios, then barycentric coordinates, then the pre-normalized rows, and finally the column-normalized topic matrix.

\subsection{Additional bridge calculations}\label{app:bridge-calculations}\label{app:bridge-lemmas}

This appendix records the bridge estimates used to compare empirical whitening by $M_d$ with fixed whitening by $M_0$. The correct perturbation object is the relative diagonal matrix
\[
R_M:=M_d^{-1/2}M_0^{1/2}-I_p,
\]
not the raw difference $M_d^{-1/2}-M_0^{-1/2}$. The raw difference can have operator norm of order $\Delta_1(D,D_0)h_{\min}^{-1/2}$, whereas the relative perturbation is uniformly $O(\Delta_1(D,D_0))$.

\begin{lemma}[Relative inverse-square-root perturbation]\label{lem:invroot_diag_perturb}
If $\Delta_1(D,D_0)\le c$ for a sufficiently small constant $c>0$, then
\[
\|R_M\|\le C\Delta_1(D,D_0),
\qquad
R_M:=M_d^{-1/2}M_0^{1/2}-I_p.
\]
Moreover,
\[
\|u_d-\xi_1\|_2\le C\Delta_1(D,D_0),
\qquad
|u_d(j)-\xi_1(j)|\le C\Delta_1(D,D_0)\sqrt{h_j}.
\]
\end{lemma}

\begin{proof}
Because all matrices are diagonal, write $m_j=M_0(j,j)$ and $\hat m_j=M_d(j,j)$. The event $\Delta_1(D,D_0)\le c$ gives
\[
|\hat m_j-m_j|\le \Delta_1(D,D_0)h_j,
\qquad
m_j\asymp h_j,
\]
hence $|\hat m_j/m_j-1|\le C\Delta_1(D,D_0)$. Applying the mean-value theorem to $x\mapsto x^{-1/2}$ on a neighborhood of one gives
\[
\left|\left(\frac{\hat m_j}{m_j}\right)^{-1/2}-1\right|\le C\Delta_1(D,D_0),
\]
which is exactly the operator bound for $R_M$. The bounds for $u_d-\xi_1$ follow from the mean-value theorem for $x^{1/2}$ and $\sum_jh_j=K$.
\end{proof}

\begin{lemma}[Bridge bound for \texorpdfstring{$\hat G_m-\tilde G_m$}{Ghat-Gtilde}]\label{lem:Gbridge_bound}
Suppose \(\Delta_1(D,D_0)\le c\). Let
\[
T_m:=M_0^{-1/2}\hat S_mM_0^{-1/2}=\tilde G_m+\frac1N I_p.
\]
Then
\[
\hat G_m-\tilde G_m
=R_MT_m+T_mR_M+R_MT_mR_M.
\]
Consequently,
\[
\|\hat G_m-\tilde G_m\|
\le C\Delta_1(D,D_0)\|T_m\|.
\]
Furthermore, whenever \(\|T_m\|\le C\),
\[
\|(\hat G_m-\tilde G_m)e_j\|_2
\le
C\sqrt{h_j}\Delta_1(D,D_0)
+C\Delta_1(D,D_0)\|(\tilde G_m-G_{0,m})e_j\|_2
+C\frac{\Delta_1(D,D_0)}{N}.
\]
\end{lemma}

\begin{proof}
Since \(M_d^{-1/2}= (I+R_M)M_0^{-1/2}\),
\[
\hat G_m+\frac1N I_p=(I+R_M)T_m(I+R_M),
\qquad
\tilde G_m+\frac1N I_p=T_m.
\]
Subtracting the two displays gives the stated bridge identity. The operator bound follows immediately from Lemma~\ref{lem:invroot_diag_perturb}:
\[
\|\hat G_m-\tilde G_m\|
\le (2\|R_M\|+\|R_M\|^2)\|T_m\|
\le C\Delta_1(D,D_0)\|T_m\|.
\]

For the column bound, expand
\[
(\hat G_m-\tilde G_m)e_j
=R_MT_me_j+T_mR_Me_j+R_MT_mR_Me_j.
\]
The first and third terms are bounded by \(C\Delta_1(D,D_0)\|T_me_j\|_2\) and a smaller quadratic analogue. Since
\[
T_me_j=(\tilde G_m-G_{0,m})e_j+G_{0,m}e_j+\frac1N e_j
\]
and the oracle row-scale bound gives \(\|G_{0,m}e_j\|_2\le C\sqrt{h_j}\), these terms contribute the stochastic column term, the deterministic \(\sqrt{h_j}\Delta_1\) term, and the identity-shift residue \(C\Delta_1/N\). For the middle term, \(R_Me_j=r_j^{(M)}e_j\) with \(|r_j^{(M)}|\le C\Delta_1(D,D_0)\), so it is controlled by the same bound. This proves the lemma.
\end{proof}

\subsubsection{Why shared empirical whitening does not destroy averaging}
The averaged empirical projector admits the decomposition
\[
\bar P-P
=
\frac1M\sum_{m=1}^M(\tilde P_m-P)
+
\frac1M\sum_{m=1}^M(\hat P_m-\tilde P_m).
\]
The first average consists of independent fixed-whitening ordinary-SVD projectors and is handled by matrix and vector Bernstein inequalities; its radius already includes the deterministic diagonal-bias term \(D_{0,m}^{\mathrm{svd}}\). The second average contains the dependence induced by the shared \(M_d\), but each summand is perturbative because it carries the additional factor \(\Delta_1(D,D_0)\).

\subsection{Additional technical details for row-wise canonical recovery}\label{app:granular-rowwise}

The proof of Theorem~\ref{thm:fed-rowwise} proceeds through projectors rather than through an arbitrary top-$K$ eigenbasis. The synchronized chain is
\[
\tilde G_m-G_{0,m}
\to \tilde P_m-P
\to \bar P^{(0)}-P
\to \bar P-P
\to \widehat P-P
\to (\hat\xi_1,\hat\Xi_\perp)
\to \hat R.
\]
The key point is that the row-wise basis bound is not a statement about an arbitrary fully rotated top-$K$ basis. The first column is fixed by the projected pooled marginal,
\[
\hat\xi_1=\frac{\widehat P M_d^{1/2}\mathbf 1}{\|\widehat P M_d^{1/2}\mathbf 1\|_2},
\]
and only the residual basis is aligned by $O_\perp\in\mathbb R^{(K-1)\times(K-1)}$.

\begin{lemma}[Combined averaged-projector row control]\label{lem:granular-combined-projector}
Under the assumptions of Theorem~\ref{thm:fed-rowwise}, with probability at least $1-\delta$,
\[
\|\bar P-P\|+\max_jh_j^{-1/2}\|(\bar P-P)e_j\|_2
\le C\varepsilon_{n,M}(\delta).
\]
\end{lemma}

\begin{proof}
Decompose \(\bar P-P=(\bar P^{(0)}-P)+(\bar P-\bar P^{(0)})\). The fixed-whitening term is controlled by Theorem~\ref{thm:projector-avg}, Lemma~\ref{lem:vector-bernstein}, and Lemma~\ref{lem:tau-bound-strong}; this term already contains the deterministic ordinary-SVD diagonal bias. The empirical-whitening projector bridge term is controlled by Lemma~\ref{lem:tau-bound-strong}, the clientwise column event from Lemma~\ref{lem:Ecol_hp}, and the pooled marginal concentration bound. This is exactly the decomposition carried out in the proof of Theorem~\ref{thm:fed-rowwise}.
\end{proof}

\begin{lemma}[From averaged-projector rows to the canonical pair]\label{lem:granular-projector-to-canonical}
If
\[
\|\bar P-P\|+\max_jh_j^{-1/2}\|(\bar P-P)e_j\|_2\le \eta
\]
with $\eta+\Delta_1(D,D_0)\le c$ for a sufficiently small constant $c>0$, then there exists an orthogonal matrix $O_\perp\in\mathbb R^{(K-1)\times(K-1)}$ such that
\[
\max_jh_j^{-1/2}
\left\|e_j^\top\bigl([\hat\xi_1,\hat\Xi_\perp O_\perp]-[\xi_1,\Xi_\perp]\bigr)\right\|_2
\le C\{\eta+\Delta_1(D,D_0)\}.
\]
\end{lemma}

\begin{proof}
Apply Lemma~\ref{lem:proj-col-from-avg} to pass from $\bar P-P$ to $\widehat P-P$. Then apply Lemma~\ref{lem:canonical-first} to recover the first direction and Lemma~\ref{lem:residual-basis} to recover the residual basis. This is exactly the canonical-basis step in Theorem~\ref{thm:fed-rowwise}.
\end{proof}

% =========================================================
\subsection{Overspecified local SVD rank}\label{app:practical-svd-rank}

This appendix gives a deterministic sufficient condition under which increasing the local SVD upload rank reduces the subspace-error upper bound in the multi-site setting. This discussion is optional and does not change Algorithm~\ref{alg:fed-topic-score} or the main theory, which use rank \(K\) spectral uploads. The real-data implementation uses a modest overspecified working rank
before the server extracts the final \(K\)-dimensional spectral subspace,
motivated by the calculation below.

For each client, fix one ordered local SVD with a deterministic measurable rule for ties, and let $Q_m^{(L)}$ project onto its first $L$ left singular vectors, where $K\le L\le p$. Thus, for a fixed client, the SVD subspaces are nested,
\[
Q_m^{(K)}\preceq Q_m^{(K+1)}\preceq\cdots.
\]
Let $P$ be the rank-$K$ oracle projector onto the common topic space. For any fixed client and any $L\ge K$, define the signal leakage
\[
\lambda_m(L):=\|(I-Q_m^{(L)})P\|.
\]
Because the subspaces are nested,
\begin{equation}\label{eq:leakage-monotone}
\lambda_m(L+1)\le \lambda_m(L),
\qquad L\ge K.
\end{equation}
This monotonicity is deterministic: projecting onto a larger local SVD subspace cannot increase the residual norm of any vector in the true signal space. The caveat is that a larger local rank can also include more off-signal directions. The next proposition gives a sufficient condition under which the improved signal retention dominates the extra contamination.

\begin{proposition}[A sufficient condition for improvement with an overspecified local rank]\label{prop:overspecified-rank-improvement}
For $L\ge K$, set
\[
\bar Q^{(L)}:=\sum_{m=1}^M\alpha_m Q_m^{(L)},
\qquad
\widehat P^{(L)}:=\text{the top-}K\text{ projector of }\bar Q^{(L)}.
\]
Define
\[
a_L:=\|(I-P)\bar Q^{(L)}P\|+
\|P-P\bar Q^{(L)}P\|,
\qquad
b_L:=\|(I-P)\bar Q^{(L)}(I-P)\|.
\]
If $1-b_L\ge c_*>0$ and $a_L\le c_*/4$, then
\begin{equation}\label{eq:rank-L-subspace-bound}
\|\widehat P^{(L)}-P\|
\le C_{c_*}a_L.
\end{equation}
Consequently, if $L_2>L_1\ge K$, $a_{L_2}<a_{L_1}\le c_*/4$, and
the off-signal bound $1-b_L\ge c_*$ holds for both ranks, then the
upper bound in~\eqref{eq:rank-L-subspace-bound} decreases when the local
upload rank is increased from $L_1$ to $L_2$.
\end{proposition}

\begin{proof}
Write $\bar Q^{(L)}$ in the block decomposition induced by $P\oplus(I-P)$:
\[
\bar Q^{(L)}=
\begin{pmatrix}
P\bar Q^{(L)}P & P\bar Q^{(L)}(I-P)\\
(I-P)\bar Q^{(L)}P & (I-P)\bar Q^{(L)}(I-P)
\end{pmatrix}.
\]
The definition of $a_L$ implies
\[
P\bar Q^{(L)}P\succeq (1-a_L)P,
\qquad
\|(I-P)\bar Q^{(L)}P\|\le a_L.
\]
The assumption on $b_L$ gives that the operator restricted to $P^\perp$ has norm at most $1-c_*$. Hence the $K$ signal eigenvalues of $\bar Q^{(L)}$ are separated from the off-signal spectrum by a gap at least $c_*/2$ after an $O(a_L)$ perturbation. Davis--Kahan applied to this block matrix yields \eqref{eq:rank-L-subspace-bound}. The final statement follows immediately.
\end{proof}

This argument concerns the multi-site regime. For $M=1$ and $L>K$, \[ \dim\{\operatorname{range}(Q^{(L)})\cap \operatorname{range}(P)^\perp\}\ge L-K>0, \] so $b_L=1$ and the proposition's off-signal condition cannot hold. Moreover, the top-$K$ subspace of a rank-$L$ projector is not unique. A single-site overspecified-rank procedure therefore requires additional spectral information or a different extraction rule. In the multi-site case, a topic-loading guarantee additionally requires the row-normalized control and smallness conditions used by the deterministic Topic-SCORE propagation lemma.

\subsection{Overview of the proof structure}\label{app:roadmap}

This appendix summarizes the dependence structure of the main arguments. The proof has three layers.

\paragraph{Layer 1: client-level bridge decomposition.}
For each client,
\[
\hat G_m-G_{0,m}
=\{\tilde G_m-\E(\tilde G_m\mid W^{(1:M)})\}
+D_{0,m}^{\mathrm{svd}}
+(\hat G_m-\tilde G_m).
\]
The first summand is the centered multinomial fluctuation under deterministic whitening \(M_0\). The second is the deterministic diagonal bias from using the practical ordinary-SVD identity shift. The third summand is the empirical-whitening correction created by replacing \(M_0\) with the shared \(M_d\). In the bridge analysis the empirical-whitening correction is controlled through the relative diagonal perturbation
\[
R_M=M_d^{-1/2}M_0^{1/2}-I_p,
\]
which avoids the spurious $h_{\min}^{-1/2}$ factor that would appear in the raw difference $M_d^{-1/2}-M_0^{-1/2}$.

\paragraph{Layer 2: projector averaging.}
The server averages the practical SVD projectors \(\hat P_m\). To recover independence, we insert only the fixed-whitening ordinary-SVD projectors \(\tilde P_m\):
\[
\bar P-P=(\bar P-\bar P^{(0)})+(\bar P^{(0)}-P),
\qquad
\bar P=\frac1M\sum_{m=1}^M \hat P_m,\quad
\bar P^{(0)}=\frac1M\sum_{m=1}^M\tilde P_m.
\]
The fixed-whitening average is controlled by the variance-bias projector averaging theorem, with the deterministic diagonal bias \(D_{0,m}^{\mathrm{svd}}\) included in its radius. The shared-whitening contribution is controlled separately through \(\Delta_1(D,D_0)\) and is retained in the deterministic radius \(\varepsilon_{n,M}(\delta)\).

\paragraph{Layer 3: canonical Topic-SCORE propagation.}
The averaged projector first gives $\widehat P-P$. The canonical first direction is then recovered as
\[
\hat\xi_1=\frac{\widehat P M_d^{1/2}\mathbf 1}{\|\widehat P M_d^{1/2}\mathbf 1\|_2},
\]
and the residual basis is aligned only by $O_\perp\in\mathbb R^{(K-1)\times(K-1)}$. The deterministic chain is
\[
\widehat P-P\to [\hat\xi_1,\hat\Xi_\perp O_\perp]\to \hat R\to \hat\Pi\to \hat A^*\to \hat A_{\mathrm{fed}}.
\]
The three errors tracked in the final propagation are
\[
\Delta_1(D,D_0),\qquad
\Delta_2=\max_jh_j^{-1/2}\|e_j^\top([\hat\xi_1,\hat\Xi_\perp O_\perp]-[\xi_1,\Xi_\perp])\|_2,
\qquad
\Err_{\mathrm{VH}}(O_\perp).
\]
They enter as
\begin{align*}
\Delta_2 &\Longrightarrow \max_j\|O_\perp^\top\hat r_j-r_j\|_2,\\
\max_j\|O_\perp^\top\hat r_j-r_j\|_2+\Err_{\mathrm{VH}}(O_\perp)
&\Longrightarrow \max_j\|\hat\pi_j-\pi_j\|_1,\\
\Delta_1+\Delta_2+\Err_{\mathrm{VH}}(O_\perp)
&\Longrightarrow \max_jh_j^{-1}\|\hat a_j^*-a_j^*\|_1,\\
\max_jh_j^{-1}\|\hat a_j^*-a_j^*\|_1
&\Longrightarrow \max_jh_j^{-1}\|\hat a_j-a_j\|_1.
\end{align*}

\paragraph{Where each theorem is used.}
Lemma~\ref{lem:Gbridge_bound} controls the empirical-whitening bridge. Lemma~\ref{lem:tau-bound-strong} supplies fixed-whitening projector moments and bias. Theorem~\ref{thm:projector-avg} and Lemma~\ref{lem:vector-bernstein} control the averaged projector. Lemmas~\ref{lem:proj-col-from-avg}, \ref{lem:canonical-first}, and \ref{lem:residual-basis} convert the averaged projector into the canonical basis. Lemma~\ref{lem:nonstochastic-fedA} then converts the canonical-basis error into the topic-matrix error.

\section{Additional simulation details}\label{app:sim-details}

\subsection{Simulation methods}\label{app:sim-methods}

The simulations follow the shared-topic federated model in
Section~\ref{subsec:model}.  In the base setting, \(M=5\) sites each contribute
\(n_m=200\) patients, giving \(n_{\rm tot}=1000\) patients in total.  The base
vocabulary size is \(p=1000\), the number of latent components is \(K=5\), and each
patient has count length \(N=2000\).  A single topic loading matrix
\(A\in\mathbb R^{p\times K}\) is shared across all sites, while the membership
weights \(w_i^{(m)}\) are generated according to either the homogeneous or
heterogeneous membership design below.

The topic loading matrix is generated with an anchor-concept structure.  For each
component \(k\), \(a_p\) concepts are designated as anchors for that component; in
the base setting \(a_p=p/100\).  An anchor concept for component \(k\) has loading
\(\delta_p\) on component \(k\), zero loading on the other components, and
\(\delta_p=1/p\) in the base setting.  The remaining non-anchor rows are generated
from independent positive random entries and rescaled so that each column of \(A\)
is a probability distribution over concepts.  Thus \(a_p\) controls the number of
rows available to identify each simplex vertex, while \(\delta_p\) controls the
signal strength of each anchor concept.  The simulation code uses a fixed base seed
and deterministic setting-specific seed offsets so that each grid value and replicate
corresponds to an independent generated dataset.

Given \(A\) and a patient membership vector \(w_i^{(m)}\), we generate raw concept
counts by
\[
    X_i^{(m)}\sim \mathrm{Multinomial}\{N,Aw_i^{(m)}\},
    \qquad
    d_i^{(m)}=X_i^{(m)}/N .
\]
The local site matrix is
\(D^{(m)}=[d_1^{(m)},\ldots,d_{n_m}^{(m)}]\).  Topic-SCORE methods use these
frequency matrices, while the LDA benchmark uses the corresponding count matrices
\(X^{(m)}=ND^{(m)}\).  We compare Federated Topic-SCORE
with pooled Topic-SCORE, which applies Topic-SCORE to the centralized
\([D^{(1)},\ldots,D^{(M)}]\), single-site Topic-SCORE, which applies
Topic-SCORE to one local site, and a federated seeded-LDA benchmark, labeled
Federated LDA in the figures.  Topic-SCORE methods use the same vertex-hunting
routine and tuning choices within a given simulation configuration.  Estimation accuracy is
measured by the permutation-aligned \(\ell_1\) distance defined in Section~\ref{sec:theory}. Each simulation setting reports medians and interquartile ranges across 30 independent
replicates.

\paragraph{Federated LDA benchmark.}
The federated LDA benchmark is a practical federated LDA-style comparator motivated
by recent work on federated LDA-based topic modeling~\citep{kushwaha2025federated}.
It matches the non-repeated-cross-site communication regime more closely than
iterative federated variational or neural topic models, while still producing an
interpretable topic loading matrix for direct comparison with \(A\).

Another study on federated LDA targets local differential privacy~\citet{wang2020fedlda}, and using it as a benchmark would conflate topic estimation accuracy with the effects of privacy noise. The federated LDA method used in our simulations thus provides a cleaner non-private baseline for evaluating topic estimation accuracy.

The benchmark method is applied on the same multinomial draws as the Topic-SCORE methods.
For LDA fitting, the frequency matrix \(D^{(m)}\) is converted back to the count
matrix \(X^{(m)}=ND^{(m)}\), with rows interpreted as concepts and columns as
patients.  Each site first fits an unsupervised LDA model with known topic number
\(K\).  The server extracts the local topic-word probability matrices, stacks the
local topics from all sites, normalizes them for cosine comparison, and clusters the
local topics into \(K\) groups.  For each group, the server selects the terms with the highest average probabilities within the group to form a seed dictionary containing one list of representative terms per topic. Sites then refit seeded LDA locally using this dictionary to guide topic estimation, and the server aggregates the resulting topic-word
matrices with sample-size weights \(n_m/n_{\rm tot}\).  The final matrix is
column-normalized and evaluated with the same permutation-aligned \(\ell_1\) loss as
the Topic-SCORE estimators.

The implementation fit the federated LDA with 20 seed terms per topic,
500 local unsupervised-LDA iterations, 500 local seeded-LDA iterations, seed weight
0.05, and sample-size weighted aggregation.  Thus the Federated LDA protocol avoids
patient-level pooling and repeated cross-site model-update communication, but the
site-level LDA fits are iterative and computationally heavier than the spectral method used by Federated Topic-SCORE.

\paragraph{Compute resources.}
The Topic-SCORE-based methods, including pooled, federated, and single-site
Topic-SCORE, are lightweight spectral procedures and each simulation replicate ran in
seconds on a local CPU machine.  The Federated LDA baseline was substantially more
computationally intensive because each replicate required local unsupervised LDA fits
followed by local seeded-LDA refits.  We therefore ran the full simulation array with
the federated LDA baseline on a CPU cluster.  The full grid consisted of 2,040
independent simulation jobs, run with up to 20 concurrent jobs on a CPU partition. The full array completed in about 5
hours of wall-clock time, corresponding to approximately 100 aggregate task-hours. 

\subsection{Homogeneous membership setting}\label{app:sim-extra}

In the homogeneous setting, all sites draw patient membership weights from the same
distribution.  The base design uses normalized positive random weights and includes
a small number \(b_n\) of pure-topic patients for each component.  The pure-topic
patients have \(w_i=e_k\) for one component \(k\), so \(b_n\) controls how strongly
the simulated patient memberships reach the corners of the membership simplex.

Beyond the main-text sensitivity analyses for \(p\), \(n_{\rm tot}\), \(N\), and
site-size imbalance, we vary additional parameters that affect the spectral-simplex
geometry.  These include the number of components \(K\), the number of anchor
concepts \(a_p\), the anchor strength \(\delta_p\), and the number of pure-topic
patients \(b_n\).  Each sensitivity analysis varies one parameter at a time while
holding the other parameters at their base values, except for the site-size
imbalance experiment, which keeps \(n_{\rm tot}=1000\) fixed and changes the
fraction of patients contributed by the site used for the single-site benchmark.

\begin{figure}[!htbp]
\centering
\includegraphics[width=\linewidth]{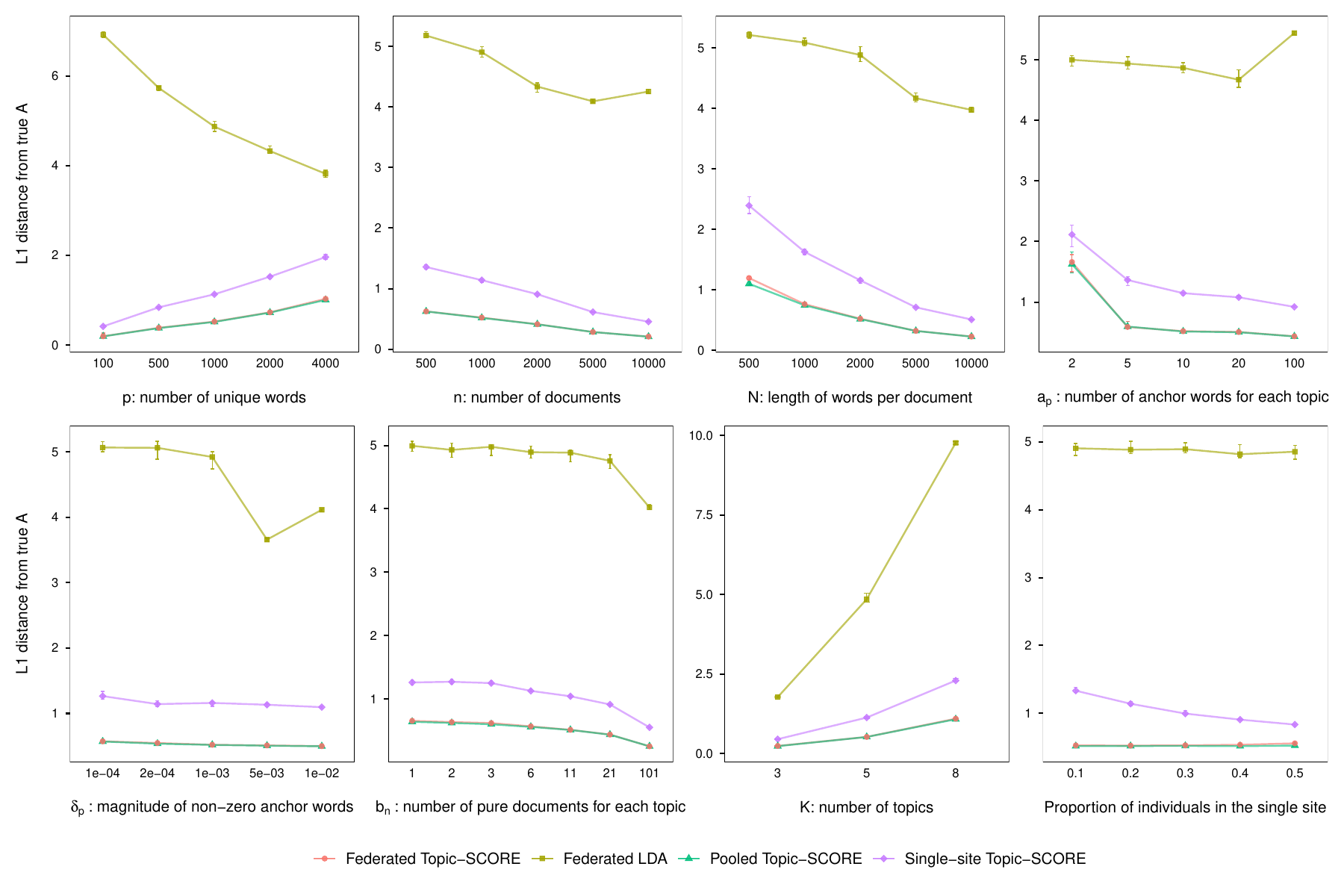}
\caption{Additional homogeneous-membership simulation sensitivity analyses. Points
show median permutation-aligned \(\ell_1\) error for estimating the true topic
loading matrix \(A\). Distinct point shapes and colors identify Federated
Topic-SCORE, Federated LDA, pooled Topic-SCORE, and single-site Topic-SCORE.
Error bars show interquartile ranges across 30 independent simulation replicates.
The Federated Topic-SCORE and pooled Topic-SCORE curves nearly overlap across the
displayed settings.}
\label{fig:sim-homo-extra}
\end{figure}

Across the additional homogeneous settings, Federated Topic-SCORE closely matches the
pooled estimator and remains more accurate than single-site estimation
(Figure~\ref{fig:sim-homo-extra}).  Larger vocabulary size and larger \(K\) make the
estimation harder, whereas larger total sample size and longer patient records reduce
error.  Increasing the number of anchor concepts or the number of pure-topic
patients improves the finite-sample geometry used by vertex hunting.  The federated
estimator is also stable under moderate site-size imbalance, indicating that the
non-iterative summary aggregation is not overly sensitive to one site contributing a
larger share of the total sample.  Federated LDA has substantially larger error
throughout these homogeneous settings.  It improves in easier regimes such as
smaller vocabularies, longer records, or stronger anchor structure, but remains less
accurate than the Topic-SCORE estimators, consistent with the original Topic-SCORE
simulation comparisons in which federated LDA underperformed spectral
Topic-SCORE recovery.

\FloatBarrier

\subsection{Heterogeneous membership setting}\label{app:sim-heterogeneous}

In the heterogeneous simulations, all sites share a common topic loading matrix \(A\). We fix \(p\), \(K\), and \(N\) at their base values and use \(M=5\) sites with \(n_m=200\) patients per site. The Dirichlet concentration parameter is fixed at 10 in the base heterogeneous grid. For skewed patients at site \(m\), let \(q^{(m)}\) denote the mean membership weight vector, with a specified mean weight assigned to one topic and the remaining weight divided equally among the other \(K-1\) topics. We sample each patient's membership weight vector as \(w_i^{(m)}\sim\operatorname{Dirichlet}(10q^{(m)})\). For balanced patients, we sample \(w_i^{(m)}\sim\operatorname{Dirichlet}(10/K,\ldots,10/K)\), giving a uniform mean membership weight vector \((1/K,\ldots,1/K)^\top\). The scenarios below introduce differences in membership weight distributions across sites by varying the mean weights and the proportions of balanced and skewed patients.

We consider four heterogeneous scenarios.  In Scenario 1, all patients at one site is skewed
toward one topic, and the weight assigned to that topic is varied. In Scenario 2, one site contains a mixture of balanced and skewed patients; the skewed
patients place \(0.9\) weight on one topic, and the proportion of skewed patients
is varied.  In Scenario 3, all patients at all sites are skewed, with different sites skewed
toward different topic, and the skewed topic weight is varied.  In
Scenario 4, all sites contain mixtures of balanced and skewed patients, with
different sites skewed toward different topic; the skewed component weight is
fixed at \(0.9\), and the skewed-patient proportion is varied.

\begin{figure}[!htbp]
\centering
\includegraphics[width=\linewidth]{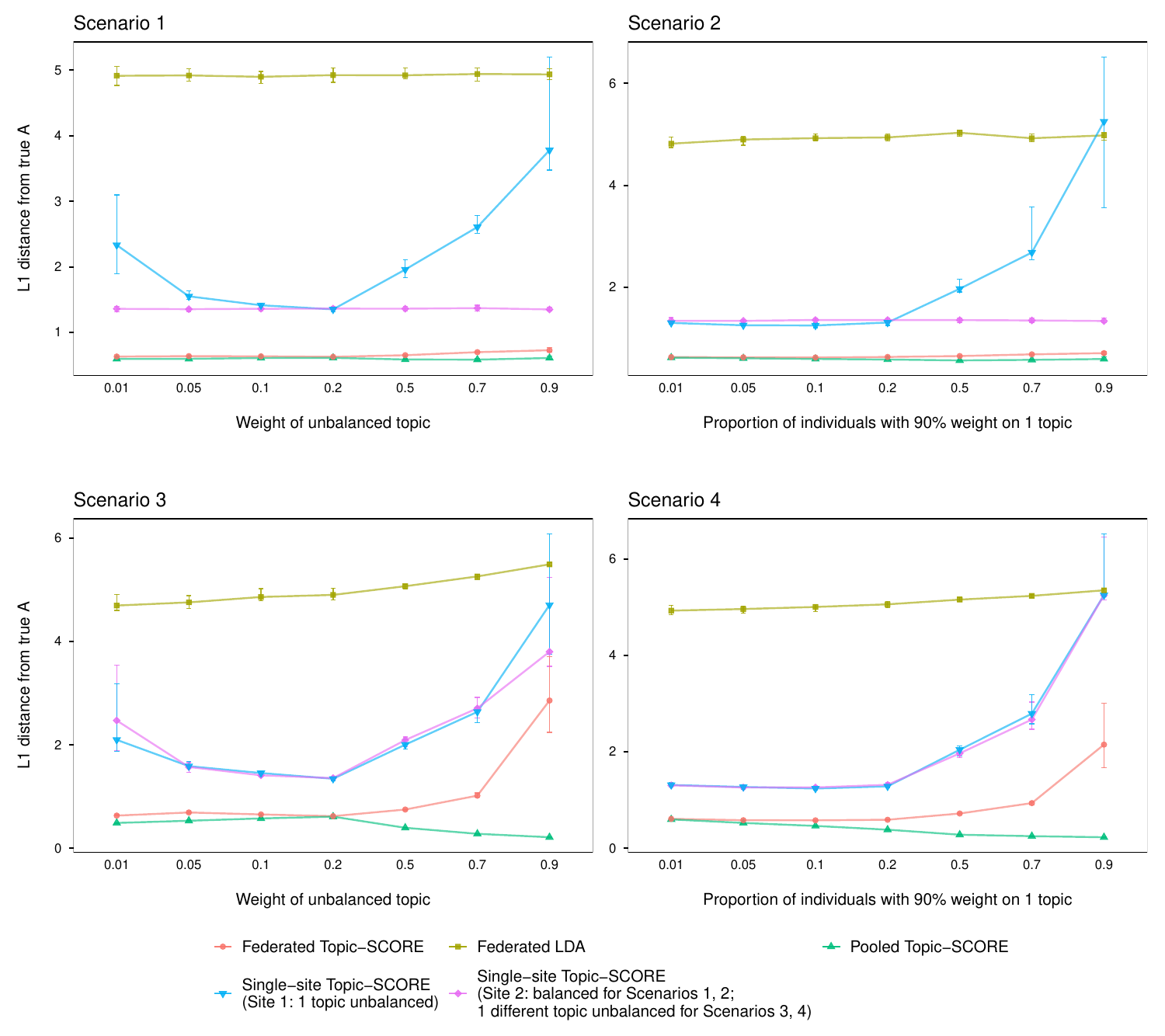}
\caption{Additional heterogeneous-membership simulation scenarios. Points show
median permutation-aligned \(\ell_1\) error for estimating the true topic loading
matrix \(A\). Distinct point shapes and colors identify methods in the shared
legend, which distinguishes the unbalanced Site 1 single-site fit from the Site 2, which is balanced in Scenarios 1 and 2 and differently skewed in Scenarios 3
and 4. Error bars show interquartile ranges across 30 independent simulation
replicates.}
\label{fig:sim-het-extra}
\end{figure}

When only one site has an imbalanced membership weight distribution and the other sites remain balanced, Federated Topic-SCORE achieves estimation accuracy similar to pooled Topic-SCORE, while single-site Topic-SCORE fitted at the imbalanced site becomes less accurate (Figure~\ref{fig:sim-het-extra}, Scenarios 1 and 2). Federated LDA changes relatively little across these settings, but its estimation error remains higher than that of Federated Topic-SCORE.

In Scenarios 3 and 4, topic imbalance affects all sites. Under severe imbalance, concentration on a single topic reduces the information available at each site about the other topics, making the full topic loading matrix harder to estimate from local data. The estimation error of Federated Topic-SCORE increases as the imbalance becomes more severe, consistent with the local richness condition in Assumption~\ref{ass:regular-plsi}, but generally remains lower than that of single-site Topic-SCORE. Federated LDA also has higher estimation error than Federated Topic-SCORE, although the difference becomes smaller under the most severe imbalance.

\FloatBarrier

\section{Real-world EHR application methods}\label{app:real-details}

\subsection{Cohort construction}\label{app:real-cohort}

The real-world application used EHR data from the Mass General Brigham healthcare
system, including Massachusetts General Hospital (MGH), Newton--Wellesley Hospital
(NWH), Brigham and Women's Hospital (BWH), and Spaulding Rehabilitation Hospital
(SRH).
Eligible patients had an inpatient acute ischemic stroke admission during January 1,
2005 through January 31, 2025.  The index event was defined as the first qualifying
inpatient acute ischemic stroke admission for each patient.  Acute ischemic stroke
was identified from the principal diagnosis code of the inpatient encounter, using
ICD-10 codes beginning with I63 and, for the ICD-9 era, codes of the form 433.x1 or
434.x1. Each patient was assigned to the facility associated with the index admission.  This
site assignment defines the local EHR matrix used by each hospital in the
federated analysis. 

\subsection{Concept construction}\label{app:real-concepts}

Observed ICD-9 and ICD-10 diagnosis codes were mapped to phecodes.  When multiple
granularities were available for a code, the pipeline used a stable coarser
representation: a one-decimal phecode when available, otherwise an integer-level
phecode, and otherwise a two-decimal phecode folded to one decimal. 

We used the diagnosis group represented by the phecodes as the EHR concepts. We then constructed a concept-by-patient matrix by counting the occurrence of each diagnosis group or phecode for each patients. The same construction could be
generalized to concepts related to medication, procedure, laboratory values, or other standardized EHR concepts once they are included in the
vocabulary.

In this analysis, we only included incident post-stroke concepts in the vocabulary to adjust for patients' baseline conditions. A concept was treated as incident for a patient if it was observed during the 30 days after the index admission and was not observed in that patient's pre-index prevalent window. The prevalent window covered the five years before the index admission and ended seven days before the index admission. 

\subsection{Concept-by-patient matrix and site contributions}
\label{app:real-matrix-sites}

For each patient, we counted retained incident concepts and formed a
concept-by-patient matrix. To improve the stability and interpretability of the recovered topics, we
retained concepts appearing in at least 20 patients and excluded concepts appearing
in more than 80\% of patients.  We also retained only patients with at least 20 total
counts across the retained concepts.  Each patient row was then normalized by its
total retained count to form concept frequencies.  In the notation of the main
method, each site stores the transpose of its local concept-by-patient matrix,
\(D^{(m)}\in\mathbb R^{p\times n_m}\).

Patients were assigned to sites where their index stroke admission happened. Patients with ambiguous multi-site assignment at the index admission were
excluded from the analysis. Sites contributing more than 100 eligible
patients were retained, yielding counts of 490 concepts for 7,057 patients across four hospitals: MGH
(\(n=4{,}358\)), NWH (\(n=154\)), BWH (\(n=2{,}254\)), and SRH
(\(n=291\)).

% CAMERA C-01: rebuttal q2yK W2, lines 133--155.
The table below summarizes demographic characteristics overall and stratified by site. The average pairwise standardized mean differences (SMDs) for age at admission and sex at birth exceeded 0.2, indicating descriptive differences in demographic composition and potential heterogeneity in population across sites. Note that SMDs are calculated for a variable only among individuals without missing value for that variable. 

\begin{table}[htbp]
\centering
\caption{Demographic characteristics overall and across hospitals. Hospitals: Massachusetts General Hospital (MGH), Newton--Wellesley Hospital (NWH), Brigham and Women's Hospital (BWH), and Spaulding Rehabilitation Hospital (SRH). SD: standard deviation; SMD: standardized mean difference, averaged across hospital pairs.}
\label{tab:camera-demographics}
\resizebox{\linewidth}{!}{%
\begin{tabular}{lrrrrrr}
\toprule
Variable & Overall & MGH & NWH & BWH & SRH & SMD \\
\midrule
number of patients (n) & 7,057 & 4,358 & 154 & 2,254 & 291 &  \\
Age at admission, years, mean (SD) & 66.8 (15.6) & 66.2 (15.8) & 74.6 (14.1) & 67.6 (15.0) & 65.8 (17.3) & 0.306 \\
\quad Missing & 4 (0.1\%) & 4 (0.1\%) & 0 (0.0\%) & 0 (0.0\%) & 0 (0.0\%) &  \\
Sex at birth, n (\%) &  &  &  &  &  & 0.227 \\
\quad Female & 855 (12.1\%) & 487 (11.2\%) & 35 (22.7\%) & 294 (13.0\%) & 39 (13.4\%) &  \\
\quad Male & 1,000 (14.2\%) & 564 (12.9\%) & 24 (15.6\%) & 346 (15.4\%) & 66 (22.7\%) &  \\
\quad Missing & 5,202 (73.7\%) & 3,307 (75.9\%) & 95 (61.7\%) & 1,614 (71.6\%) & 186 (63.9\%) &  \\
Race, n (\%) &  &  &  &  &  & 0.135 \\
\quad White & 5,246 (74.3\%) & 3,251 (74.6\%) & 126 (81.8\%) & 1,675 (74.3\%) & 194 (66.7\%) &  \\
\quad Non-White & 1,384 (19.6\%) & 818 (18.8\%) & 25 (16.2\%) & 469 (20.8\%) & 72 (24.7\%) &  \\
\quad Missing & 427 (6.1\%) & 289 (6.6\%) & 3 (1.9\%) & 110 (4.9\%) & 25 (8.6\%) &  \\
Ethnic group, n (\%) &  &  &  &  &  & 0.145 \\
\quad Hispanic & 190 (2.7\%) & 123 (2.8\%) & 0 (0.0\%) & 62 (2.8\%) & 5 (1.7\%) &  \\
\quad Non-Hispanic & 5,632 (79.8\%) & 3,416 (78.4\%) & 137 (89.0\%) & 1,880 (83.4\%) & 199 (68.4\%) &  \\
\quad Missing & 1,235 (17.5\%) & 819 (18.8\%) & 17 (11.0\%) & 312 (13.8\%) & 87 (29.9\%) &  \\
\bottomrule
\end{tabular}}
\end{table}
\FloatBarrier

\subsection{Model implementation and comparisons}\label{app:real-model}

We used \(K=5\) latent components for interpretability. Pooled
Topic-SCORE was fit to the combined concept-patient matrix as a centralized reference. While the pooling procedure is not the deployable in general setting when sites cannot share individual data with one another, this approach is useful for benchmarking in our demonstrating example. Federated Topic-SCORE was fit to the four site matrices without
sharing patient-level records.  The federated implementation used global rank
\(K=5\) and local working rank \(L=10\), consistent with the overspecified-rank
principle discussed in Section~\ref{subsec:algorithm-main} and
Appendix~\ref{app:practical-svd-rank}.

Single-site Topic-SCORE was fit separately to each site. Federated and single-site estimates were compared with the pooled reference using the permutation-aligned \(\ell_1\) distance between topic loading matrices. Since vertex hunting involves random starts, benchmark comparisons were repeated over 20 replicates.

\subsection{Patient subgrouping and survival analysis}
\label{app:real-subgroup-survival}

After the server estimates the shared component or topic loading matrix \(\widehat A_{\rm fed}\), it
returns \(\widehat A_{\rm fed}\) to the
participating sites.  Each site then estimates local patient membership weights
\(\widehat W^{(m)}\) from its own \(D^{(m)}\), using the weighted least-squares step
in~\eqref{eq:wls-step-main}. 
For our demonstrating subgroup analysis, each patient was assigned to a a subgroup corresponding to the topic with the largest estimated membership weight. We then fit the Kaplan-Meier survival curve for each subgroup as a post-hoc description Patients were followed from the date of index stroke admission until death, with censoring at last encounter or the end of available follow-up.

% CAMERA C-02: complete outcome-agnostic version, rebuttal 108--126.
\subsection{Outcome-agnostic patient profile and subgroup discovery utility of representation learning, topic modeling, and Federated Topic-SCORE applied to EHR analysis}
\label{app:real-downstream}

Representation learning applied to EHR analysis can decompose complex EHR data structure and generate compact patient representations or profiles. The learned representations could potentially be used on various complementary and distinct machine learning tasks such as unsupervised patient subgrouping, supervised patient outcome prediction and generative modeling. While supervised learning and generative modeling could be achieved by other dedicated models, the unsupervised data structure discovery and outcome-agnostic patient profile and subgroup learning is a unique utility of representation learning. Specifically, a representation learning approach can summarize a multi-institutional population through compact and shared clinical profiles that potentially reveal intrinsic biological mechanisms of a disease process and quantify how strongly each patient expresses each profile, without optimizing for a particular outcome. Such outcome-agnostic representations can support patient subgroup identification, characterize clinical heterogeneity across subgroups, and be examined against multiple subsequent outcomes simultaneously and guide comprehensive precision care.

Among representation learning approaches, topic modeling remains actively developed and applied in both methodological studies and scientific applications due to its unique advantage in computation and interpretability. Recent work has used or extended topic models to identify EHR comorbidity subtypes with differential genetic risk~\citep{jiang2023agedependent}, derive reproducible post-acute COVID-19 subphenotypes across healthcare systems~\citep{zhang2023postacute}, model heterogeneous longitudinal clinical notes~\citep{ye2024dynamic}, and integrate pretrained language-model embeddings into topic estimation~\citep{austern2025pptm}. Topic modeling provides direct interpretation at both the representation component and patient levels. Each topic or representation is represented by nonnegative loadings over observed clinical concepts that sum to one, and each patient is represented by nonnegative membership weights over topics that also sum to one. As such, topic loadings naturally characterize the concept composition of each representation or clinical profile, and topic weights naturally characterize the profile structure of a patient. Topic weights can then be used to construct outcome-agnostic patient subgroups that may inform subgroup-specific care strategies given their prognosis for various outcomes, as shown by our real data application.

Federated Topic-SCORE naturally extends the outcome-agnostic patient profile and subgroup discovery utility of representation learning and topic modeling in multi-institutional EHR settings, with its advantage of a non-iterative scheme that significantly mitigates communication and computation costs. To evaluate Federated Topic-SCORE relative to pooled and federated baselines, we applied the same analysis in Section 4.2 to pooled Topic-SCORE and Federated LDA in comparison against Federated Topic-SCORE and compared the degree of separation among the survival curves obtained from the three approaches, each of which learned topic loadings and weights from all sites and is comparable to the others. We evaluated the differentiation in survival patterns across subgroups using the median of pairwise differences (comparing two subgroups at a time) in subgroup-specific restricted mean survival time (RMST) and obtained patient-level bootstrap 95\% confidence intervals (with 1,000 bootstrap samples for each estimator) to quantify the uncertainty, given the widely acknowledged clinical interpretability of RMST.

In addition to the mortality outcome that we examined in the main text, we also examined two more outcomes whose differentiated risk may imply targeted stroke care strategies: the post-index stroke recurrent ischemic stroke admission and intracerebral hemorrhage (ICH) admission, to demonstrate the outcome-agnostic feature of the use case. Given the competing risks imposed by mortality on these two outcomes, we examined the pairwise difference in subgroup-specific restricted mean time lost (RMTL) to evaluate the differentiation in time-to-event patterns across subgroups. Notably, the time-to-event outcome framework is applied to all three outcomes due to commonly observed right-censoring due to loss-to-follow-up in EHR data, but our approach can be generalized to categorical or continuous outcomes.

\begin{table}[htbp]
\centering
\caption{Median pairwise RMST/RMTL difference for 365-day mortality, recurrent ischemic stroke readmission, or intracerebral hemorrhage (ICH) readmission. Values are the median absolute pairwise RMST (mortality) or RMTL (recurrent stroke and ICH) differences (in days) across the 10 comparisons among five dominant-topic patient groups, reported in days with patient-level bootstrap 95\% confidence intervals in parentheses.}
\label{tab:camera-outcome-separation}
\resizebox{\linewidth}{!}{%
\begin{tabular}{lrrr}
\toprule
Outcome & Pooled Topic-SCORE & Federated Topic-SCORE & Federated LDA \\
\midrule
Mortality (RMST) & 41.5 (29.6, 53.2) & 42.2 (28.7, 53.1) & 13.8 (8.1, 22.0) \\
Recurrent stroke (RMTL) & 9.9 (5.2, 21.2) & 9.3 (3.4, 15.2) & 9.0 (5.9, 15.2) \\
ICH (RMTL) & 6.2 (4.8, 8.1) & 5.8 (3.6, 10.8) & 3.8 (2.3, 7.0) \\
\bottomrule
\end{tabular}}

\end{table}

As shown in Table~\ref{tab:camera-outcome-separation}, the mortality differentiation of Federated Topic-SCORE closely overlaps that of the gold standard pooled estimate. Both pooled and Federated Topic-SCORE reached a higher level of RMST difference than Federated LDA, with non-overlapping confidence intervals. The difference in RMTL of recurrent stroke and ICH readmission also shows that the Federated Topic-SCORE approaches pooled Topic-SCORE and has a higher level of differentiation than the federated baseline in point estimate, although with overlapping confidence intervals.

\section{Code and data availability}
\label{app:code-data}

The R and Python implementations and simulation scripts, together with dependency specifications and instructions for running the simulations, are provided at \url{https://github.com/zroyan/Federated-Topic-SCORE}. Synthetic simulation inputs are generated by the scripts. Patient-level EHR data used in the real-world application are not publicly available because they contain protected health information and are governed by institutional data-use restrictions.

\end{document}